\documentclass[11pt]{article}

\usepackage{acl}

\usepackage{amsmath}
\usepackage{times}
\usepackage{latexsym}

\usepackage[T1]{fontenc}
\usepackage[utf8]{inputenc}
\usepackage{microtype}
\usepackage{colortbl}
\usepackage{float}
\usepackage{comment}
\usepackage{array}
\usepackage{adjustbox} 
\usepackage{booktabs}
\usepackage[edges]{forest}
\definecolor{royalblue}{rgb}{0.2549, 0.4118, 0.8824}
\definecolor{limegreen}{rgb}{0.1961, 0.8039, 0.1961}
\definecolor{deeppurple}{rgb}{0.5647, 0.2118, 0.6118}
\definecolor{hidden-draw}{rgb}{0.5,0.5,0.5} 

\usepackage{multirow}
\usepackage{stfloats}

\title{The Visual Counter Turing Test (VCT\textsuperscript{2}): A Benchmark for Evaluating AI-Generated Image Detection and the Visual AI Index (V\textsubscript{AI})} 

\newcommand*{\affaddr}[1]{#1}
\newcommand*{\affmark}[1][*]{\textsuperscript{#1}}

\author{
\parbox{\textwidth}{
\centering
\textbf{Nasrin Imanpour}\affmark[1,*], \textbf{Abhilekh Borah}\affmark[2,*], \textbf{Shashwat Bajpai}\affmark[3,*], \textbf{Subhankar Ghosh}\affmark[4,*], \\\textbf{Sainath Reddy Sankepally}\affmark[5,*], 
\textbf{Hasnat Md Abdullah}\affmark[6], \textbf{Nishoak Kosaraju}\affmark[7], \textbf{Shreyas Dixit}\affmark[8], \textbf{Ashhar Aziz}\affmark[9], \textbf{Shwetangshu Biswas}\affmark[10], 
\textbf{Vinija Jain}\affmark[11], \textbf{Aman Chadha}\affmark[12,13]\footnotemark[2], \textbf{Song Wang}\affmark[14], \textbf{Amit Sheth}\affmark[1], \textbf{Amitava Das}\affmark[15] \\
\vspace{0.4em}
{\normalfont\mdseries
\affaddr{\affmark[1]University of South Carolina, USA},
\affaddr{\affmark[2]Manipal University Jaipur, India},
\affaddr{\affmark[3]BITS Pilani, Hyderabad, India}, 
\affaddr{\affmark[4]Xpectrum AI, USA},
\affaddr{\affmark[5]International Institute of Information Technology, India},
\affaddr{\affmark[6]Texas A\&M University, USA}, 
\affaddr{\affmark[7]Carnegie Mellon University, USA},
\affaddr{\affmark[8]Vishwakarma Institute of Information Technology, India},
\affaddr{\affmark[9]IIIT Delhi, India}, 
\affaddr{\affmark[10]National Institute of Technology, Silchar, India},
\affaddr{\affmark[11]Amazon AI, USA},
\affaddr{\affmark[12]Stanford University, USA},
\affaddr{\affmark[13]Amazon GenAI, USA},
\affaddr{\affmark[14]Shenzhen University of Advanced Technology, China},
\affaddr{\affmark[15]BITS Pilani, Goa, India}
}
}
}

\begin{document}
\maketitle

\renewcommand{\thefootnote}{\fnsymbol{footnote}}
\footnotetext[1]{These authors contributed equally to this work.}
\footnotetext[2]{Work does not relate to position at Amazon.}

\begin{abstract}
\vspace{-3mm}
The rapid progress and widespread availability of text-to-image (T2I) generative models have heightened concerns about the misuse of AI-generated visuals, particularly in the context of misinformation campaigns. Existing AI-generated image detection (AGID) methods often overfit to known generators and falter on outputs from newer or unseen models. We introduce the \textbf{Visual Counter Turing Test (VCT\textsuperscript{2})}, a comprehensive benchmark of 166,000 images, comprising both real and synthetic prompt-image pairs produced by six state-of-the-art T2I systems: Stable Diffusion 2.1, SDXL, SD3 Medium, SD3.5 Large, DALL·E 3, and Midjourney 6. We curate two distinct subsets: \textit{COCO\textsubscript{AI}}, featuring structured captions from MS COCO, and \textit{Twitter\textsubscript{AI}}, containing narrative-style tweets from The New York Times. Under a unified zero-shot evaluation, we benchmark 17 leading AGID models and observe alarmingly low detection accuracy, 58\% on COCO\textsubscript{AI} and 58.34\% on Twitter\textsubscript{AI}. To transcend binary classification, we propose the \textbf{Visual AI Index (V\textsubscript{AI})}, an interpretable, prompt-agnostic realism metric based on twelve low-level visual features, enabling us to quantify and rank the perceptual quality of generated outputs with greater nuance. Correlation analysis reveals a moderate inverse relationship between V\textsubscript{AI} and detection accuracy: Pearson $\rho$ of $-0.532$ on COCO\textsubscript{AI} and $\rho$ of $-0.503$ on Twitter\textsubscript{AI}, suggesting that more visually realistic images tend to be harder to detect, a trend observed consistently across generators. We release 
\href{https://huggingface.co/datasets/NasrinImp/COCO_AI}{COCO\textsubscript{AI}}, 
\href{https://huggingface.co/datasets/NasrinImp/Twitter_AI}{Twitter\textsubscript{AI}}, 
and all \href{https://github.com/nasrinimapour/VCT2}{codes} 
to catalyze future advances in generalized AGID and perceptual realism assessment.
\end{abstract}

\vspace{-10pt}

\begin{figure}[htbp]
        \centering
        \includegraphics[width=0.5\columnwidth]{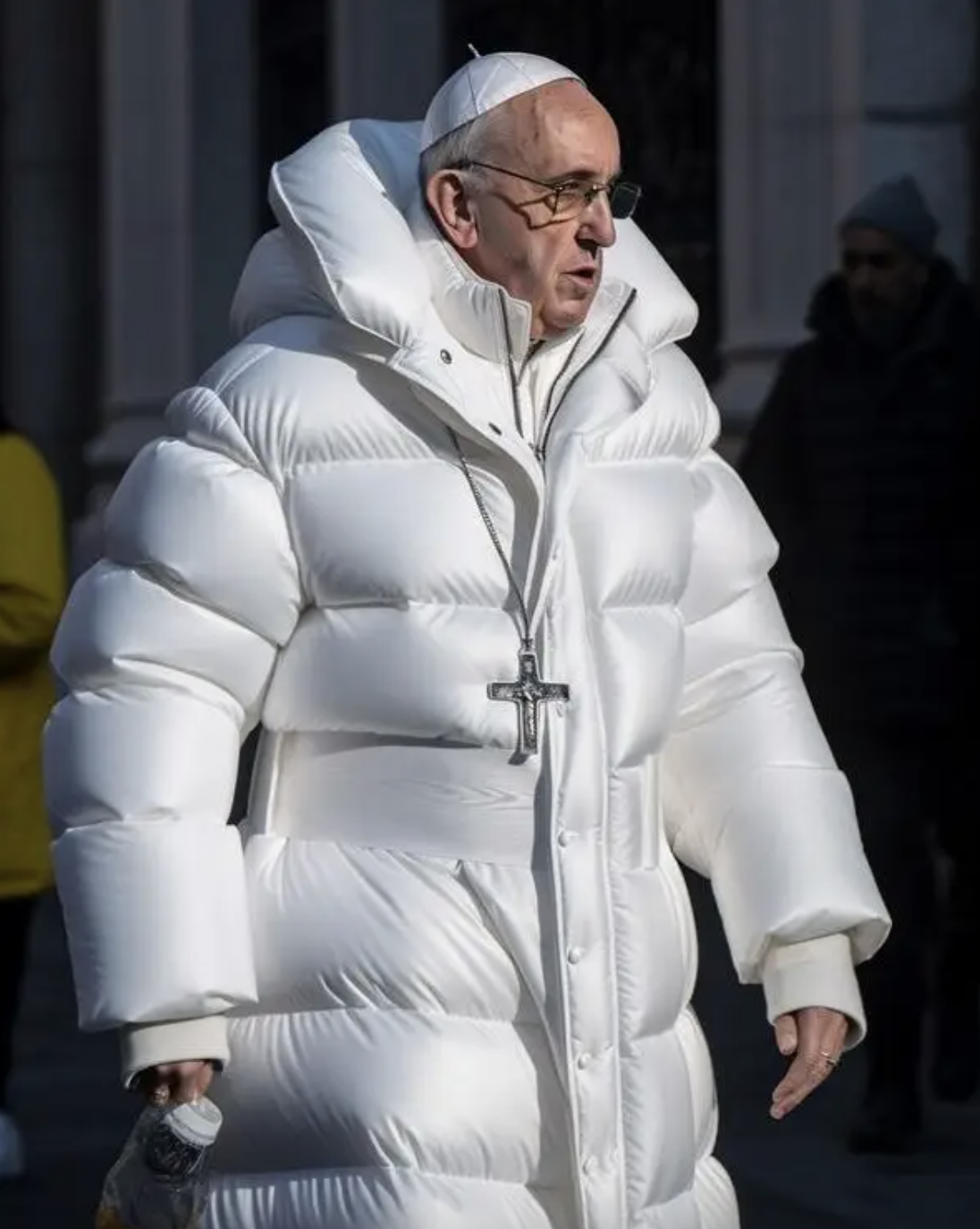}
        \caption{An AI-generated image of Pope Francis wearing a gigantic white puffer jacket went viral on social media platforms like Reddit and Twitter (X) in March 2023. This image sparked widespread media discussions on the potential misuse of generative AI technologies, becoming an iconic example of AI-generated misinformation. For more details, see the \href{https://www.forbes.com/sites/mattnovak/2023/03/26/that-viral-image-of-pope-francis-wearing-a-white-puffer-coat-is-totally-fake/}{Forbes story}.}
    \label{fig:comprehensive1}
    \vspace{-7mm}
\end{figure}
    
\section{Introduction}
\label{sec:intro}

The rapid advancement of text-to-image (T2I) generative models, such as Stable Diffusion~\cite{rombach2022high,podell2023sdxl,esser2024scaling}, DALL·E~\cite{ramesh2021zero,ramesh2022hierarchical,betker2023improving}, Midjourney~\cite{Midjourney2024}, and Imagen~\cite{saharia2022photorealistic}, has revolutionized visual content creation. These models unlock powerful creative workflows and democratize image synthesis at scale. However, their widespread accessibility also raises critical concerns about visual misinformation and content authenticity. As illustrated in Figure~\ref{fig:comprehensive1}, synthetic images can convincingly mimic journalistic or photographic style, blurring the boundary between real and generated content. This growing threat has prompted global attention. In March 2023, an open letter~\cite{aihalt2023} warned that generative AI could destabilize the global information ecosystem. The European Commission reported a significant decline in online content moderation accuracy, from 90.4\% in 2020 to just 64.4\% in 2022~\cite{EU_code_of_conduct_2022}. Meanwhile, social platforms process over 3.2 billion images and 720,000 hours of video daily~\cite{CONVERSATION2023}, with synthetic media projected to account for 90\% of online content by 2026~\cite{europol2024facing}.

Despite increasing demand for reliable detection tools, existing AI-generated image detection (AGID) methods often fail to generalize to images from unseen generators or real-world contexts. Watermark-based approaches remain fragile, easily circumvented via cropping, filtering, or adversarial manipulation~\cite{zhao2025invisible}. Meanwhile, prior AGID benchmarks~\cite{zhu2023gendet,sha2023fake} suffer from limited real-image diversity, narrow prompt coverage, outdated model inclusion, and closed access, impeding rigorous evaluation and progress.

To address these limitations, we introduce the \textbf{Visual Counter Turing Test (VCT\textsuperscript{2})}, a large-scale benchmark dataset for zero-shot AGID evaluation. VCT\textsuperscript{2} contains approximately 166,000 images, including 26,000 real prompt-image pairs and 140,000 synthetic images produced by six state of the art T2I models: Stable Diffusion 2.1, SDXL, SD3 Medium, SD3.5 Large, DALL·E 3, and Midjourney 6, spanning both open-source and proprietary systems. The prompts in VCT\textsuperscript{2} are drawn from two semantically distinct sources to capture both structured and open-ended language. The COCO\textsubscript{AI} subset uses object-centric captions from MS COCO~\cite{lin2014microsoft}, a staple in vision-language research. The Twitter\textsubscript{AI} subset comprises narrative-style tweets authored by The New York Times (\texttt{@nytimes}), providing real-world, journalistic prompts rich in nuance and context. This diversity allows us to evaluate AGID methods across a wide range of generation styles and domains. 

To enable more nuanced evaluation beyond binary classification, we introduce the \textbf{Visual AI Index (V\textsubscript{AI})}. This model-agnostic, interpretable metric quantifies the perceptual realism of an image based solely on its visual content. V\textsubscript{AI} produces a scalar score derived from twelve handcrafted, low-level image features, including texture complexity, frequency-domain statistics, Haralick features, and image sharpness. These features have been selected based on their empirically observed alignment with human judgments of realism \cite{wang2004image,haralick2007textural,canny2009computational,corvi2023intriguing}. Correlation analysis further supports the utility of V\textsubscript{AI} as a proxy for detection difficulty: we observe a moderate inverse relationship between V\textsubscript{AI} scores and AGID detection accuracy across generative models (Pearson $\rho = -0.503$ on Twitter\textsubscript{AI} and $\rho = -0.532$ on COCO\textsubscript{AI}), indicating that more visually realistic images tend to be harder to detect. Our realism scores offer a prompt and model-agnostic lens into the perceptual quality of generated images.

We evaluate \textbf{17 AGID} methods under a standardized zero-shot setting, using publicly available implementations and default model checkpoints. Our goal is to assess how well these methods generalize across a variety of text-to-image models, including open-source systems like Stable Diffusion 2.1, SDXL, SD3 Medium, and SD3.5 Large, as well as proprietary models such as DALL·E 3 and Midjourney 6, and across two domains: structured and high quality MS COCO captions and images and narrative-style tweets and real world images from The New York Times.
Experimental results (Section~\ref{sec:results}) reveal model generalization gaps by noticeable detection performance degradation, with average detection accuracy of 58\% on COCO\textsubscript{AI} and 58.34\% on Twitter\textsubscript{AI}. We observe lower detection accuracy on COCO\textsubscript{AI} compared to Twitter\textsubscript{AI}. This is likely because COCO prompts produce images that are more photo-realistic and visually similar to real photos. In contrast, Twitter\textsubscript{AI} generations often include creative or unusual visual patterns, leading to more detectable differences. Notably, DALL·E 3 and SD3.5 consistently yield the lowest detection accuracy across both domains. To summarize, our main contributions are:

(i) We introduce the Visual Counter Turing Test (VCT\textsuperscript{2}) benchmark to evaluate the generalization capabilities of AI-generated image detection methods across diverse prompt styles and real image sources, including MS COCO and Twitter, as well as six state-of-the-art synthetic image generators. 

(ii) We define the VisualAI Index (V\textsubscript{AI}), a scalar metric to quantify perceptual realism based on twelve interpretable low-level visual features.


\section{Recent Advances in AI-Generated Image Detection Techniques}
\label{sec:sota}

AI-generated image detection (AGID) is becoming increasingly vital as synthetic content continues to grow in both photorealism and scale. Detection methods vary widely in their design assumptions, feature representations, and robustness to distribution shifts, such as changes in generative models, prompt styles, or real image characteristics that diverge from photographic norms.

To facilitate systematic evaluation, we categorize AGID approaches into three broad groups:

\begin{enumerate}
    \item[(i)] Generation Artifact-Based Methods: These methods target low-level signals introduced during the image synthesis process, such as upsampling artifacts, denoising residuals, or color inconsistencies. While often computationally efficient, they tend to be fragile under post-processing or model variation.

    \item[(ii)] Feature Representation-Based Methods: These approaches rely on high-level semantic or perceptual features extracted using convolutional neural networks (CNNs), Vision Transformers (ViTs), or CLIP-style encoders, computational models that have demonstrated strong performance across a wide range of vision tasks \cite{imanpour2021memory,bagheri2023automated,rajeoni2024vascular,cai2024einet,zhao2025dpseg}. Such methods generally exhibit robust cross-domain generalization capabilities; however, they may struggle to capture subtle, fine-grained artifacts often present in generative content.

    \item[(iii)] Hybrid Methods: These approaches combine both low-level artifact features (e.g., frequency, texture, or pixel-level traces) and high-level semantic representations (e.g., CNN, ViT, or CLIP embeddings).
They often employ contrastive learning, multi-modal embeddings, or text–image alignment to integrate these complementary cues and enhance robustness under distributional shifts between real and synthetic data
\end{enumerate}
\vspace{-2mm}
Figure~\ref{fig:detection_overview} illustrates this taxonomy. We evaluate 17 publicly available AGID models spanning all three categories, selected based on their methodological diversity, recent relevance, and open-source availability. Each method is tested in a standardized zero-shot setting using its default checkpoint, without any fine-tuning on the VCT\textsuperscript{2} benchmark. This taxonomy provides a reference framework for interpreting detection trends discussed in Section~\ref{sec:results}, with further implementation details outlined in Appendix~\ref{agid_techniques}.

\begin{figure}[!htb]
  \centering
  \resizebox{0.48\textwidth}{!}{%
    \begin{forest}
      forked edges,
      for tree={
        grow=east,
        reversed=true,
        anchor=base west,
        parent anchor=east,
        child anchor=west,
        base=center,
        font=\Large,
        rectangle,
        draw=hidden-draw,
        rounded corners,
        align=center,
        text centered,
        minimum width=5em,
        edge+={darkgray, line width=1pt},
        s sep=3pt,
        inner xsep=2pt,
        inner ysep=3pt,
        line width=0.8pt,
        ver/.style={rotate=90, child anchor=north, parent anchor=south, anchor=center},
      },
      where level=1{text width=15em,font=\Large,}{},
      where level=2{text width=14em,font=\Large,}{},
      where level=3{minimum width=10em,font=\Large,}{},
      where level=4{text width=26em,font=\Large,}{},
      where level=5{text width=28em,font=\Large,}{},
        [
          AI Generated Image Detection, ver,text width=20em, for tree={fill=gray!50} 
          [
            Generation Artifact-Based Detection, text width=27em, for tree={fill=cyan!30}
            [
              NPR \cite{tan2023rethinking}, text width=28em, for tree={fill=cyan!30}  
            ]
            [
              DM Image Detection \cite{corvi2023detection}, text width=28em, for tree={fill=cyan!30}  
            ]
            [
              Fake Image Detection \cite{doloriel2024frequency}, text width=28em, for tree={fill=cyan!30}  
            ]  
            [
              DRCT \cite{chendrct}, text width=28em, for tree={fill=cyan!30}  
            ]
          ]
          [
            Feature Representation-Based Detection, text width=27em, for tree={fill=magenta!30} 
            [
              CNNDetection \cite{wang2020cnn}, text width=28em, for tree={fill=magenta!30}  
            ]
            [
              GAN Image Detection \cite{mandelli2022detecting}, text width=28em, for tree={fill=magenta!30}  
            ]
            [
              DIRE \cite{wang2023dire}, text width=28em, for tree={fill=magenta!30}  
            ]
            [
              LASTED \cite{lasted}, text width=28em, for tree={fill=magenta!30}  
            ]
            [
              De-Fake \cite{sha2023fake}, text width=28em, for tree={fill=magenta!30}  
            ]
            [
              Deep Fake Detection \cite{aghasanli2023interpretable}, text width=28em, for tree={fill=magenta!30}  
            ]
            [
              SSP \cite{chen2024single}, text width=28em, for tree={fill=magenta!30}  
            ]
            [
               RINE \cite{koutlis2024leveraging}, text width=28em, for tree={fill=magenta!30}
            ]
            [
               OCC-CLIP \cite{liumodel}, text width=28em, for tree={fill=magenta!30}
            ]
            [
               Universal Fake Detect \cite{ojha2023fakedetect}, text width=28em, for tree={fill=magenta!30}
            ]
            [
               C2P-CLIP \cite{tan2024c2p}, text width=28em, for tree={fill=magenta!30}
            ]
          ]
          [
            Hybrid Techniques, text width=27em, for tree={fill=orange!30}
            [
               AIDE \cite{yan2024sanity}, text width=28em, for tree={fill=orange!30}  
            ]
            [
               FatFormer \cite{liu2024forgeryaware}, text width=28em, for tree={fill=orange!30}
            ]
          ]
        ]
    \end{forest}
  }
  \vspace{-7mm}
       \caption{The taxonomy of AI-generated image detection techniques, categorized into three main groups: Generation Artifact-Based Detection, Feature Representation-Based Detection, and Hybrid Techniques.}
\end{figure}

  \label{fig:detection_overview}

\section{The Visual Counter Turing Test (VCT\textsuperscript{2}) Benchmark Dataset}
\label{sec:dataset}

We introduce the Visual Counter Turing Test (VCT\textsuperscript{2}), a large-scale benchmark designed to evaluate AI-generated image detection (AGID) techniques. VCT\textsuperscript{2} includes the following:
\begin{itemize}
    \item Approximately 26,000 real image–prompt pairs, combining our curated Twitter dataset and the benchmark MS COCO dataset;
    \item Approximately 140,000 synthetic images, generated using six state-of-the-art text-to-image models; open-source models: Stable Diffusion 2.1, SDXL, SD3 Medium, SD3.5 Large; and two proprietary models: DALL·E 3, Midjourney 6.
\end{itemize}
\begin{itemize}
    \item In total, around 166,000 images derived from 26,000 unique prompts.
\end{itemize}

 This scale provides a balance between structured, caption-based content and naturalistic, real-world prompts, positioning VCT\textsuperscript{2} among the most comprehensive AGID datasets to date. All real images and prompts are organized into two structured subsets: Twitter\textsubscript{AI}\footnote{\url{https://huggingface.co/datasets/NasrinImp/Twitter_AI}} 
and COCO\textsubscript{AI}\footnote{\url{https://huggingface.co/datasets/NasrinImp/COCO_AI}}, which are publicly released.

\subsection{Prompt Sources and Coverage}

To ensure diversity in both semantic content and visual generation styles, we curated prompts from two distinct and complementary sources:

\begin{itemize}
    \item $\sim$10,000 benchmark prompts from the MS COCO dataset \cite{lin2014microsoft}, focused on object-centric and everyday scenes;
    \item $\sim$16,000 real-world prompts from the \texttt{@nytimes} Twitter account 2011--2023. To assess topical diversity, we identified ten topics and associated keywords, and then assigned each tweet to its most probable topic. Table~\ref{tab:nyt_topic_clusters} shows the ten dominant clusters. These clusters reflect both editorial depth and real-world content breadth. Their presence enhances the semantic realism of our benchmark and supports rigorous AGID evaluation across multiple domains.
\end{itemize}

\begin{table*}[t]
\centering
\small
\caption{Topic Clusters in the NYT Twitter Subset.}
\vspace{-2mm}
\label{tab:nyt_topic_clusters}
\begin{tabular}{|p{5.7cm}|c|p{7.3cm}|}
\hline
\rowcolor{gray!15}
\textbf{Topic Cluster} & \textbf{Tweet Count} & \textbf{Top Keywords} \\
\hline
Daily Briefings and News Summaries     & 1129 & \textit{know, need, day, morning, briefing, evening} \\
New York City and Culture     & 1961 & \textit{new, york, city, times, books, critics} \\
Art, Movies, and Obituaries& 631 & \textit{photo, review, obituary, art, movie, critic} \\
Health, COVID-19, and Breaking News   & 1436 & \textit{coronavirus, health, opinion, news, breaking, people} \\
Opinion Pieces and Societal Reflections& 914 & \textit{nytopinion, life, young, america, death, ebola} \\
Travel and International Destinations&605&\textit{hours, italy, florida, china, japan, park}\\
World Events and Sports & 903  & \textit{world, cup, photos, team, war, country}\\
Lifestyle and City Aesthetics & 1164 & \textit{like, looks, look, city, love, idea}\\
Time, Life Stories, and Incarceration &1200&\textit{years, life, ago, prison, time, close}\\
Home, Food, and Leisure & 965 & \textit{make, recipes, summer, home, simple, best}\\
miscellaneous &5001 & \textit{--}\\
\hline
\end{tabular}
\end{table*}

\subsection{Real Twitter Prompt-Image Dataset Collection and Processing}
\label{appendix:twitter}

To construct a diverse and reliable dataset of real Twitter images, we employed an automated data collection pipeline using Python and Selenium. We focused on tweets from \texttt{@nytimes} (The New York Times) due to its editorial credibility, rigorous fact-checking, and diverse topical coverage. 

\textbf{Data Collection.} Our pipeline sampled tweets spanning a 12-year period (2011-2023), retaining only those with attached media. The goal was to align real images with captions that could feasibly be used to generate synthetic counterparts.

\textbf{Definition of Real Images.} We define “real” images as those not generated by AI. This includes natural photographs as well as editorial media such as UI screenshots, infographics, and photojournalistic illustrations, provided they are not produced using generative models. This definition reflects the ambiguity present in real-world detection scenarios, where non-photographic content may still be authentic.

\textbf{Data Filtration and Preprocessing.} To ensure quality and consistency, we applied several filtering steps: (i) removal of duplicate tweets and media; (ii) exclusion of irrelevant content such as word games or puzzles; and (iii) filtering of non-English tweets. Additionally, preprocessing involved removal of hashtags and URLs, and retention of only alphanumeric characters to facilitate downstream analysis and clustering.

\subsection{Benchmark Contributions}

VCT\textsuperscript{2} offers several key advantages over existing benchmarks:

\begin{itemize}
    \item \textbf{Scale}. VCT\textsuperscript{2} contains 166,000 images generated from 26,000 unique prompts. 
    
    \item \textbf{Model diversity.} VCT\textsuperscript{2} includes six cutting-edge generative models supporting broader evaluation across current-generation image generators.
    
    \item \textbf{Prompt realism.} VCT\textsuperscript{2} uniquely combines benchmark-style prompts from MS COCO with naturalistic, real-world prompts curated from a 12-year archive of \texttt{@nytimes} tweets, capturing diverse linguistic styles and topics.
    
    \item \textbf{Mixed-media realism.} The real image subset includes ambiguous formats such as infographics, UI screenshots, and editorial photos, reflecting the heterogeneous content encountered in real-world detection scenarios.
    
    \item \textbf{Public accessibility.} All prompts, real and synthetic images, and evaluation scripts for 17 AGID baselines are publicly released to facilitate reproducibility and comparative benchmarking.
\end{itemize}

\noindent \textit{To our knowledge, VCT\textsuperscript{2} is the first large-scale AGID benchmark to pair real-world journalistic prompts with diverse state-of-the-art text-to-image models, providing a robust and publicly available testbed for evaluating both detection performance and perceptual realism across different prompt domains and generative model types.}

\section{Evaluation and Results}
\label{sec:results}

\begin{table*}[!b]
    \centering
    \caption{Overall accuracy (Acc), recall (R), and precision (P) across COCO\textsubscript{AI} synthetic datasets generated from MS COCO prompts. All values are in \%. Color-coded: Green (\(\geq90\%\)), Yellow-Green (80--89\%), Yellow (70--79\%), Orange (60--69\%), red ($<$60\%).
} 
 \vspace{-0.3cm} 
    \small
    \renewcommand{\arraystretch}{0.8}
    \setlength{\extrarowheight}{1.2pt}
    \adjustbox{width=\textwidth}{%
    \begin{tabular}{p{6.5cm}p{.5cm}p{.5cm}p{.5cm}p{.5cm}p{.5cm}p{.5cm}p{.5cm}p{.5cm}p{.5cm}p{.5cm}p{.5cm}p{.5cm}p{.5cm}p{.5cm}p{.5cm}p{.5cm}p{.5cm}p{.5cm}}
    
        \toprule
        \multirow{2}{*}{ \small Method} & \multicolumn{3}{c}{ \small SD2.1} &\multicolumn{3}{c}{ \small SDXL} & \multicolumn{3}{c}{ \small SD3 Medium} & \multicolumn{3}{c}{ \small SD3.5 Large} & \multicolumn{3}{c}{ \small DALL.E 3} & \multicolumn{3}{c}{ \small Midjourney 6} \\  
        \cmidrule{2-19}
        & \small Acc & \small R & \small P & \small Acc & \small R & \small P & \small Acc & \small R & \small P & \small Acc & \small R & \small P & \small Acc & \small R & \small P & \small Acc & \small R & \small P  \\  
        \toprule
         \small CNNDetection \cite{wang2020cnn} & \cellcolor{red}49.94 &\cellcolor{red}0.03 &\cellcolor{orange}65.11 & \cellcolor{red}49.96 &\cellcolor{red}0.07 &\cellcolor{yellow}77.52 & \cellcolor{red}49.93 &\cellcolor{red}0.01 & \cellcolor{lime}81.16 & \cellcolor{red}49.99 &\cellcolor{red} 0.14 & \cellcolor{red}33.04 & \cellcolor{red}49.93 &\cellcolor{red}0.00 & \cellcolor{red}35.13 &\cellcolor{red}49.95 & \cellcolor{red}0.05 & \cellcolor{orange}63.15 \\
        \small NPR \cite{tan2023rethinking}  & \cellcolor{red}26.76 & \cellcolor{red}1.89  & \cellcolor{red}34.26 & \cellcolor{red}26.68 & \cellcolor{red}1.73 & \cellcolor{red}33.15 & \cellcolor{red}27.96 & \cellcolor{red}4.29 & \cellcolor{red}34.41 & \cellcolor{yellow}70.32 & \cellcolor{red}48.37 &\cellcolor{yellow} 79.44 & \cellcolor{red}25.81 & \cellcolor{red}0.00 & \cellcolor{red}41.13 & \cellcolor{red}25.81 & \cellcolor{red}0.00 & \cellcolor{red}48.13 \\
        \small DM Image Detection \cite{corvi2023detection}  & \cellcolor{lime}83.92 & \cellcolor{orange}67.92 & \cellcolor{green}99.40 & \cellcolor{orange}69.96 & \cellcolor{red}40.00 & \cellcolor{green}98.91 & \cellcolor{orange}63.58 & \cellcolor{red}27.23 & \cellcolor{green}98.04 &\cellcolor{red} 38.58 & \cellcolor{red} 32.06 & \cellcolor{red} 0.07 & \cellcolor{red}49.96 & \cellcolor{red}0.00 & \cellcolor{red}40.00 & \cellcolor{red}51.73 & \cellcolor{red}3.52 & \cellcolor{lime}87.04 \\  
        
        \small Fake Image Detection \cite{doloriel2024frequency}  & \cellcolor{red}49.84 & \cellcolor{red}0.49 & \cellcolor{orange}63.58 & \cellcolor{red}49.83 & \cellcolor{red}0.48 & \cellcolor{orange}66.68 & \cellcolor{red}50.02 & \cellcolor{red}0.86 & \cellcolor{orange}66.91 & 
        \cellcolor{red}{48.24}& \cellcolor{red}{0.11}& \cellcolor{orange}{62.57}& \cellcolor{orange}49.59 & \cellcolor{red}0.00 & \cellcolor{red}34.90 & \cellcolor{red}49.79 & \cellcolor{red}0.40 & \cellcolor{orange}62.89 \\  
        \small DIRE \cite{wang2023dire}  & \cellcolor{red}47.08 & \cellcolor{green}93.40 & \cellcolor{red}37.66 & \cellcolor{red}49.67 & \cellcolor{green}98.57 & \cellcolor{red}47.07 & \cellcolor{red}48.59 & \cellcolor{green}96.40 & \cellcolor{red}38.88 &\cellcolor{red} 50.63 & \cellcolor{green}99.23 &\cellcolor{red} 58.68 & \cellcolor{red}48.89 & \cellcolor{green}97.01 & \cellcolor{red}43.25 & \cellcolor{red}50.04 & \cellcolor{green}99.31 & \cellcolor{red}52.74 \\ 
         \small LASTED \cite{lasted} & \cellcolor{red}54.00 & \cellcolor{red}8.67 & \cellcolor{red}56.62 & \cellcolor{orange}61.13 & \cellcolor{red}9.86 & \cellcolor{orange}61.20 & \cellcolor{red}51.87 & \cellcolor{red}9.61 & \cellcolor{red}57.67 & \cellcolor{red}55.21 & \cellcolor{red}10.11 &\cellcolor{red} 57.35 & \cellcolor{orange}66.18 & \cellcolor{red}44.85 & \cellcolor{yellow}76.21 & \cellcolor{orange}68.21 & \cellcolor{red}14.37 & \cellcolor{orange}63.14 \\ 
         \small GAN Image Detection \cite{mandelli2022detecting} & \cellcolor{red}51.87 & \cellcolor{lime}82.93  & \cellcolor{red}51.16 & \cellcolor{red}56.35 & \cellcolor{green}91.75 & \cellcolor{red}53.72 & \cellcolor{red}58.26 & \cellcolor{green}95.35 & \cellcolor{red}54.74 & \cellcolor{red}45.61 & \cellcolor{yellow}79.08 &\cellcolor{red} 47.37 & \cellcolor{red}48.10 & \cellcolor{yellow}74.93 & \cellcolor{red}48.77 & \cellcolor{red}57.15 & \cellcolor{green}93.42 & \cellcolor{red}54.14 \\  
         \small AIDE \cite{yan2024sanity} & \cellcolor{orange}60.30 & \cellcolor{red}20.98  & \cellcolor{green}93.77 & \cellcolor{orange}64.34 & \cellcolor{red}28.91 & \cellcolor{green}96.75 & \cellcolor{red}57.11 & \cellcolor{red}14.45 & \cellcolor{green}94.28 &\cellcolor{red}50.83 &\cellcolor{red}5.01&\cellcolor{red}52.31 & \cellcolor{orange!60}50.00 & \cellcolor{red}0.02 & \cellcolor{orange}61.23 & \cellcolor{yellow}76.01 & \cellcolor{red}52.25 & \cellcolor{green}96.92 \\  
        \small SSP \cite{chen2024single} & \cellcolor{red}50.15 & \cellcolor{green}99.63 & \cellcolor{red}50.07 & \cellcolor{red}49.95 & \cellcolor{green}99.63 & \cellcolor{red}49.97 & \cellcolor{red}50.34 & \cellcolor{green}99.63 & \cellcolor{red}50.17 &\cellcolor{red}50.30&\cellcolor{green}99.48&\cellcolor{red}50.29& \cellcolor{red}49.91 & \cellcolor{green}99.63 & \cellcolor{red}49.95 & \cellcolor{red}49.95 & \cellcolor{green}99.63 & \cellcolor{red}49.97 \\  
        \small FatFormer \cite{liu2024forgeryaware} & \cellcolor{red}50.00 & \cellcolor{red}0.00 & \cellcolor{red}0.00 & \cellcolor{red} 50.00& \cellcolor{red} 0.00& \cellcolor{red}0.00& \cellcolor{red}50.00 & \cellcolor{red}0.01 & \cellcolor{green}100 &\cellcolor{red}50.28&\cellcolor{red}0.00&\cellcolor{red}0.00& \cellcolor{red}48.01& \cellcolor{red}0.00 & \cellcolor{red}0.00 & \cellcolor{red}48.01 & \cellcolor{red}0.00 & \cellcolor{red}0.00 \\  
        \small DRCT (ConvB) \cite{chendrct} & \cellcolor{green}98.76 & \cellcolor{green}99.61 & \cellcolor{green}97.94 & \cellcolor{green}96.83 & \cellcolor{green}95.75 & \cellcolor{green}97.86 & \cellcolor{lime}80.72 & \cellcolor{orange}63.54 & \cellcolor{green}96.81 &\cellcolor{yellow}78.58 &\cellcolor{red}59.05&\cellcolor{green}96.51& \cellcolor{red}49.99 & \cellcolor{red}2.08 & \cellcolor{red}49.76 & \cellcolor{orange}67.48 & \cellcolor{red}37.06 & \cellcolor{green}94.65 \\  
        \small DRCT (UnivFD) \cite{chendrct} & \cellcolor{lime}88.57 & \cellcolor{green}96.98 & \cellcolor{lime}83.02 & \cellcolor{lime}89.45 & \cellcolor{green}98.73 & \cellcolor{lime}83.27 & \cellcolor{lime}84.90 & \cellcolor{lime}89.64 & \cellcolor{lime}81.88 &\cellcolor{lime}83.09&\cellcolor{lime}84.09&\cellcolor{lime}82.29& \cellcolor{yellow}79.98 & \cellcolor{yellow}79.80 & \cellcolor{lime}80.09 & \cellcolor{lime}89.64 & \cellcolor{green}99.12 & \cellcolor{lime}83.32 \\  
        \small RINE \cite{koutlis2024leveraging} & \cellcolor{yellow}74.43 & \cellcolor{red}49.63 & \cellcolor{green}98.49 & \cellcolor{red}56.47 & \cellcolor{red}13.71 & \cellcolor{green}94.76 & \cellcolor{orange}61.99 & \cellcolor{red}24.75 & \cellcolor{green}97.03 &\cellcolor{red}55.34&\cellcolor{red}27.72&\cellcolor{green}95.34& \cellcolor{red}50.05 & \cellcolor{red}0.87 & \cellcolor{red}53.37 & \cellcolor{orange}63.13 & \cellcolor{red}27.02 & \cellcolor{green}97.27 \\  
        \small OCC-CLIP \cite{liumodel} & \cellcolor{red}51.49 & \cellcolor{green}92.28 & \cellcolor{red}50.82 & \cellcolor{red}47.11 & \cellcolor{red}14.95 & \cellcolor{red}41.91 & \cellcolor{red}50.60 & \cellcolor{orange}66.03 & \cellcolor{red}50.46 &
        \cellcolor{red}49.08&
        \cellcolor{red}50.67&
        \cellcolor{orange}67.63& \cellcolor{yellow}78.82 & \cellcolor{red}50.28 & \cellcolor{red}55.04 & \cellcolor{yellow}75.04 & \cellcolor{red}53.60 &
        \cellcolor{red}50.03 \\  
        \small De-Fake \cite{sha2023fake} & \cellcolor{green}92.37 & \cellcolor{green}97.90 & \cellcolor{lime}88.15 & \cellcolor{green}91.23 & \cellcolor{green}95.62 & \cellcolor{lime}87.90 & \cellcolor{green}91.30 & \cellcolor{green}95.76 & \cellcolor{lime}87.92 &\cellcolor{red}52.57&\cellcolor{lime}86.05&\cellcolor{red}5.11& \cellcolor{green}90.58 & \cellcolor{green}94.31 & \cellcolor{lime}87.76 & \cellcolor{lime}86.22 & \cellcolor{lime}85.59 & \cellcolor{lime}86.68 \\  
        \small Deep Fake Detection \cite{aghasanli2023interpretable} & \cellcolor{red}49.49 & \cellcolor{red}49.49 & \cellcolor{red}49.03 & \cellcolor{red}51.43 & \cellcolor{red}51.43 & \cellcolor{red}49.65 & \cellcolor{red}49.85 & \cellcolor{red}49.85 & \cellcolor{red}49.97 &\cellcolor{red} 50.66& \cellcolor{red}50.66&\cellcolor{red}52.19& \cellcolor{red}52.73 & \cellcolor{red}52.73 & \cellcolor{red}53.02 & \cellcolor{red}52.87 & \cellcolor{red}52.87 & \cellcolor{red}54.09 \\ 
        \small Universal Fake Detect \cite{ojha2023fakedetect} &\cellcolor{yellow} 74.42 & \cellcolor{yellow}77.15 & \cellcolor{yellow}73.15 & \cellcolor{orange}69.18 &\cellcolor{orange} 65.84 &\cellcolor{yellow} 70.56 &\cellcolor{yellow} 70.11 & \cellcolor{orange}68.79 & \cellcolor{yellow}70.66 & \cellcolor{red}57.46 &\cellcolor{orange}60.10  & \cellcolor{red}57.09 & \cellcolor{red}50.00 & \cellcolor{green}99.99 & \cellcolor{red}50.00 & \cellcolor{red}53.23 & \cellcolor{yellow}76.28 &\cellcolor{red} 52.21 \\
        \small C2P-CLIP \cite{tan2024c2p} & \cellcolor{red}53.38 & \cellcolor{red}7.53 & \cellcolor{green}90.73 &
        \cellcolor{red} 53.69& \cellcolor{red} 8.15 & \cellcolor{green} 91.30 &
        \cellcolor{red} 55.50 & \cellcolor{red}11.77 &\cellcolor{green}93.87&\cellcolor{red}52.13&\cellcolor{red}4.40&\cellcolor{green}97.01& 
        \cellcolor{red}49.76 &
        \cellcolor{red}0.29 & \cellcolor{red}27.36 & \cellcolor{red}50.31 & \cellcolor{red}1.40 & \cellcolor{orange}64.52 \\
        \bottomrule
    \end{tabular}
}

     \label{tab:res_coco} 
     \vspace{0.4cm}
\end{table*}
\begin{table*}

    \centering
     \caption{Overall accuracy (Acc), recall (R), and precision (P) across Twitter\textsubscript{AI} synthetic datasets generated from Twitter prompts. Midjourney 6 is not included as it blocks image generation for most Twitter prompts. All values are in \%. Color-coded: Green (\(\geq90\%\)), Yellow-Green (80--89\%), Yellow (70--79\%), Orange (60--69\%), red ($<$60\%).
} 
\vspace{-2mm}
    \small
    \renewcommand{\arraystretch}{0.8}
    \setlength{\extrarowheight}{1.2pt}
    \adjustbox{width=\textwidth}{%
    \begin{tabular}{p{6.5cm}p{.5cm}p{.5cm}p{.5cm}p{.5cm}p{.5cm}p{.5cm}p{.5cm}p{.5cm}p{.5cm}p{.5cm}p{.5cm}p{.5cm}p{.5cm}p{.5cm}p{.5cm}p{.5cm}p{.5cm}p{.5cm}}
        \toprule
        \multirow{2}{*}{ \small Method} & \multicolumn{3}{c}{ \small SD2.1} & \multicolumn{3}{c}{ \small SDXL} & \multicolumn{3}{c}{ \small SD3 Medium} & \multicolumn{3}{c}{ \small SD3.5 Large} & \multicolumn{3}{c}{ \small DALL.E 3} \\  
        \cmidrule{2-16}
        & \small Acc & \small R & \small P & \small Acc & \small R & \small P & \small Acc & \small R & \small P & \small Acc & \small R & \small P & \small Acc & \small R & \small P \\  
        \toprule
        \small CNNDetection \cite{wang2020cnn} & \cellcolor{red}50.00 & \cellcolor{red}0.06 & \cellcolor{red}52.21 & \cellcolor{red}49.98 & \cellcolor{red}0.03 & \cellcolor{red}59.98 & \cellcolor{red}50.19 & \cellcolor{red}0.44 & \cellcolor{yellow}74.35 & \cellcolor{red}50.34 &\cellcolor{red} 0.76 &\cellcolor{yellow} 76.04 & \cellcolor{red}49.97 & \cellcolor{red}0.01 & \cellcolor{red}34.59  \\ 
        \small NPR \cite{tan2023rethinking} & \cellcolor{red}50.23 & \cellcolor{red}2.22 & \cellcolor{red}50.89 & \cellcolor{red}50.46 & \cellcolor{red}2.68 & \cellcolor{orange}60.58 & \cellcolor{red}51.45 & \cellcolor{red}4.66 & \cellcolor{orange}68.26 &\cellcolor{red} 52.12 &\cellcolor{red} 7.81 &\cellcolor{orange} 67.44 & \cellcolor{red}49.12 & \cellcolor{red}0.00 & \cellcolor{red}42.20  \\ 
        \small DM Image Detection \cite{corvi2023detection} & \cellcolor{lime}88.31 & \cellcolor{yellow}77.57 & \cellcolor{green}97.82 & \cellcolor{yellow}73.82 & \cellcolor{red}48.58 & \cellcolor{green}93.74 & \cellcolor{orange}65.15 & \cellcolor{red}31.24 & \cellcolor{green}90.34 &\cellcolor{yellow} 63.56 &\cellcolor{red} \cellcolor{red}28.19 &\cellcolor{lime} 89.66 & \cellcolor{red}49.53 & \cellcolor{red}0.00 & \cellcolor{red}33.34 \\ 
        \small Fake Image Detection \cite{doloriel2024frequency} & \cellcolor{red}49.86 & \cellcolor{red}0.53 & \cellcolor{red}56.33 & \cellcolor{red}49.88 & \cellcolor{red}0.58 & \cellcolor{orange}60.83 & \cellcolor{red}50.35 & \cellcolor{red}1.51 & \cellcolor{orange}66.15 &
        \cellcolor{red}48.57&
        \cellcolor{red}1.12&
        \cellcolor{orange}62.13& 
        \cellcolor{red}49.59 & \cellcolor{red}0.01 & \cellcolor{red}33.11  \\  
        \small DIRE \cite{wang2023dire} & \cellcolor{red}43.90 & \cellcolor{lime}86.95 & \cellcolor{red}36.20 & \cellcolor{red}48.57 & \cellcolor{green}96.29 & \cellcolor{red}46.29 & \cellcolor{red}48.49 & \cellcolor{green}96.13 & \cellcolor{red}38.48 & \cellcolor{red}49.41 &\cellcolor{green} 98.01 &\cellcolor{red} 45.11 & \cellcolor{red}46.33 & \cellcolor{green}91.81 & \cellcolor{red}36.19  \\
        \small LASTED \cite{lasted} & \cellcolor{yellow}77.60 & \cellcolor{red}1.93 & \cellcolor{red}59.60 & \cellcolor{lime}83.60 & \cellcolor{red}2.75 & \cellcolor{orange}66.04 & \cellcolor{lime}83.24 & \cellcolor{red}2.75 & \cellcolor{orange}61.52 & \cellcolor{lime}82.71 & \cellcolor{red}2.59 & \cellcolor{orange}61.90 & \cellcolor{yellow}78.77 & \cellcolor{red}25.57 & \cellcolor{yellow}76.81  \\
        \small GAN Image Detection \cite{mandelli2022detecting} & \cellcolor{red}53.26 & \cellcolor{yellow}77.37 & \cellcolor{red}52.25 & \cellcolor{red}55.84 & \cellcolor{lime}82.36 & \cellcolor{red}53.86 & \cellcolor{orange}60.01 & \cellcolor{green}91.04 & \cellcolor{red}56.21 &\cellcolor{red} 54.78 &\cellcolor{lime} 80.60 & \cellcolor{red}53.19 & \cellcolor{red}53.99 & \cellcolor{yellow}79.44 & \cellcolor{red}52.68  \\  
        \small AIDE \cite{yan2024sanity} & \cellcolor{red}55.69 & \cellcolor{red}11.81 & \cellcolor{lime}81.98 & \cellcolor{orange}60.43 & \cellcolor{red}21.29 & \cellcolor{lime}89.61 & \cellcolor{red}56.49 & \cellcolor{red}13.41 & \cellcolor{lime}87.40 &\cellcolor{red}56.57&\cellcolor{red}6.16\cellcolor{red}&\cellcolor{red}58.42&        \cellcolor{red}49.93 & \cellcolor{red}0.25 & \cellcolor{red}43.61  \\  
        \small SSP \cite{chen2024single} & \cellcolor{red}49.91 & \cellcolor{green}99.66 & \cellcolor{red}49.95 & \cellcolor{red}50.20 & \cellcolor{green}99.66 & \cellcolor{red}50.10 & \cellcolor{red}50.20 & \cellcolor{green}99.66 & \cellcolor{red}50.10 &\cellcolor{red}54.94&\cellcolor{green}99.33&\cellcolor{red}55.04& \cellcolor{red}50.18 & \cellcolor{green}99.66 & \cellcolor{red}50.10   \\  
        \small FatFormer \cite{liu2024forgeryaware} & \cellcolor{red}50.04& \cellcolor{red}0.08 & \cellcolor{green}100 & \cellcolor{red} 50.04& \cellcolor{red} 0.08& \cellcolor{green}100& \cellcolor{red}50.00 & \cellcolor{red}0.00 & \cellcolor{red}0.00 &\cellcolor{red}55.10&\cellcolor{red}0.02&\cellcolor{green}100& \cellcolor{red}50.02& \cellcolor{red}0.00 &\cellcolor{red}0.00  \\  
        \small DRCT (ConvB) \cite{chendrct} & \cellcolor{green}96.81 & \cellcolor{green}99.77 & \cellcolor{green}94.20 & \cellcolor{green}93.96 & \cellcolor{green}94.05 & \cellcolor{green}93.87 & \cellcolor{yellow}71.79 & \cellcolor{red}49.73 & \cellcolor{lime}89.01 &\cellcolor{yellow}77.87&\cellcolor{red}59.54&\cellcolor{green}90.35& \cellcolor{red}47.31 & \cellcolor{red}0.76 & \cellcolor{red}11.02   \\  
        \small DRCT (UnivFD) \cite{chendrct} & \cellcolor{orange}67.47 & \cellcolor{green}96.73 & \cellcolor{orange}61.02 & \cellcolor{orange}68.32 & \cellcolor{green}98.43 & \cellcolor{orange}61.44 & \cellcolor{orange}64.81 & \cellcolor{green}91.40 & \cellcolor{red}59.67 &\cellcolor{orange}62.94&\cellcolor{lime}85.60&\cellcolor{red}55.68& \cellcolor{red}53.76 & \cellcolor{yellow}69.30 & \cellcolor{red}52.87  \\  
        \small RINE \cite{koutlis2024leveraging} & \cellcolor{yellow}77.07 & \cellcolor{red}55.40 & \cellcolor{green}97.79 & \cellcolor{red}57.86 & \cellcolor{red}16.97 & \cellcolor{green}93.13 & \cellcolor{orange}62.13 & \cellcolor{red}25.50 & \cellcolor{green}95.32 &\cellcolor{orange}66.36&\cellcolor{red}44.37&\cellcolor{green}94.36& \cellcolor{red}49.61 & \cellcolor{red}0.48 & \cellcolor{red}27.64  \\  
        \small OCC-CLIP \cite{liumodel} & \cellcolor{red}46.88 & \cellcolor{yellow}74.11 & \cellcolor{red}47.98 & \cellcolor{red}45.67 & \cellcolor{red}51.17 & \cellcolor{red}46.10 & \cellcolor{red}48.84 & \cellcolor{orange}67.54 & \cellcolor{red}49.16 &
        \cellcolor{red}47.82 &
        \cellcolor{orange}66.16 &
        \cellcolor{red}48.03 & \cellcolor{red}47.75 & \cellcolor{red}45.63 & \cellcolor{red}49.72  \\  
        \small De-Fake \cite{sha2023fake} & \cellcolor{lime}81.13 & \cellcolor{green}91.51 & \cellcolor{yellow}75.78 & \cellcolor{yellow}78.16 & \cellcolor{lime}85.57 & \cellcolor{yellow}74.53 & \cellcolor{yellow}79.39 & \cellcolor{lime}88.03 & \cellcolor{yellow}72.06 &
        \cellcolor{red}40.80 &
        \cellcolor{red}0.00 &
        \cellcolor{red}0.00 & \cellcolor{yellow}79.95 & \cellcolor{lime}89.14 & \cellcolor{yellow}75.29  \\  
        \small Deep Fake Detection \cite{aghasanli2023interpretable} & \cellcolor{red}50.80 & \cellcolor{red}50.80 & \cellcolor{red}51.84 & \cellcolor{red}53.64 & \cellcolor{red}53.64 & \cellcolor{red}56.59 & \cellcolor{red}51.44 & \cellcolor{red}51.44 & \cellcolor{red}51.51 &\cellcolor{red}49.19&\cellcolor{red}49.19&\cellcolor{red}56.38& \cellcolor{red}55.30 & \cellcolor{red}55.30 & \cellcolor{orange}60.34  \\  
        \small Universal Fake Detect \cite{ojha2023fakedetect} & \cellcolor{yellow}72.88 &\cellcolor{yellow} 74.17 &\cellcolor{yellow} 72.31 &\cellcolor{orange} 69.47 &\cellcolor{yellow} 73.91 &\cellcolor{orange} 67.89 &\cellcolor{orange} 68.41 & \cellcolor{yellow}72.58 &\cellcolor{orange} 67.00 &\cellcolor{red}55.38 &\cellcolor{red} 45.60& \cellcolor{red}56.69& \cellcolor{red}50.00 &\cellcolor{green} 99.99 &\cellcolor{red} 50.00  \\ 
        \small C2P-CLIP \cite{tan2024c2p} & \cellcolor{red} 52.21 
        &\cellcolor{red} 6.74  &\cellcolor{lime} 88.76 &\cellcolor{red} 53.28 &\cellcolor{red} 7.98
        &\cellcolor{green}  91.25
        &\cellcolor{red} 47.92 & \cellcolor{red} 0.97 &\cellcolor{red} 26.17 &\cellcolor{red}54.16 &\cellcolor{red}8.47 & \cellcolor{green}98.39& \cellcolor{red} 49.73 &\cellcolor{red} 0.21 &\cellcolor{red} 53.45 \\ 
        \bottomrule
    \end{tabular}
    }    
    \vspace{-0.3cm} 
   
    \label{tab:res_tweeter} 
\end{table*}

We evaluate the VCT\textsuperscript{2} benchmark under zero-shot settings using 17 state-of-the-art AI-generated image detection (AGID) methods. These span artifact-, feature-, and hybrid-based approaches. In the following, we present detection performance, examine cross-domain and cross-model generalization, and analyze detector sensitivity to detector type.

\subsection{Evaluation Protocol}
\vspace{-2mm}
To simulate real-world deployment, we assess all detectors without fine-tuning. Public checkpoints and default hyperparameters are used. Performance is measured separately on COCO\textsubscript{AI} and Twitter\textsubscript{AI} subsets, reporting accuracy, precision, and recall. Results are summarized in Tables~\ref{tab:res_coco} and~\ref{tab:res_tweeter}.

\subsection{Cross-Domain and -Model Trends}
\begin{figure}[h]
\centering
\includegraphics[width=\linewidth]{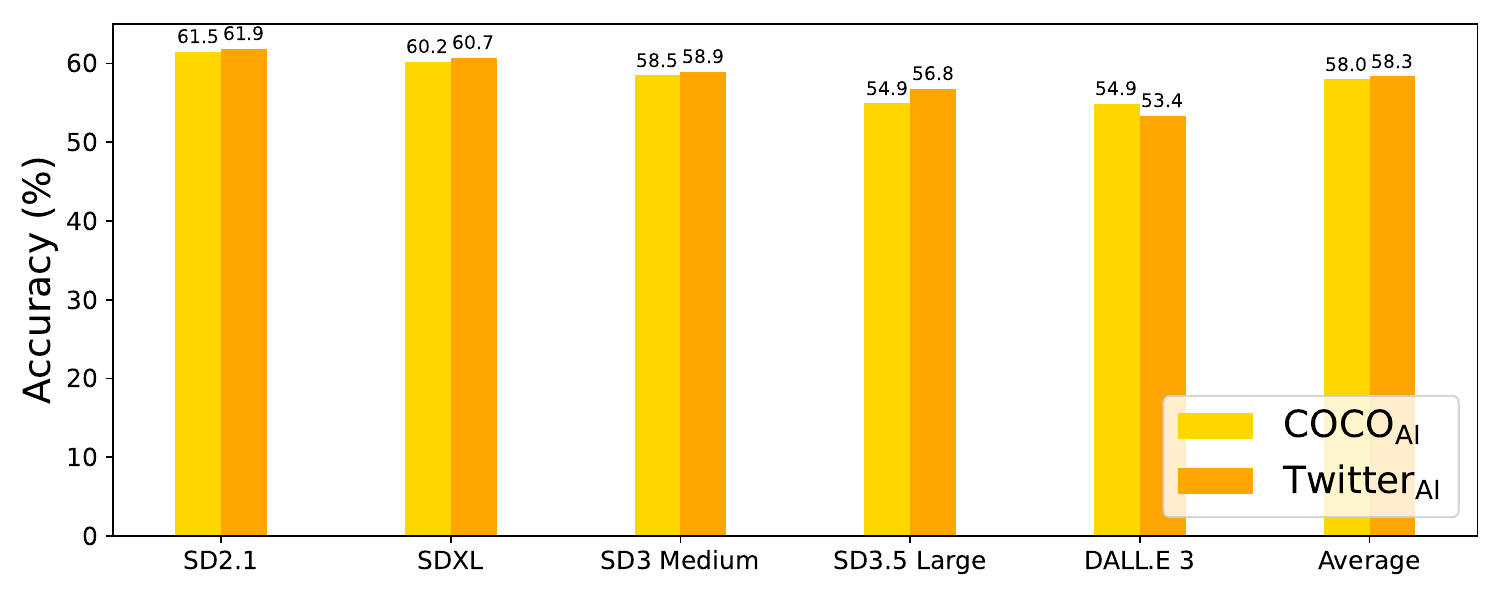}
\vspace{-7mm}

\caption{Average detection accuracy across the COCO\textsubscript{AI} and Twitter\textsubscript{AI} subsets for each generator.}
\vspace{-4mm}

\label{fig:det_gap}
\end{figure}

\vspace{-2mm}
Figure~\ref{fig:det_gap} presents the average detection accuracies per generator across the two domains. Overall, detection performance is low, with accuracy dropping further on COCO\textsubscript{AI} compared to Twitter\textsubscript{AI}.

Detection performance also varies across generators. Images from earlier models like SD2.1 and SDXL remain relatively detectable. In contrast, newer or proprietary models such as SD3.5 Large and DALL·E 3 yield significantly lower detection results, suggesting that existing detectors may be overfitted to older, synthetic image distributions. 

\subsection{Comparative Detector Performance}
We analyze per detector performance in Appendix \ref{app:det}.
The results indicate that there is no one-size-fits-all solution for detecting AI-generated images. Different generative models pose unique challenges, and the performance of each detection method varies based on its ability to identify specific artifacts. De-Fake and DRCT were the most consistent performers, highlighting their robustness across models. The latter collapse on proprietary models due to reliance on low-level artifacts often absent in advanced generators. Conversely, feature-based and contrastive methods benefit from semantic representations, allowing stronger generalization to unseen prompt styles and model outputs. Future research should aim to improve detection for models like SD 3.5 Large, Midjourney 6 and DALL.E 3, where many techniques struggled.

\section{The Visual AI Index (V\textsubscript{AI})}
\label{sec:vai}

We introduce the \textit{Visual AI Index} (V\textsubscript{AI}), an interpretable, prompt-agnostic metric that scores the perceptual realism of images based on low-level visual features. V\textsubscript{AI} provides a continuous score that reflects where an image lies along a spectrum of visual realism.
Many real images in web-scale datasets (e.g., news media, social platforms) are not pristine photographs; they may include screenshots, digital graphics, or compressed visuals. These images often lack sharpness, contrast, or structure. V\textsubscript{AI} quantifies perceptual quality by learning to score realism using a combination of handcrafted visual features, independent of prompts or model-specific information.

\subsection{Feature Design}

V\textsubscript{AI} uses twelve visual features grouped into three categories:

\textbf{(i) Texture and Frequency:} Texture Complexity, Haralick Contrast, Haralick Correlation, Haralick Energy, Frequency Mean, Frequency Standard Deviation.

\textbf{(ii) Sharpness and Structure:} Image Sharpness, Image Smoothness, Image Contrast.

\textbf{(iii) Color and Semantics:} Color Distribution Consistency, Object Coherence, Contextual Relevance.

\textbf{Texture Complexity} quantifies the variety and unpredictability of an image's texture. It is determined by computing the entropy of the normalized Local Binary Pattern (LBP) histogram of the grayscale image using the formula $-\sum_{k=0}^{P-1} \tilde{H}_{LBP}(k) \log_2 (\tilde{H}_{LBP}(k) + \epsilon)$. Here, $\tilde{H}_{LBP}(k)$ represents the normalized histogram value for LBP bin $k$, and $P$ is the total number of bins in the LBP histogram. The small constant $\epsilon$ (in our case, $1 \times 10^{-6}$) is used to avoid taking the logarithm of zero.

Haralick features are texture descriptors computed from the gray-level co-occurrence matrix (GLCM), which encodes the frequency \( G(i,j) \) of pixel intensity pairs \( (i, j) \) occurring at a fixed spatial offset.
We use three common features:

\textbf{Haralick Contrast} is defined as \( \sum_{i,j} (i - j)^2 G(i,j) \), capturing local intensity variation.

\textbf{Haralick Correlation} is computed as \( \sum_{i,j} \frac{(i - \mu_i)(j - \mu_j) G(i,j)}{\sigma_i \sigma_j} \), where \( \mu_i, \mu_j \) and \( \sigma_i, \sigma_j \) are the means and standard deviations of the marginal GLCM distributions. It measures linear dependency between pixel pairs.

\textbf{Haralick Energy} (Angular Second Moment) is given by \( \sum_{i,j} G(i,j)^2 \), reflecting texture uniformity—higher values imply more homogeneous regions.

These values are averaged across multiple angles (e.g., \( 0^\circ \), \( 45^\circ \), \( 90^\circ \), \( 135^\circ \)) to ensure rotation-invariant descriptors.

We extract frequency-domain features using the 2D Fast Fourier Transform (FFT) of the grayscale image \( I \). Let \( \hat{I}(u,v) = \mathrm{FFT2}(I) \) denote the Fourier-transformed image, and let \( M(u,v) = |\hat{I}(u,v)| \) be the magnitude spectrum.

\textbf{Frequency Mean} is defined as \( \text{FreqMean} = \frac{1}{HW} \sum_{u=1}^{H} \sum_{v=1}^{W} M(u,v) \), where \( H \times W \) is the image resolution.

\textbf{Frequency Standard Deviation} is given by \( \text{FreqStd} = \sqrt{ \frac{1}{HW} \sum_{u=1}^{H} \sum_{v=1}^{W} \left( M(u,v) - \text{FreqMean} \right)^2 } \).

These two features capture the spectral energy and its variation. Higher values indicate detailed or noisy content, while lower values reflect smoother textures.

\textbf{Image Sharpness} is quantified as $\max(|I - I_{\text{blurred}}|)
$. $I$ and $I_{\text{blurred}}$ denote the grayscale and blurred image with Gaussian kernel, respectively. 

\textbf{Image Smoothness} evaluates how consistent the image's texture is.
It is quantified as $\frac{1}{1 + \text{var}(\Delta I)}$, where $\Delta I$ denotes the Laplacian of the grayscale image $I$.

\textbf{Image Contrast} measures the degree of variation in intensity across an image.
It is quantified by calculating the standard deviation of the pixel values in the grayscale image, expressed as $\text{std}(I)$. 

\textbf{Color Distribution Consistency} evaluates the variability in an image's color distribution by analyzing the standard deviation of the normalized color histogram in the HSV color space. It is calculated as $\text{std}(\tilde{H}_{HSV}(h, s, v))$, where \(\text{std}(\cdot)\) denotes the standard deviation of the normalized histogram \(\tilde{H}_{HSV}(h, s, v)\) for hue \(h\), saturation \(s\), and value \(v\).

\textbf{Object Coherence} evaluates the extent and clarity of edge detection in an image, providing insight into the consistency of object boundaries. 
It is determined using $\frac{\sum_{i,j} E(i, j)}{\sum_{i,j} 1}$, where \( E(i, j) \) represents the value of the Canny edge image at pixel \((i, j)\), and the \(\sum_{i,j} 1\) represents the total number of pixels in the image. 

\textbf{Contextual Relevance} evaluates the distribution of edge strengths across the image.
It is given by $\text{var}(\sqrt{{G_x}^2 + {G_y}^2})$, where \(\text{var}(\cdot)\) denotes the variance, and \(G_x\) and \(G_y\) are the gradients computed using the Sobel filter in the horizontal and vertical directions, respectively.

Each low-level feature is first standardized using Z-score normalization as
$
f_i(x) = \frac{v_i(x) - \mu_i}{\sigma_i},
$
where \(v_i(x)\) is the raw value of feature \(i\) for image \(x\), and \(\mu_i\), \(\sigma_i\) are the mean and standard deviation of that feature across the dataset.

To compute the Visual AI Index, we learn a set of weights that quantify how strongly each normalized feature contributes to the perceived realism of an image. We employ a logistic regression model to distinguish between real (label \(y = 1\)) and synthetic (label \(y = 0\)) images. Given a 12-dimensional feature vector \(x = [f_1, f_2, \dots, f_{12}]\), the model estimates the probability that an image is real as
$
p(x) = \frac{1}{1 + e^{-w^\top x}},
$
where \(w\) is the weight vector. The weights are learned by minimizing the binary cross-entropy loss
$
\mathcal{L}(w) = \frac{1}{N} \sum_{i=1}^{N} \left[ -y_i \log p(x_i) - (1 - y_i) \log (1 - p(x_i)) \right].
$
After optimization, the final Visual AI Index is defined as
$
\text{V\textsubscript{AI}}(x) = p(x; w^*) = \frac{1}{1 + e^{-w^{*\top} x}},
$
where \(w^*\) denotes the optimized weights obtained after training. We train two models separately, one for COCO\textsubscript{AI} and one for Twitter\textsubscript{AI}, each tailored to the distribution of real images in its respective domain. Table~\ref{tab:VAI-weights} reports the final learned weights.

\vspace{-3mm}
\begin{table}[h]
\footnotesize
\centering
\caption{Learned V\textsubscript{AI} feature weights for COCO\textsubscript{AI} and Twitter\textsubscript{AI} domains.}
\vspace{-3mm}

\label{tab:VAI-weights}
\begin{tabular}{lrr}
\toprule
\textbf{Feature} & \textbf{COCO\textsubscript{AI}} & \textbf{Twitter\textsubscript{AI}} \\
\midrule
Texture Complexity        & 4.13  & 1.15 \\
Color Dist. Consistency   & $-$0.15 & $-$0.05 \\
Object Coherence          & $-$0.87 & 1.56 \\
Contextual Relevance      & $-$1.33 & 2.42 \\
Haralick Contrast         & 1.02  & $-$4.52 \\
Haralick Correlation      & $-$0.42 & $-$0.07 \\
Haralick Energy           & $-$1.46 & $-$1.59 \\
Freq. Mean                & $-$0.87 & $-$1.36 \\
Freq. Std                 & $-$1.30 & $-$0.20 \\
Image Smoothness          & 0.37  & $-$0.06 \\
Image Sharpness           & 2.08  & 2.26 \\
Image Contrast            & 0.50  & $-$0.18 \\
\bottomrule
\end{tabular}
\vspace{-2mm}
\end{table}

\subsection{V\textsubscript{AI} Analysis}
\begin{figure*}[t]
\centering
\minipage{0.10\columnwidth}
\vspace{18mm}\includegraphics[height=4cm,width=5.5cm]{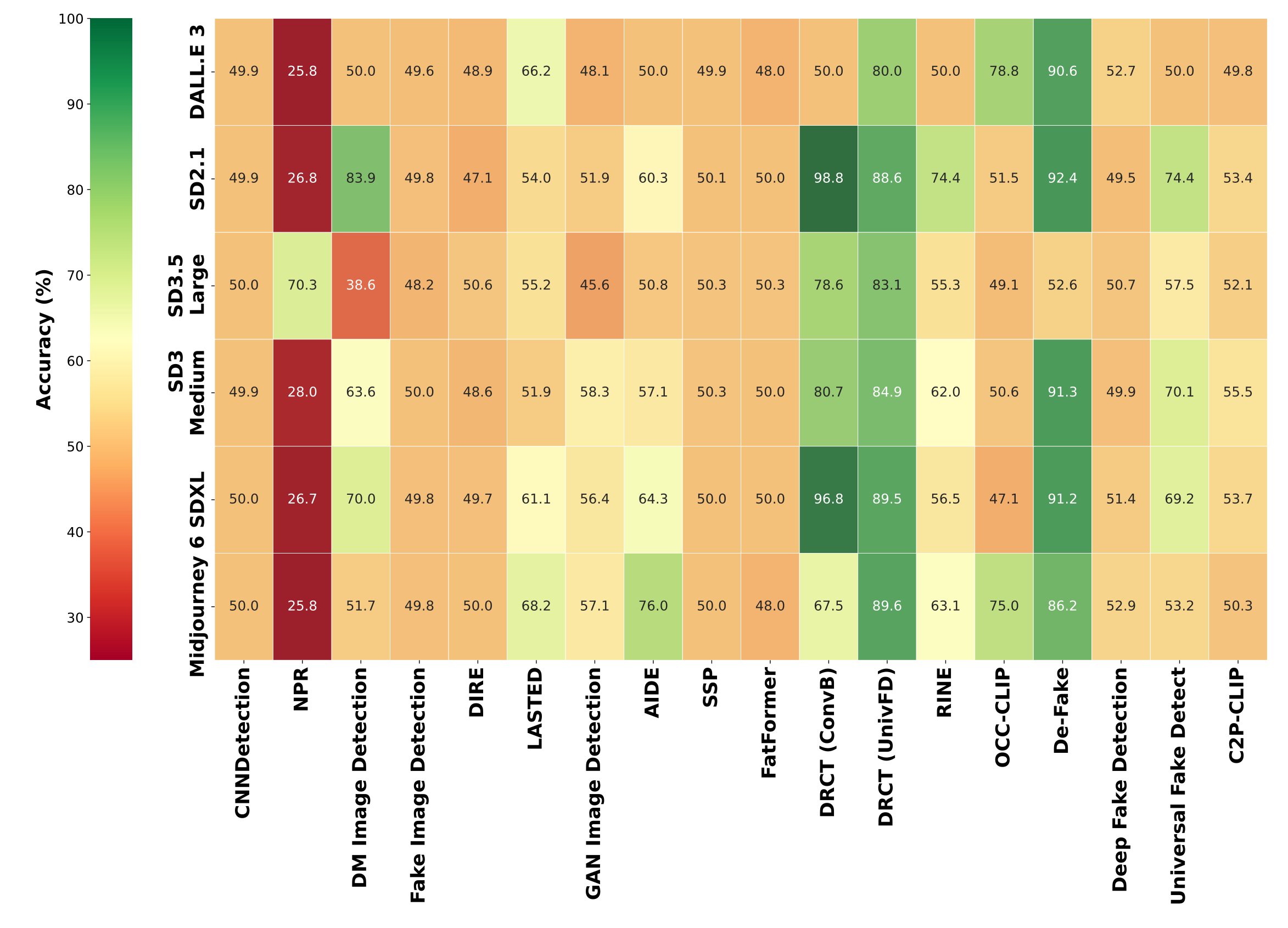}
\endminipage
\hspace{4.8cm}
\minipage{0.74\columnwidth}
\resizebox{\columnwidth}{!}{%
\begin{tabular}{lll}
\toprule
\textbf{Text-to-Image Model} & \textbf{$V_{AI}$ (0-1)}
\\ \midrule
\vspace{4.2mm}
\textbf{Real}   &   0.685  \begin{tikzpicture}
\centering
\draw[green, very thick, fill=green] (0,0)  rectangle (6.85,.1);
\end{tikzpicture}
\\
\vspace{4.2mm}
\textbf{DALL.E 3}      &   0.626  \begin{tikzpicture}
\centering
\draw[lime, very thick, fill=lime] (0,0)  rectangle (6.26,.1);
\end{tikzpicture}  
\\
\vspace{4.2mm}
\textbf{SD2.1}    &     0.568  \begin{tikzpicture}
\centering
\draw[yellow, very thick, fill=yellow] (0,0)  rectangle (5.68,.1);
\end{tikzpicture}
\\
\vspace{4.4mm}
\textbf{SD3.5 Large}    &     0.552  \begin{tikzpicture}
\centering
\draw[yellow, very thick, fill=yellow] (0,0)  rectangle (5.52,.1);
\end{tikzpicture}
\\
\vspace{4.4mm}
\textbf{SD3 Medium}    &    0.536  \begin{tikzpicture}
\centering
\draw[yellow, very thick, fill=yellow] (0,0)  rectangle (5.36,.1);
\end{tikzpicture} 
\\
\vspace{4.4mm}
\textbf{SDXL}    &    0.496  \begin{tikzpicture}
\centering
\draw[orange, very thick, fill=orange] (0,0)  rectangle (4.96,.1);
\end{tikzpicture} 
\\
\vspace{4.4mm}
\textbf{Midjourney 6}   &   0.432 \begin{tikzpicture}
\centering
\draw[red, very thick, fill=red] (0,0)  rectangle (4.32,.1);
\end{tikzpicture} 
\\
\bottomrule
\end{tabular}%
}
\endminipage
\minipage{0.10\columnwidth}
\includegraphics[height=3cm]{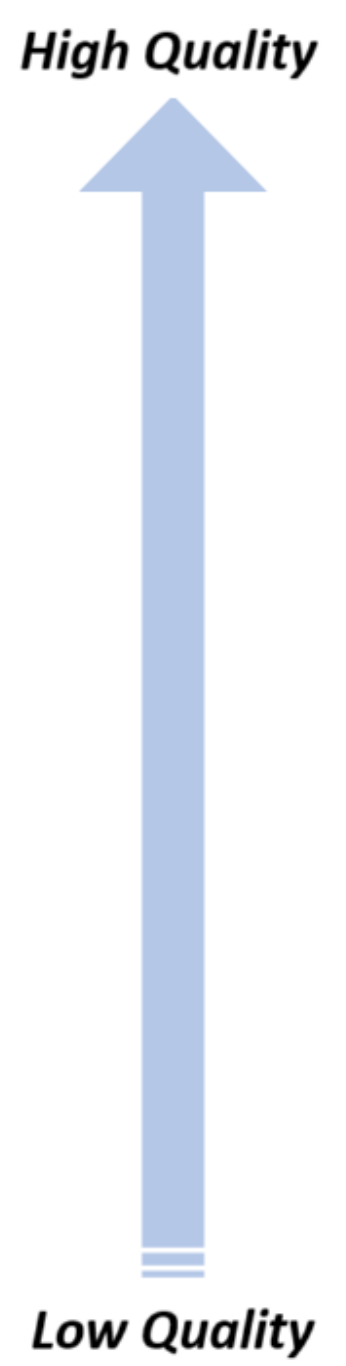}
\endminipage

\caption{Right: $V_{AI}$ scores of COCO\textsubscript{AI} dataset. Left: Accuracy heat maps showing the average accuracy of each AGID method.
}
\label{fig:adi_coco}
\vspace{-2mm}
\end{figure*}

\begin{figure*}[t]
\centering
\vspace{-7mm}

\minipage{0.10\columnwidth}
\vspace{16mm}\includegraphics[height=4cm,width=5.5cm]{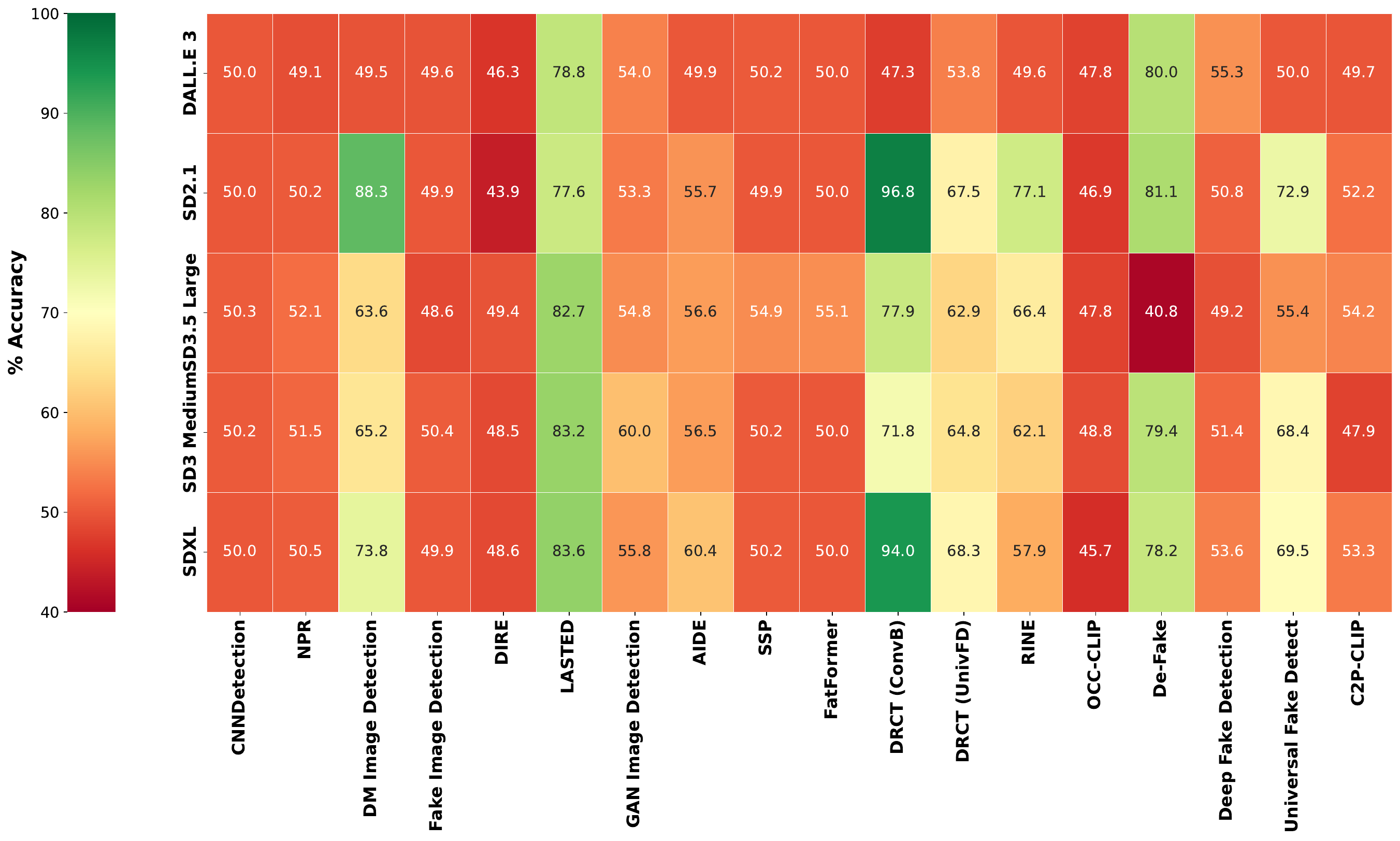}
\endminipage
\hspace{4.8cm}
\minipage{0.74\columnwidth}
\resizebox{\columnwidth}{!}{%
\begin{tabular}{lll}
\toprule
\textbf{Text-to-Image Model} & \textbf{$V_{AI}$ (0-1)}
\\ \midrule
\vspace{4.5mm}
\textbf{Real}   &   0.624  \begin{tikzpicture}
\centering
\draw[green, very thick, fill=green] (0,0)  rectangle (6.24,.1);
\end{tikzpicture}
\\
\vspace{4.5mm}
\textbf{DALL-E 3}   &   0.593  \begin{tikzpicture}
\centering
\draw[lime, very thick, fill=lime] (0,0)  rectangle (5.93,.1);
\end{tikzpicture}
\\
\vspace{5mm}
\textbf{SD2.1 }   &   0.588  \begin{tikzpicture}
\centering
\draw[yellow, very thick, fill=yellow] (0,0)  rectangle (5.88,.1);
\end{tikzpicture}
\\
\vspace{5mm}
\textbf{SD3.5 Large}      &   0.580  \begin{tikzpicture}
\centering
\draw[yellow, very thick, fill=yellow] (0,0)  rectangle (5.80,.1);
\end{tikzpicture}  
\\
\vspace{5mm}
\textbf{SD3 Medium}    &    0.579  \begin{tikzpicture}
\centering
\draw[yellow, very thick, fill=yellow] (0,0)  rectangle (5.79,.1);
\end{tikzpicture} 
\\
\vspace{3mm}
\textbf{SDXL }   &   0.573 \begin{tikzpicture}
\centering
\draw[orange, very thick, fill=orange] (0,0)  rectangle (5.73,.1);
\end{tikzpicture}  
\\
\bottomrule
\end{tabular}%
}
\endminipage
\minipage{0.10\columnwidth}
\includegraphics[height=3cm]{sec/images/arrow_fit.pdf}
\endminipage

\caption{Right: $V_{AI}$ scores of Twitter\textsubscript{AI} dataset. Left: Accuracy heat maps showing the average accuracy of each AGID method.
}
\label{fig:adi_twitter}
\end{figure*}

We report the average V\textsubscript{AI} scores for real and generated images across the COCO\textsubscript{AI} and Twitter\textsubscript{AI} subsets in Figures \ref{fig:adi_coco} and \ref{fig:adi_twitter}. As expected, real images achieve the highest V\textsubscript{AI} scores in both domains, reflecting the benchmark’s ability to assign higher realism to naturally occurring images. Among generative models, DALL{\tiny·}E~3 obtains the highest V\textsubscript{AI} in both subsets (0.626 for COCO\textsubscript{AI}, 0.593 for Twitter\textsubscript{AI}), indicating its outputs most closely align with real images in terms of low-level features such as texture complexity, edge coherence, and color consistency. A cluster of diffusion-based models, SD2.1, SD3 Medium, and SD3.5 Large, follow DALL{\tiny·}E~3 with relatively similar V\textsubscript{AI} scores, suggesting comparable levels of photorealism. SDXL ranks lower across both domains, i.e. 0.496 COCO\textsubscript{AI} and 0.573 Twitter\textsubscript{AI}, which may be attributed to its tendency toward stylistic exaggeration or generation artifacts that deviate from natural image statistics. These artifacts can influence frequency-domain, edge-based, or texture descriptors negatively, despite the model's high perceptual fidelity. Midjourney yields the lowest V\textsubscript{AI} in the COCO\textsubscript{AI} subset (0.432) and is excluded from the Twitter\textsubscript{AI} analysis due to the unavailability of corresponding generated images.  

The accuracy heat maps in Figures \ref{fig:adi_coco} and \ref{fig:adi_twitter} highlight differences in AGID methods across models. The results indicates that the texture and artifact characteristics differ significantly across models, affecting detection reliability. While some detection methods, like De-Fake and DRCT, performed consistently well, the $V_{AI}$ scores reveal that realism score of generated image plays a significant role in detection difficulty. 

Factor-level analysis is provided in the Appendix \ref{sec:vai}, where we highlight specific contributing factors through LBP (Local Binary Pattern) analysis, and pairwise factor plots.

\subsection{Correlation with Detection Accuracy}
\label{sec:correlation}

To evaluate whether the Visual AI Index aligns with the difficulty of detecting AI-generated images, we compute the Pearson correlation coefficient between average V\textsubscript{AI} scores and AGID detection accuracy across five generative models. The Pearson correlation coefficient $\rho$ is defined as:

\[
\rho = \frac{\sum_{i=1}^{n} (x_i - \bar{x})(y_i - \bar{y})}{\sqrt{\sum_{i=1}^{n} (x_i - \bar{x})^2} \sqrt{\sum_{i=1}^{n} (y_i - \bar{y})^2}},
\]

where $x_i$ and $y_i$ represent the V\textsubscript{AI} score and AGID accuracy for model $i$, and $\bar{x}$ and $\bar{y}$ are their respective means. The coefficient $\rho$ ranges from $-1$ to $1$: a value near $1$ implies a strong positive correlation, near $-1$ implies a strong negative correlation, and a value near $0$ suggests no linear relationship. We compute $\rho$ separately for the Twitter\textsubscript{AI} and COCO\textsubscript{AI} datasets. As shown in Table \ref{tab:correlation}, our results indicate a moderate inverse relationship: models with higher visual realism tend to be harder to detect. However, the correlations are not statistically significant, likely due to the small number of generative models, i.e. $n = 5$, and should be interpreted cautiously.

\begin{table}[h]
\small
\centering
\caption{Pearson correlation between V\textsubscript{AI} and AGID detection accuracy.}
\vspace{-3mm}

\begin{tabular}{lcc}
\toprule
\textbf{Dataset} & \textbf{Pearson $\rho$} & \textbf{p-value} \\
\midrule
Twitter\textsubscript{AI} & $-0.503$ & $0.388$ \\
COCO\textsubscript{AI}    & $-0.532$ & $0.356$ \\
\bottomrule
\end{tabular}
\vspace{-2mm}
\label{tab:correlation}
\end{table}






\section{Conclusion}
\label{sec:conclusion}

In this paper, we introduced (VCT$^2$), a comprehensive benchmark for evaluating AI-generated image detection (AGID) across diverse generative models, including cutting-edge proprietary systems like DALL·E 3 and Midjourney 6. By incorporating both real-world prompts and standardized captions, VCT² offers a challenging, realistic dataset for assessing generalization. The VCT$^2$ benchmark provides a critical resource for evaluating AGID techniques under challenging and varied conditions, highlighting performance gaps and guiding the development of more robust detection methods.

To assess the realism of images, we present the Visual AI Index ($V_{AI}$) that evaluates characteristics like texture complexity, Haralick correlation, frequency mean, and image sharpness. Our findings reveal that real images generally achieve higher $V_{AI}$ scores than AI-generated images.

\section*{Limitations and Future Work}

While our work provides a strong foundation for evaluating AGID methods and realism metrics, future directions include expanding to diverse domains (e.g., social platforms, synthetic video), integrating temporal and multimodal features into V\textsubscript{AI}, and adapting it for localization or attribution. We also plan to explore human alignment and psychometric grounding of these continuous realism scores. As generative models evolve, updating the benchmark and exploring hybrid detection techniques will be key to ensuring resilience against increasingly sophisticated AI imagery.

\bibliography{main}

\clearpage

\section*{Appendix}

\appendix

\section{Detection Techniques}
\label{agid_techniques}
This appendix provides detailed descriptions of the \textbf{17} AI-generated image detection (AGID) techniques evaluated on our benchmark. These methods span three major detection paradigms: generation artifact-based, feature representation-based, and hybrid approaches. This taxonomy is designed to reflect the breadth of design assumptions across the literature and serves as the foundation for our performance analysis in Section~\ref{sec:results}. 

Artifact-based methods exploit low-level visual artifacts, such as frequency distortions, edge inconsistencies, or upsampling traces, introduced during the image generation process. Feature-based methods, in contrast, analyze semantic-level inconsistencies by leveraging deep neural representations from CNNs, vision transformers, or CLIP encoders. Hybrid methods combine both low-level and high-level signals, often incorporating alignment objectives or learned fusion strategies to improve robustness.

The detectors described here were selected based on recency, diversity, and public availability, and represent both classical and state-of-the-art AGID strategies. Each technique is evaluated under zero-shot settings using default public checkpoints, and grouped by detection paradigm below.

\subsection{Generation Artifact-Based Detection}
Generation artifact-based detection techniques focus on identifying visual artifacts produced during the generation process, analyzing both spatial and frequency domains.

\citep{tan2023rethinking} found that the up-sampling operator introduces artifacts not only in frequency patterns but also in pixel arrangements within images. The authors introduce the concept of Neighboring Pixel Relationships to capture and characterize these generalized structural artifacts caused by up-sampling operations.

\citep{corvi2023detection} observed that synthetic images, especially those generated by diffusion models like GLIDE and Stable Diffusion, exhibit distinctive differences in mid-to-high frequency signals compared to real images. However, this distinction is less pronounced in images produced by newer models, such as DALL-E and ADM. Although their method accurately distinguishes synthetic and real images in controlled settings, it struggles in real-world scenarios.

\citep{doloriel2024frequency} explored masked image modeling for universal fake image detection. Their approach involves both spatial and frequency domain masking, leading to a deepfake detector based on frequency masking.

\citep{chendrct} enhance detector generalization diffusion generated images by generating hard samples through high-quality diffusion reconstruction. These reconstructed images, which closely resemble real ones but retain subtle artifacts, train detectors to differentiate between real and generated images, including those from unseen diffusion models.

\subsection{Feature Representation-Based Detection}
Feature representation-based detection methods distinguish real images from synthesized images by leveraging deep learning models to extract and analyze complex visual features.

\citep{wang2020cnn} proposed a universal detector using a ResNet-50 classifier \cite{he2016deep} with random blur and JPEG compression data augmentation. When trained on images generated by a single CNN generator (ProGAN), their model demonstrated strong generalization across unseen architectures, including StyleGAN2 \cite{karras2020analyzing} and StyleGAN3 \cite{karras2021alias}.

\citep{mandelli2022detecting} developed a GAN-generated image detector based on an ensemble of CNNs. Their method emphasizes generalization by ensuring orthogonal results from CNNs and prioritizing original images during testing.

\citep{wang2023dire} introduced a technique that measures the error between an input image and its reconstructed counterpart generated by a pre-trained diffusion model. They observed that diffusion-generated images are more accurately reconstructed than real images, highlighting a key discrepancy for detection.

\citep{lasted} employed language-guided contrastive learning to capture inherent differences in the distributions of real and synthetic images. Their method augments training images with designed textual labels, enabling joint image-text contrastive learning for forensic feature extraction.

\citep{sha2023fake} addressed the challenges of fake image detection and attribution. Their approach involves: 
(i) building a machine learning classifier to detect fake images generated by various text-to-image models, including DALL-E 2, Stable Diffusion, GLIDE, and Latent Diffusion, and benchmark prompt-image datasets such as MS COCO and Flickr30k \cite{young2014image}; 
(ii) attributing fake images to their respective generative models to enhance accountability; and 
(iii) examining how prompts influence detection and attribution performance.

\citep{aghasanli2023interpretable} introduced a deepfake detection method that combines fine-tuned Vision Transformers (ViTs) with Support Vector Machines (SVMs). Their method provides interpretability by analyzing the SVMs' support vectors to distinguish between real and fake images generated by various diffusion models.

\citep{chen2024single} proposed a straightforward method that extracts the simplest patch from an image and sends its noise pattern to a binary classifier, demonstrating effectiveness with minimal complexity.

\citep{koutlis2024leveraging} utilized intermediate outputs from CLIP's image encoder for enhanced AI-generated image detection. They introduced a Trainable Importance Estimator to dynamically assess the contributions of each Transformer block, boosting generalization across generative models.

\citep{liumodel} presented OCC-CLIP, a CLIP-based framework for few-shot one-class classification. This method is particularly effective when only a few images generated by a model are available, and access to the model's parameters is restricted. OCC-CLIP combines high-level and adversarial data augmentation techniques to attribute images to specific generative models accurately.

To enhance generalization to unseen generative models, \cite{ojha2023fakedetect} propose an approach that avoids explicitly training a classifier to distinguish real from fake images. Instead, their method leverages the feature space of large pre-trained vision-language models and employs techniques such as nearest neighbor classification.

\cite{tan2024c2p} enhance the image encoder's ability to detect deepfakes by integrating category-related prompts into the text encoder of CLIP.

\subsection{Hybrid Techniques}
Hybrid techniques combine low-level artifact analysis with high-level semantic feature extraction to effectively distinguish AI-generated images from real ones.

\cite{yan2024sanity} propose a hybrid-feature model that integrates high-level semantic information (using CLIP) with low-level artifact analysis to improve detection robustness.

\cite{liu2024forgeryaware} incorporate a forgery-aware adapter that integrates local forgery traces from both image and frequency domains. Their method employs language-guided alignment, using contrastive objectives between image features and text prompts to enhance generalization.

To guide our benchmark evaluation, we selected 17 state-of-the-art AGID methods spanning all three categories. This categorization enables us to evaluate model robustness from complementary perspectives: from low-level artifact exploitation to high-level semantic inconsistency analysis.

\section{Detection Performance Overview}
\label{app:det}
Tables \ref{tab:res_coco} and \ref{tab:res_tweeter} provide the performance of detection techniques across COCO\textsubscript{AI} and Twitter\textsubscript{AI} synthetic datasets, respectively. The metrics measured are Accuracy (Acc), Recall (R), and Precision (P), providing insights into each model's ability to differentiate real from AI-generated images.
Below, we analyze performance across different detection techniques.
\begin{itemize}
    \item \textbf{CNNDetection, NPR and Fake Image Detection}: These methods showed variable results, characterized by low recall but higher precision across several models. This indicates a tendency to correctly identify generated images when detected, but with many instances being missed (false negatives).
    
    \item \textbf{DM Image Detection and De-Fake}: DM Image Detection demonstrated high precision across all models, particularly excelling with Stable Diffusion versions and Midjourney 6, effectively capturing generated images. De-Fake consistently maintains strong metrics across SD (2.1, XL and 3), DALL.E 3, and Midjourney 6 but struggles with SD3.5 Large images, exhibiting lower accuracy, precision, and recall. This drop in performance likely results from SD3.5’s refined generation and post-processing that minimize the artifacts and noise patterns AGID techniques depend on.

    \item \textbf{GAN Image Detection, SSP and DIRE}: These methods had mixed performance, particularly excelling in recall. 

    \item \textbf{DRCT (ConvB and UnivB)}: Both versions of DRCT showed strong accuracy, recall, and precision across most models but experienced a slight performance drop with DALL.E 3, indicating challenges with proprietary models.

    \item \textbf{OCC-CLIP and Deep Fake Detection}: OCC-CLIP had lower recall with SDXL but balanced performance for DALL.E 3 and Midjourney 6; while Deep Fake Detection demonstrated steady, consistent performance, with all of its metrics remaining within a similar range.

    \item \textbf{Universal Fake Detect}: Universal Fake Detect performed better on SD (2.1, XL, and 3) models but its performance dropped when applied to SD3.5, DALL.E 3, and Midjourney 6. Notably, we observed a significant increase in recall for DALL.E 3-generated images across both datasets.

    \item \textbf{C2P CLIP}: C2P CLIP consistently performs poorly with low accuracy and recall, clearly showing that it often misses AI-generated images. Although its precision remains high across both datasets overall, it declines significantly for DALL.E 3 images in both datasets and for SD3 images in the Twitter dataset.
    
\end{itemize}

\begin{table*}[ht]
\centering
\caption{Real images and synthetic images generated by different models.}
\label{tab:img_samples}
\resizebox{\linewidth}{!}{%
\begin{tabular}{ccccccc}
\toprule

Real Image & SD2.1 & SDXL & SD3 Medium & SD3.5 Large & DALL.E 3 & Midjourney 6\\ 
\toprule

\includegraphics[width=0.125\textwidth, height = 0.1\textwidth]{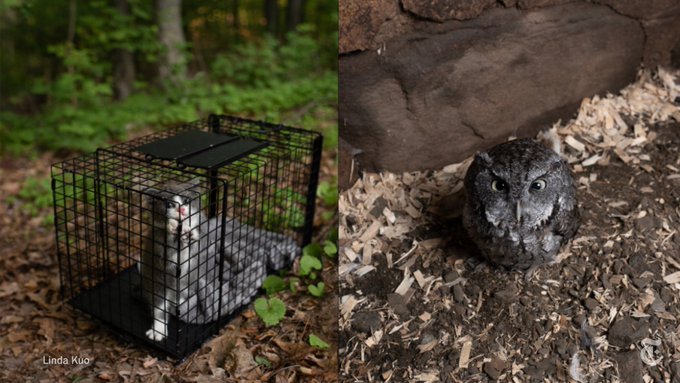} & \includegraphics[width=0.125\textwidth, height = 0.1\textwidth]{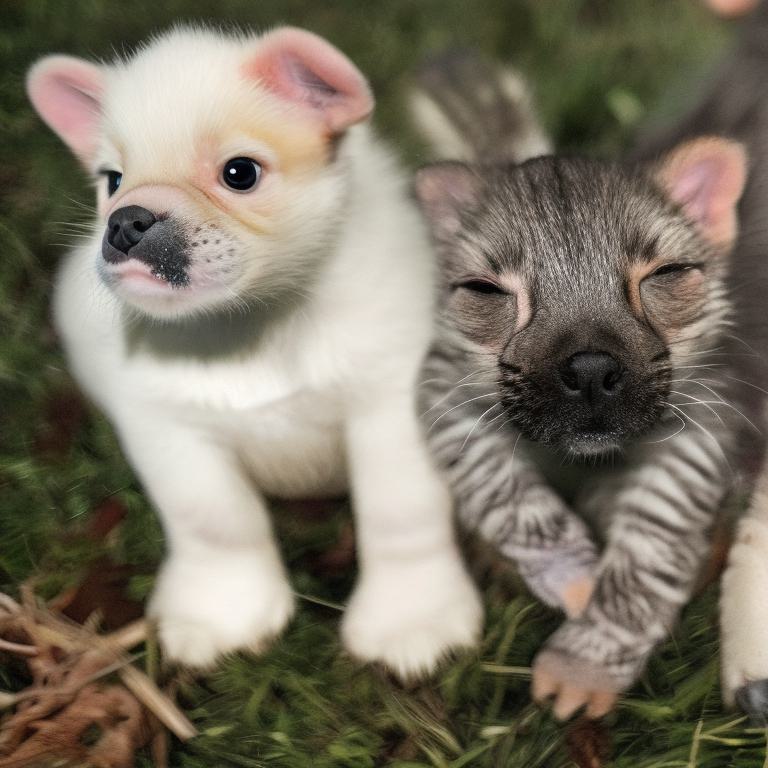} & \includegraphics[width=0.125\textwidth, height = 0.1\textwidth]{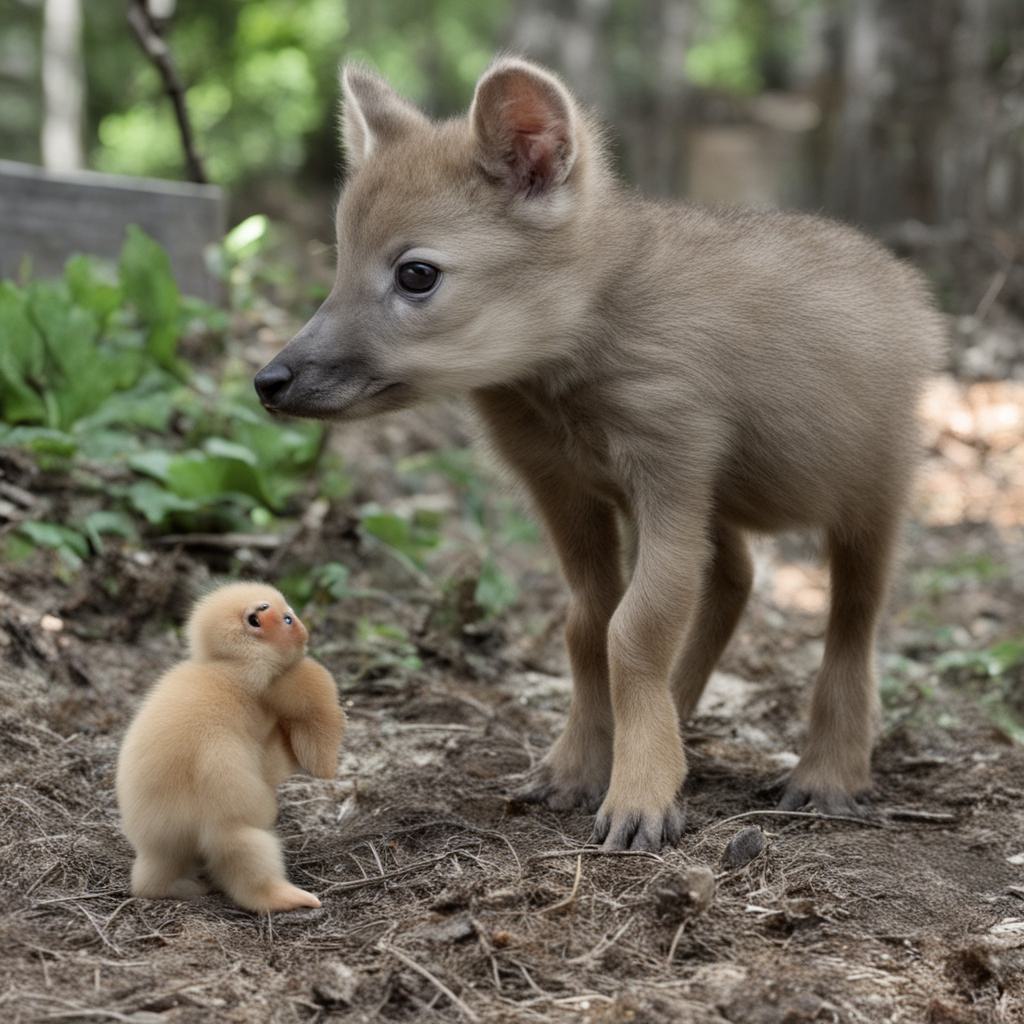} & \includegraphics[width=0.125\textwidth, height = 0.1\textwidth]{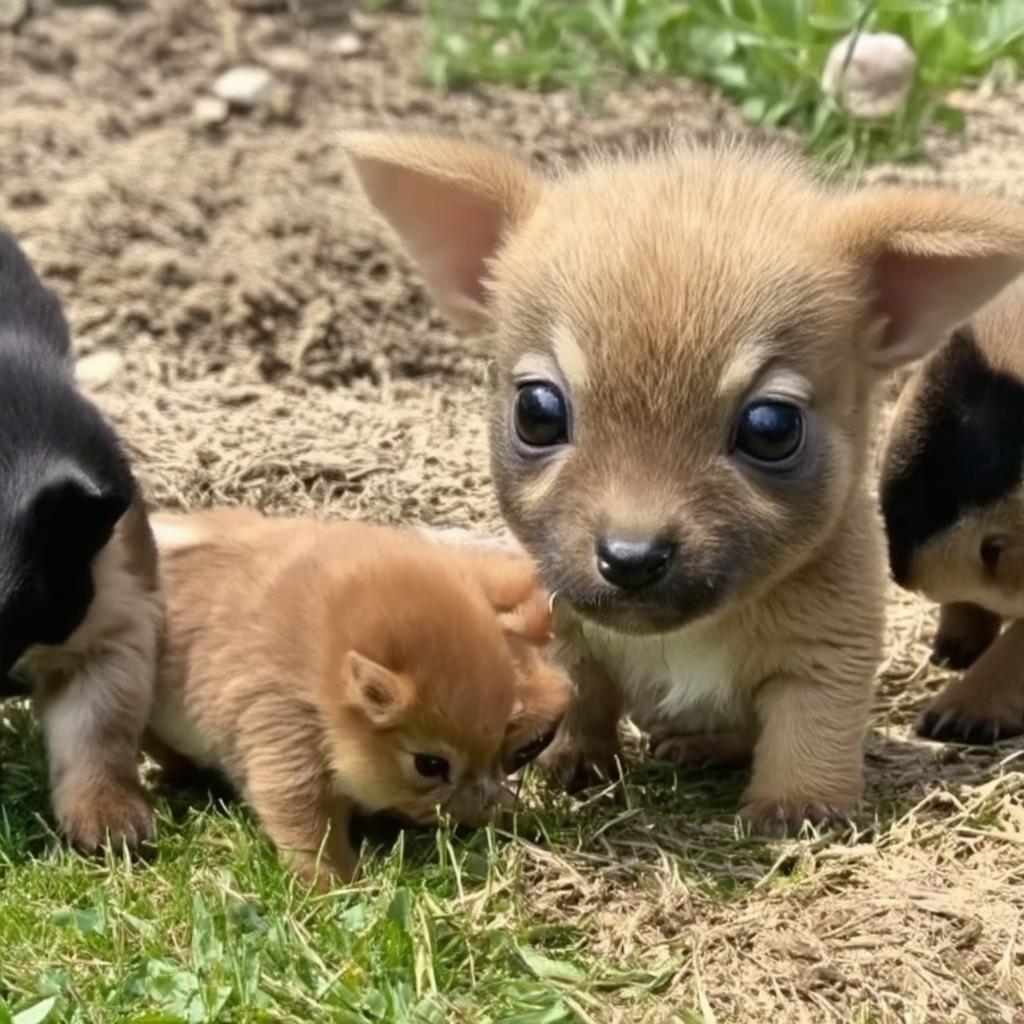} & \includegraphics[width=0.125\textwidth, height = 0.1\textwidth]{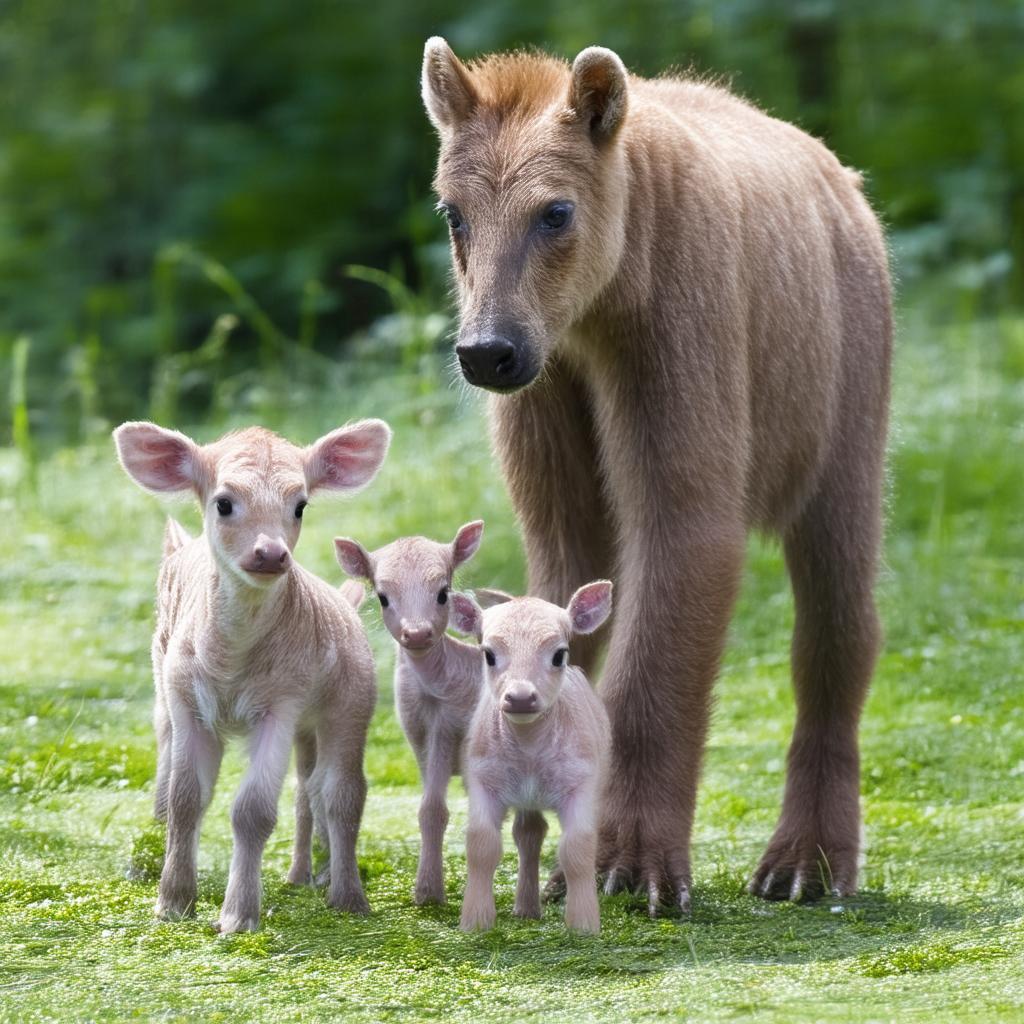} & \includegraphics[width=0.125\textwidth, height = 0.1\textwidth]{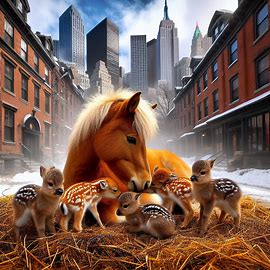} & \includegraphics[width=0.125\textwidth, height = 0.1\textwidth]{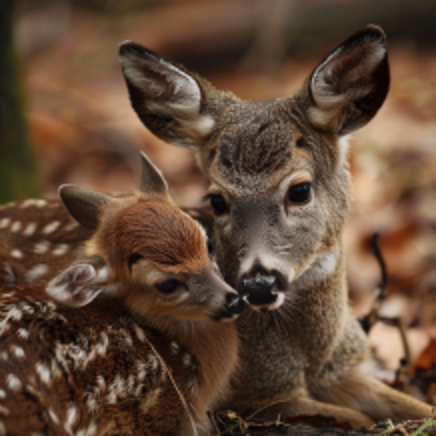} \\

\includegraphics[width=0.125\textwidth, height = 0.1\textwidth]{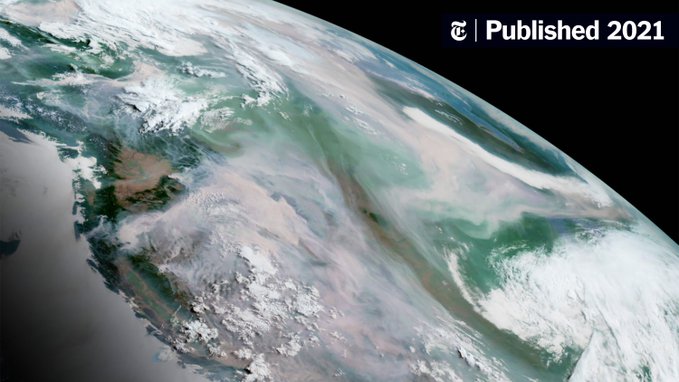} & \includegraphics[width=0.125\textwidth, height = 0.1\textwidth]{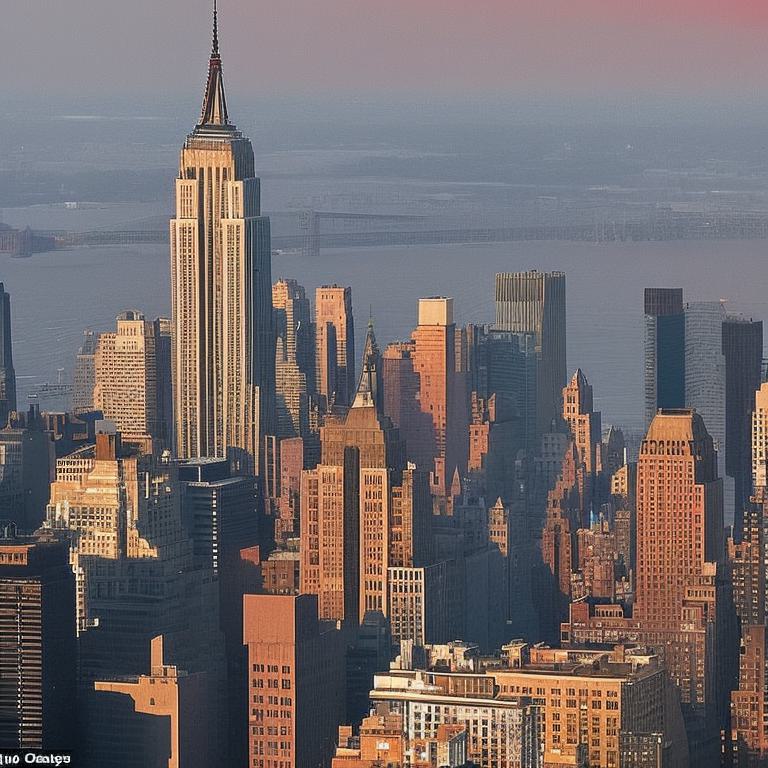} & \includegraphics[width=0.125\textwidth, height = 0.1\textwidth]{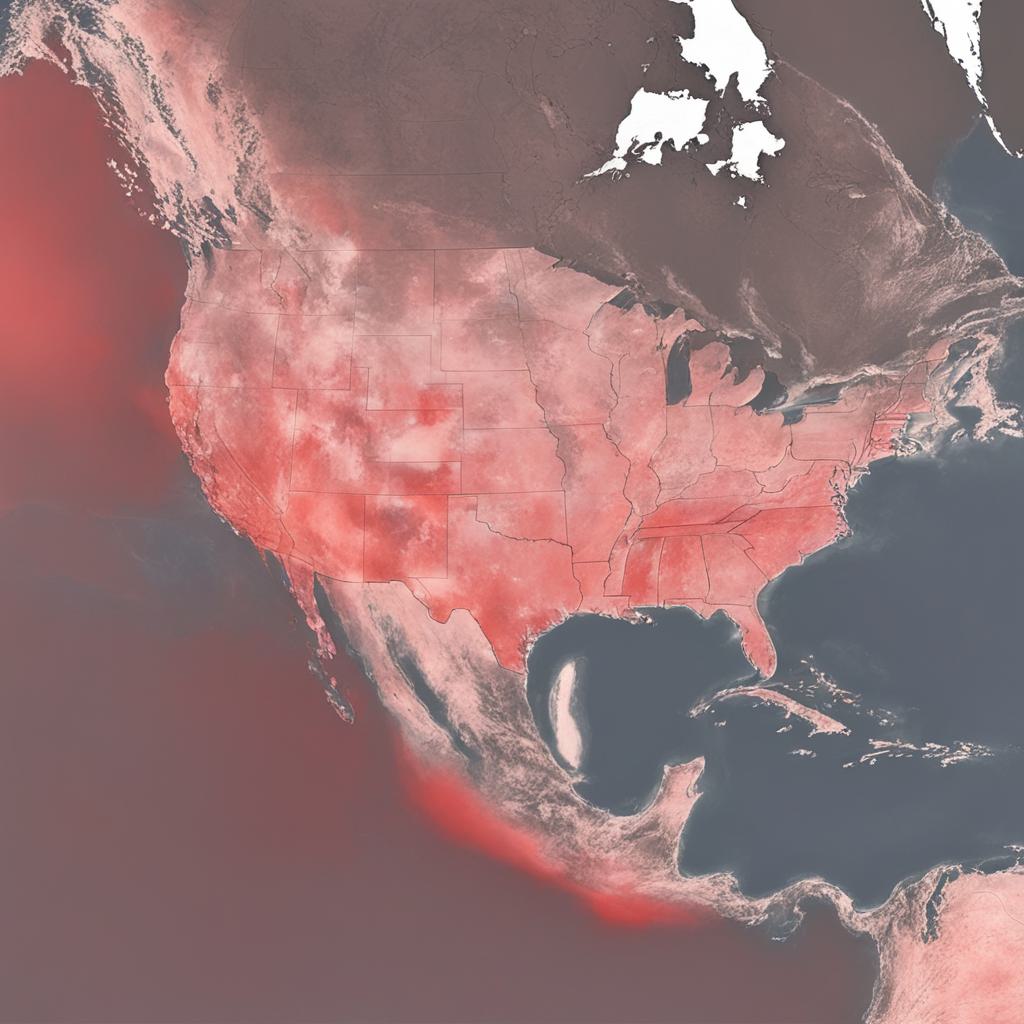} & \includegraphics[width=0.125\textwidth, height = 0.1\textwidth]{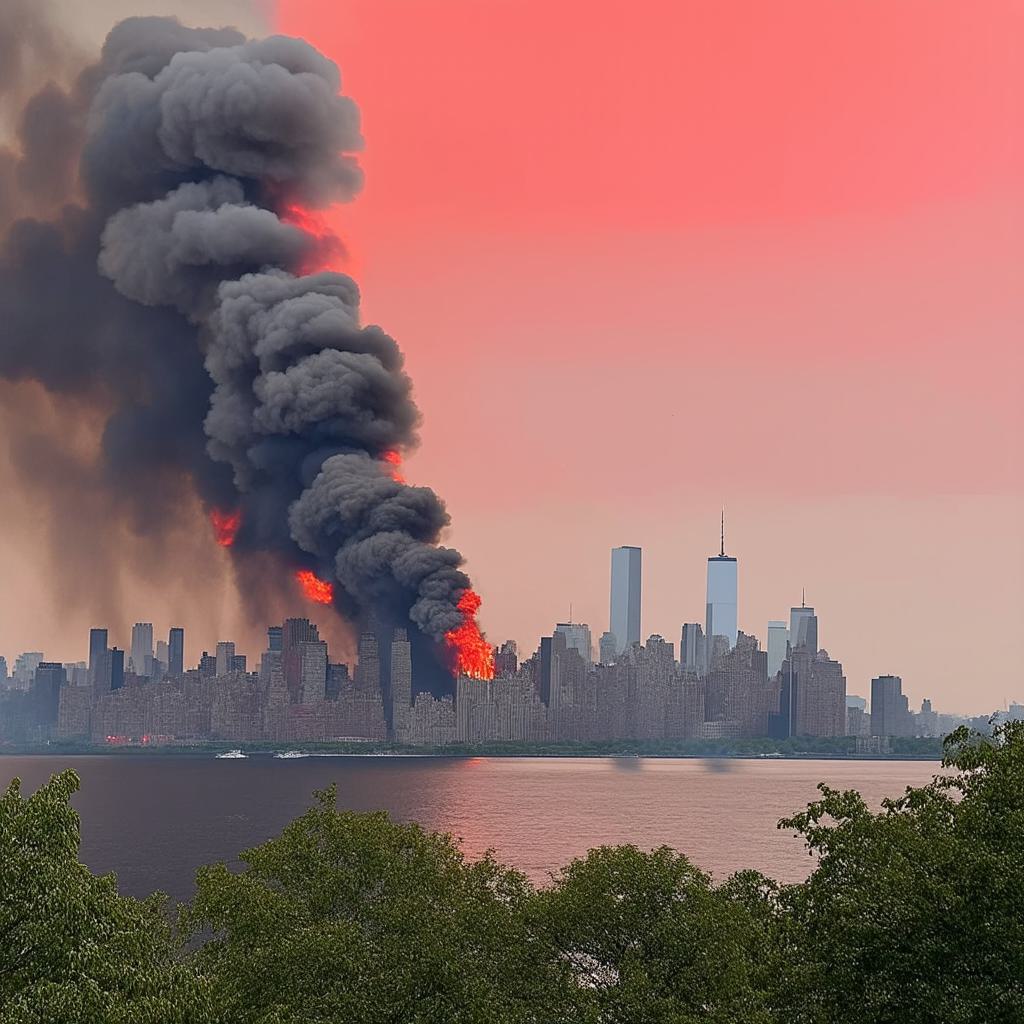} & \includegraphics[width=0.125\textwidth, height = 0.1\textwidth]{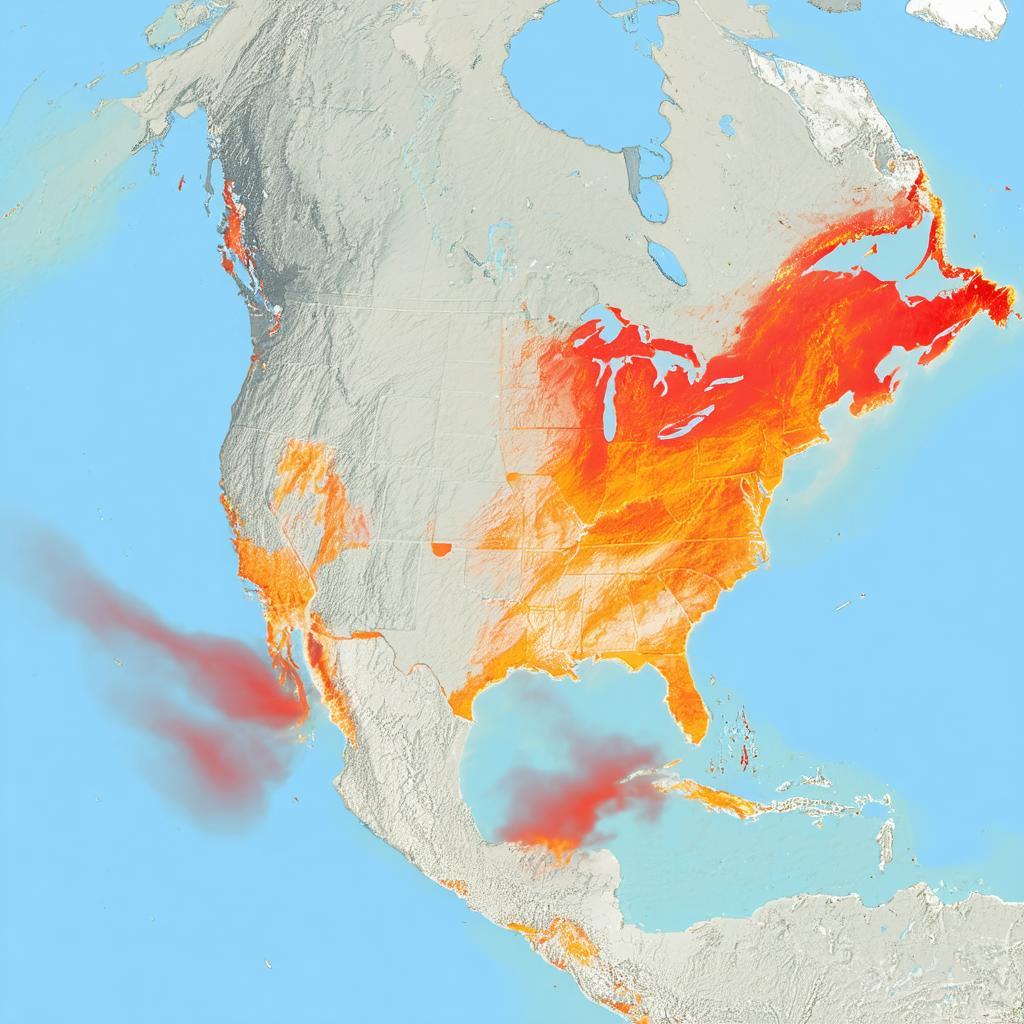} & \includegraphics[width=0.125\textwidth, height = 0.1\textwidth]{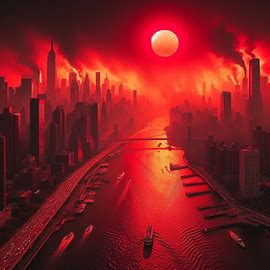} & \includegraphics[width=0.125\textwidth, height = 0.1\textwidth]{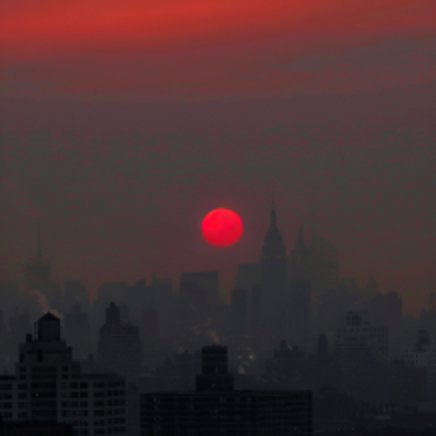} \\

\includegraphics[width=0.125\textwidth, height = 0.1\textwidth]{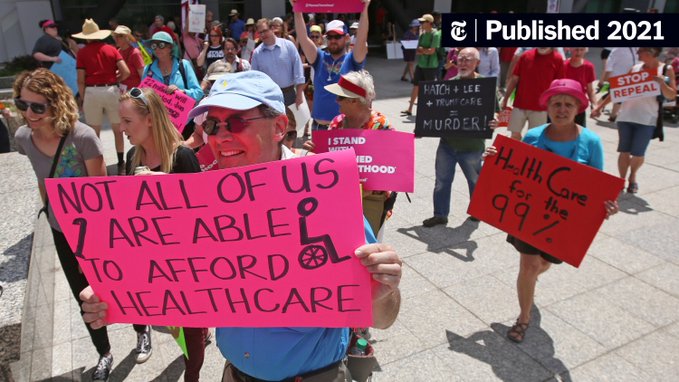} & \includegraphics[width=0.125\textwidth, height = 0.1\textwidth]{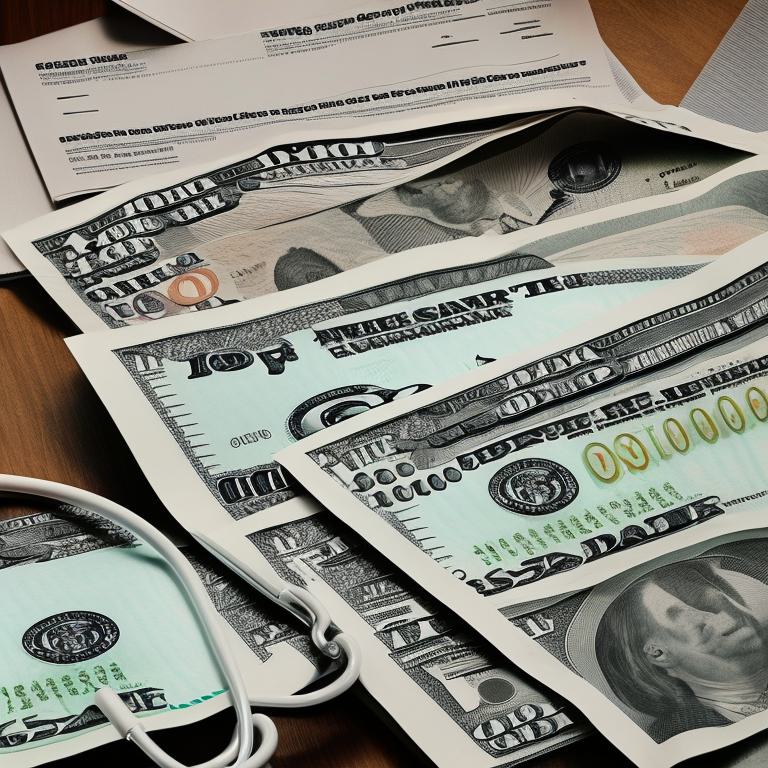} & \includegraphics[width=0.125\textwidth, height = 0.1\textwidth]{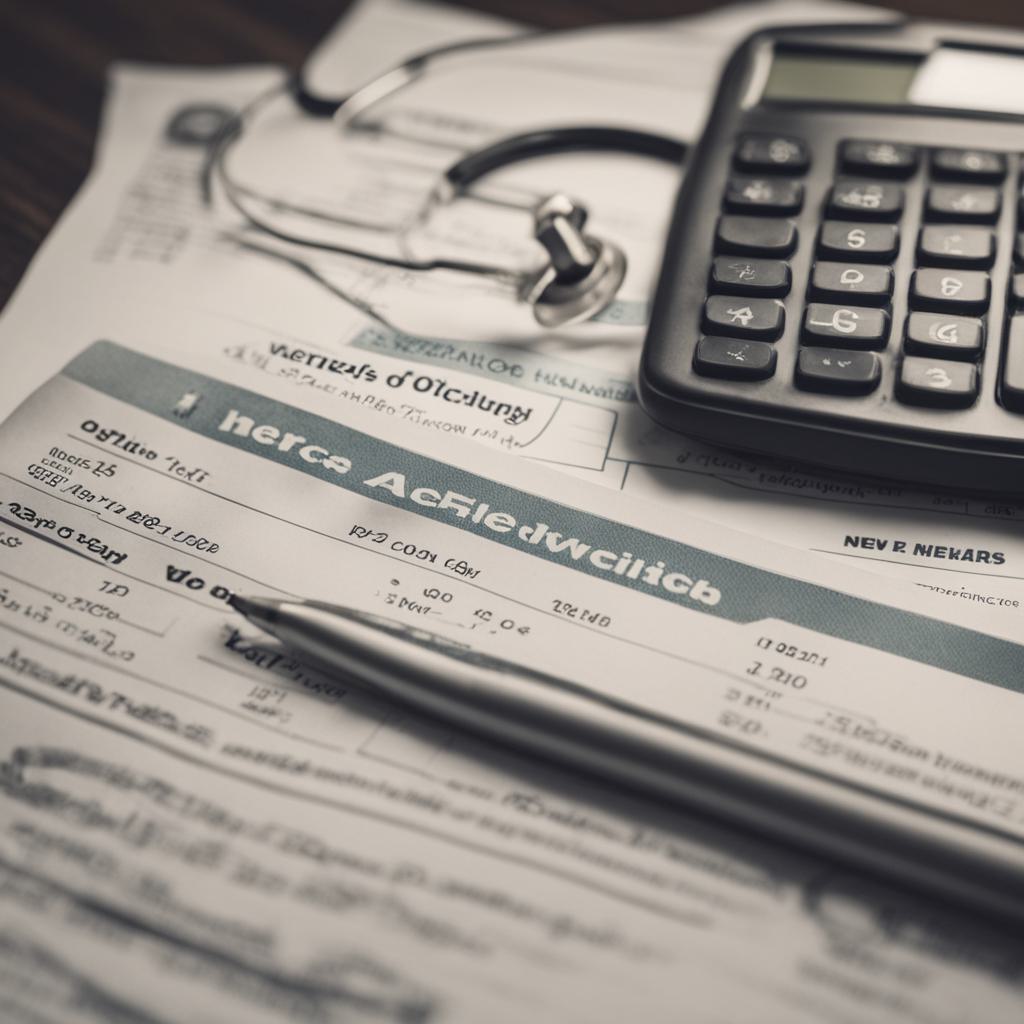} & \includegraphics[width=0.125\textwidth, height = 0.1\textwidth]{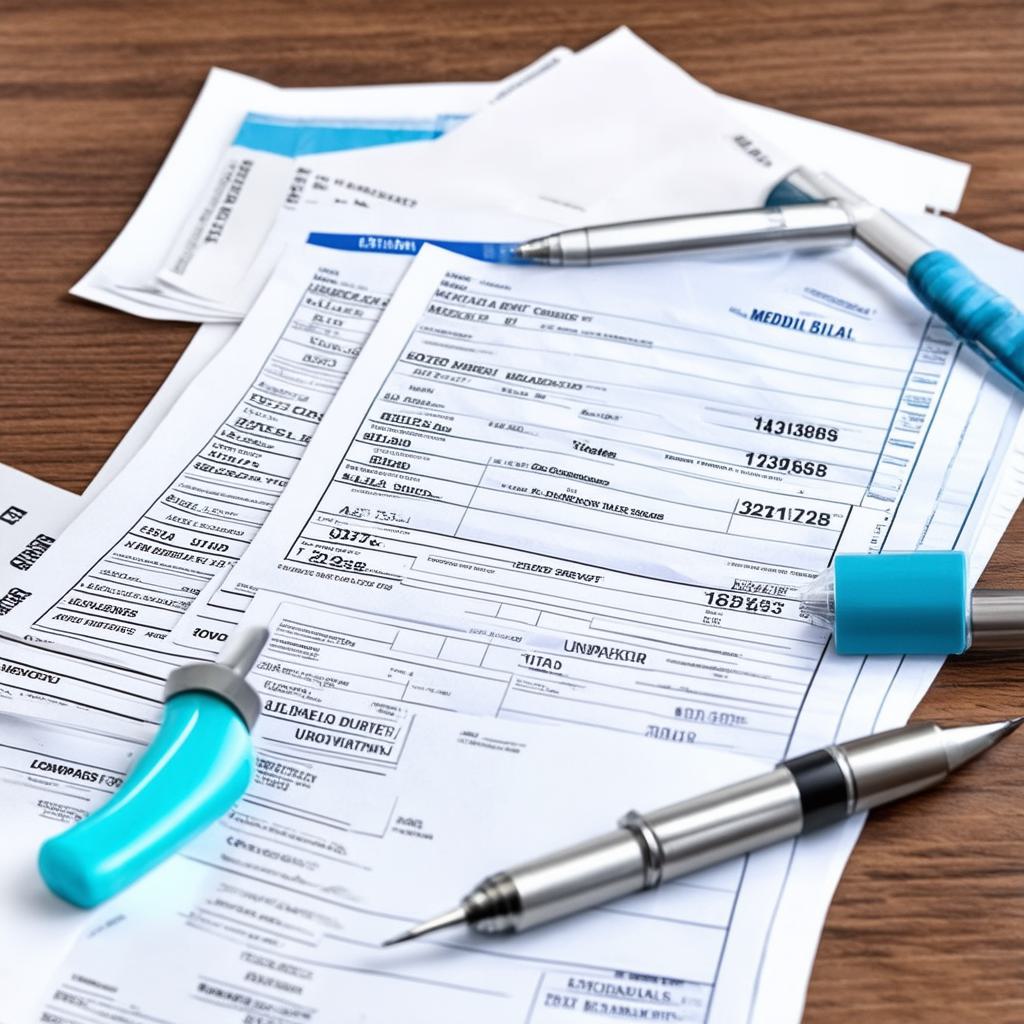} & \includegraphics[width=0.125\textwidth, height = 0.1\textwidth]{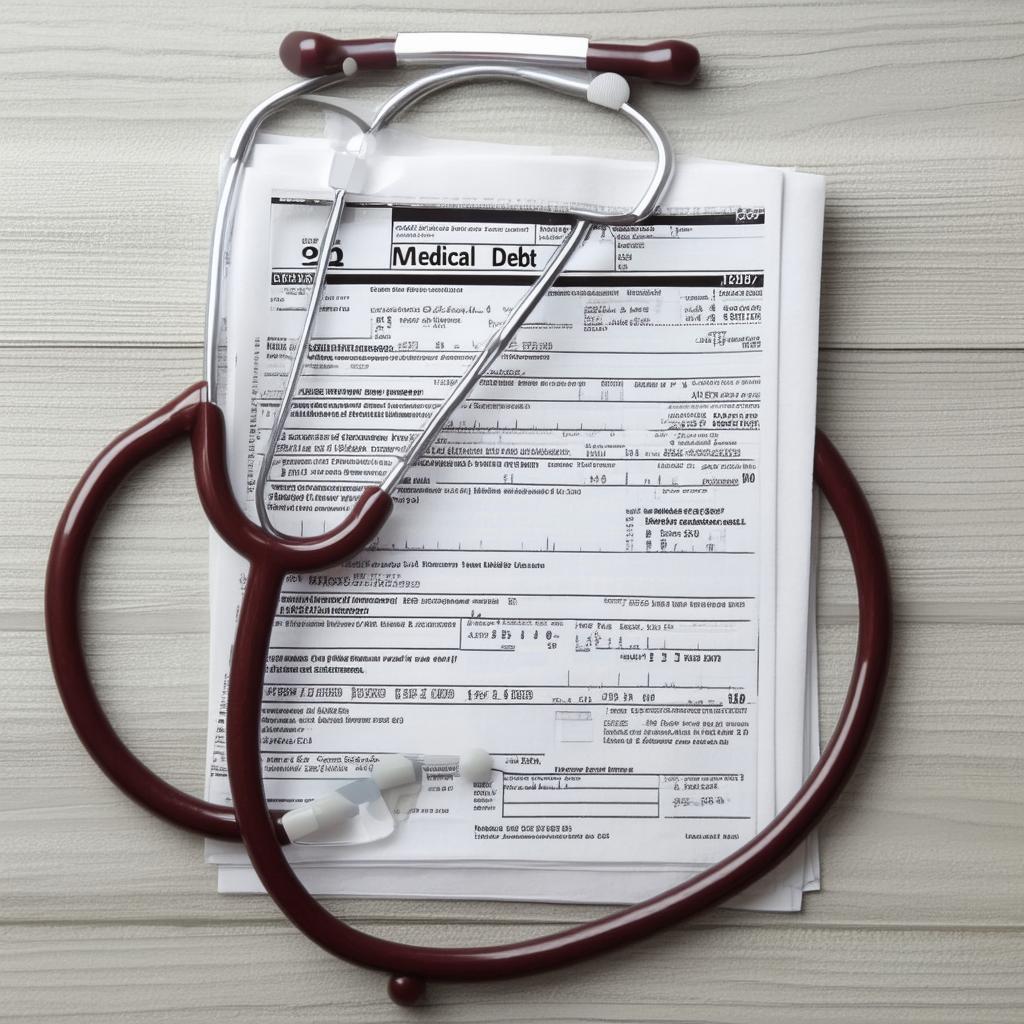} & \includegraphics[width=0.125\textwidth, height = 0.1\textwidth]{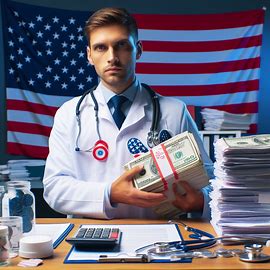} & \includegraphics[width=0.125\textwidth, height = 0.1\textwidth]{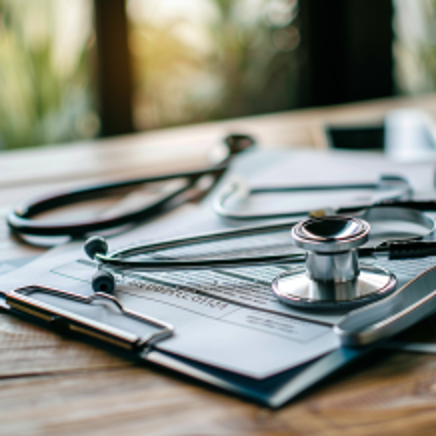} \\ 

\includegraphics[width=0.125\textwidth, height = 0.1\textwidth]{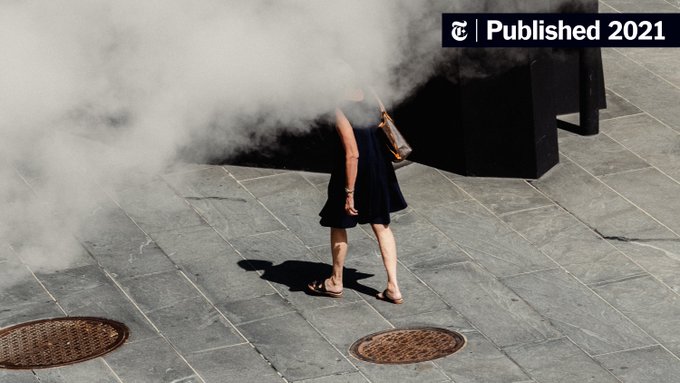} & \includegraphics[width=0.125\textwidth, height = 0.1\textwidth]{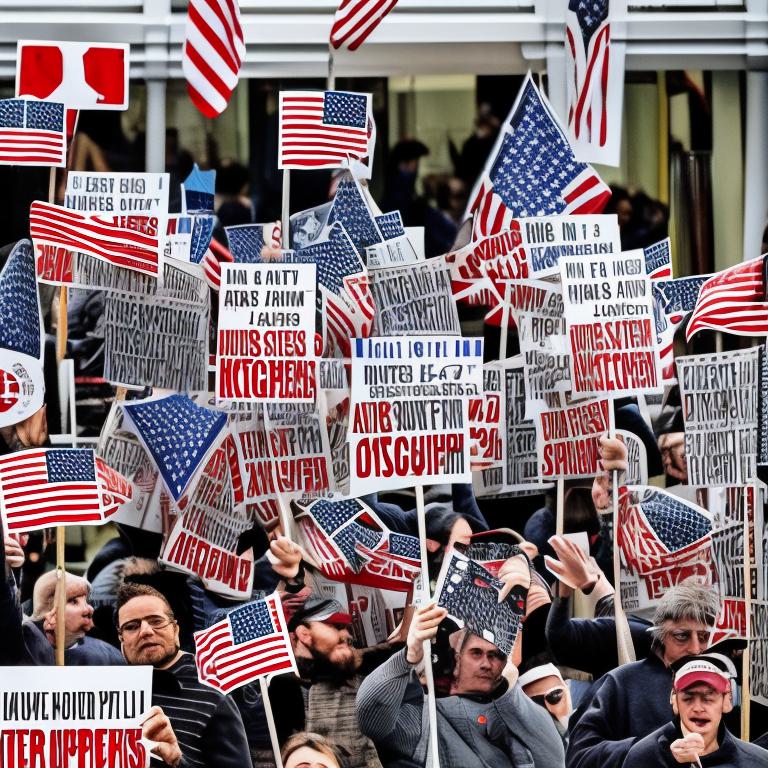} & \includegraphics[width=0.125\textwidth, height = 0.1\textwidth]{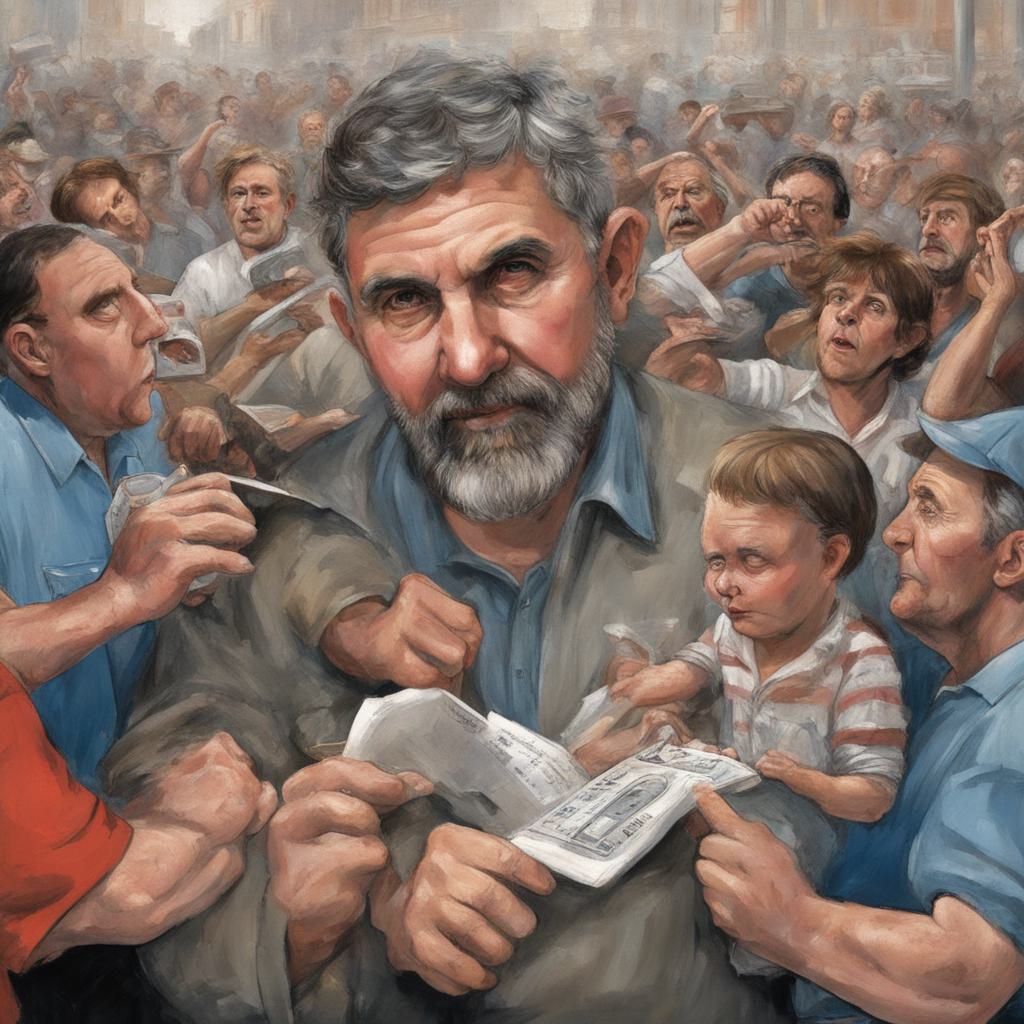} & \includegraphics[width=0.125\textwidth, height = 0.1\textwidth]{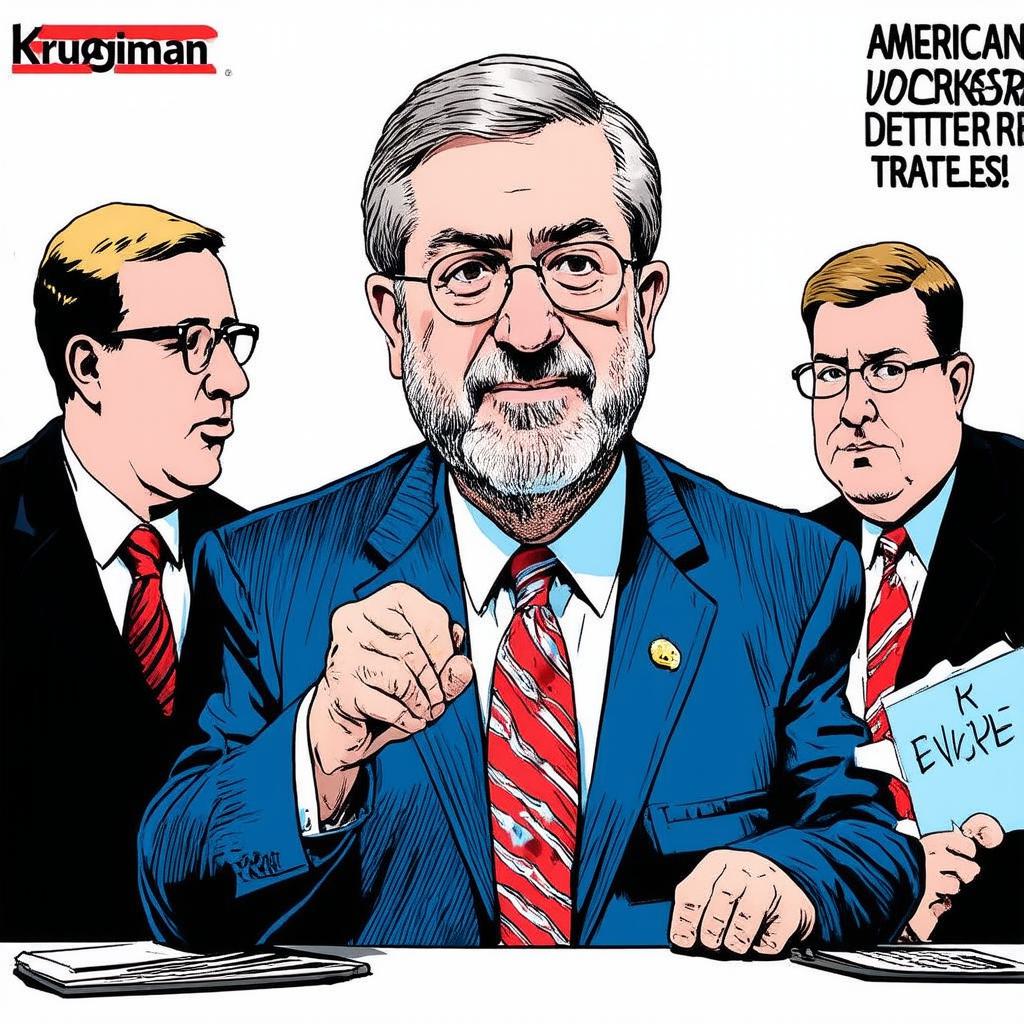} & \includegraphics[width=0.125\textwidth, height = 0.1\textwidth]{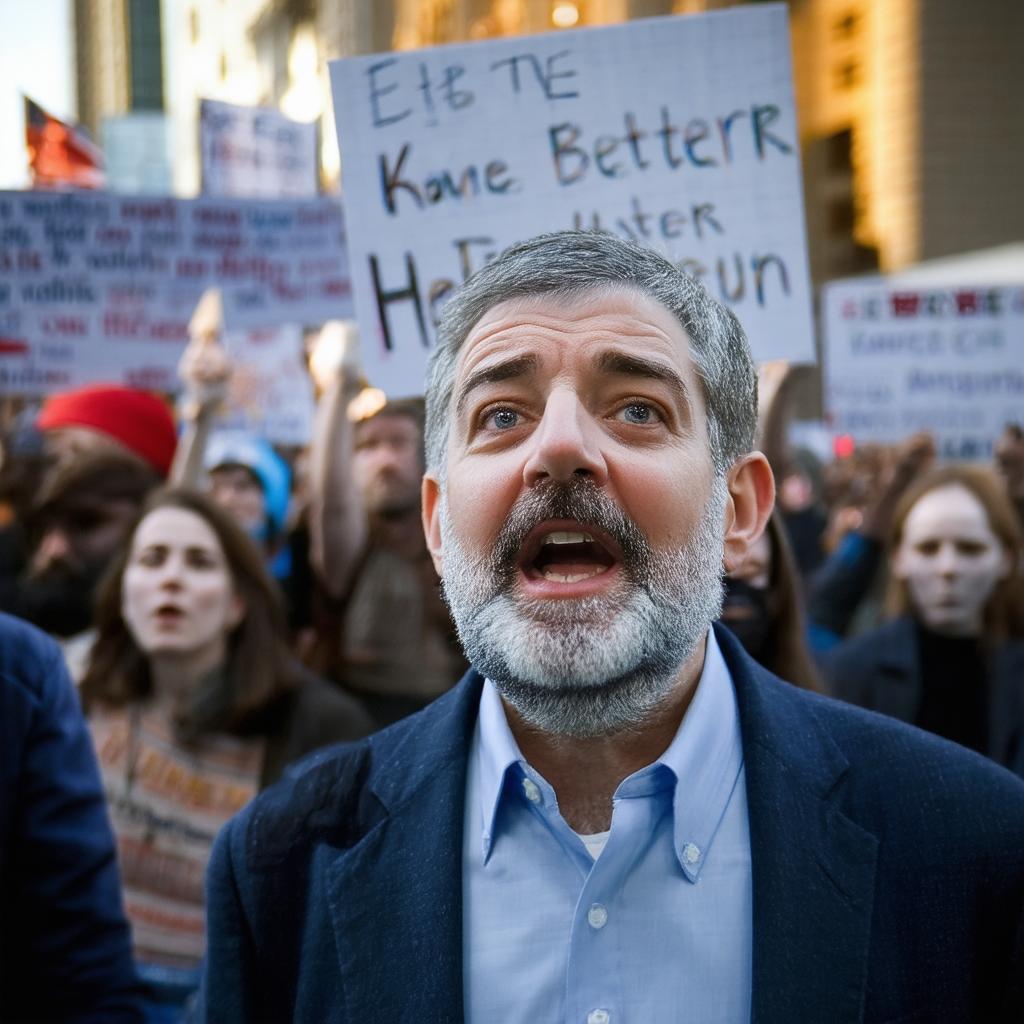} & \includegraphics[width=0.125\textwidth, height = 0.1\textwidth]{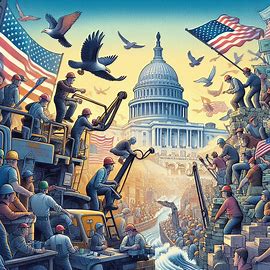} & \includegraphics[width=0.125\textwidth, height = 0.1\textwidth]{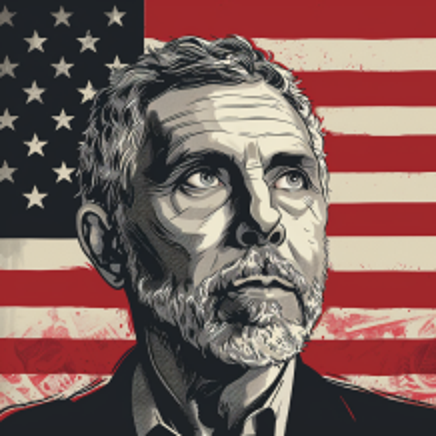} \\

\includegraphics[width=0.125\textwidth, height = 0.1\textwidth]{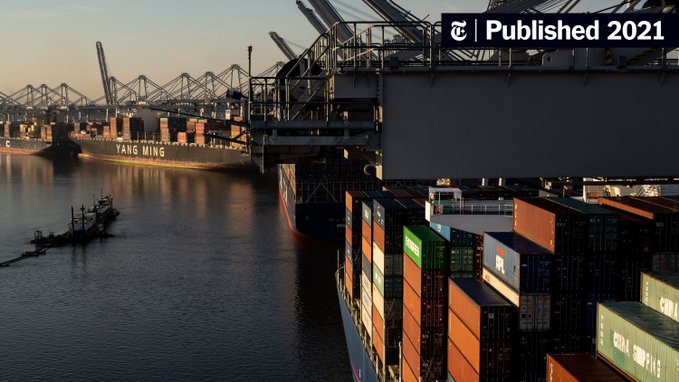} & \includegraphics[width=0.125\textwidth, height = 0.1\textwidth]{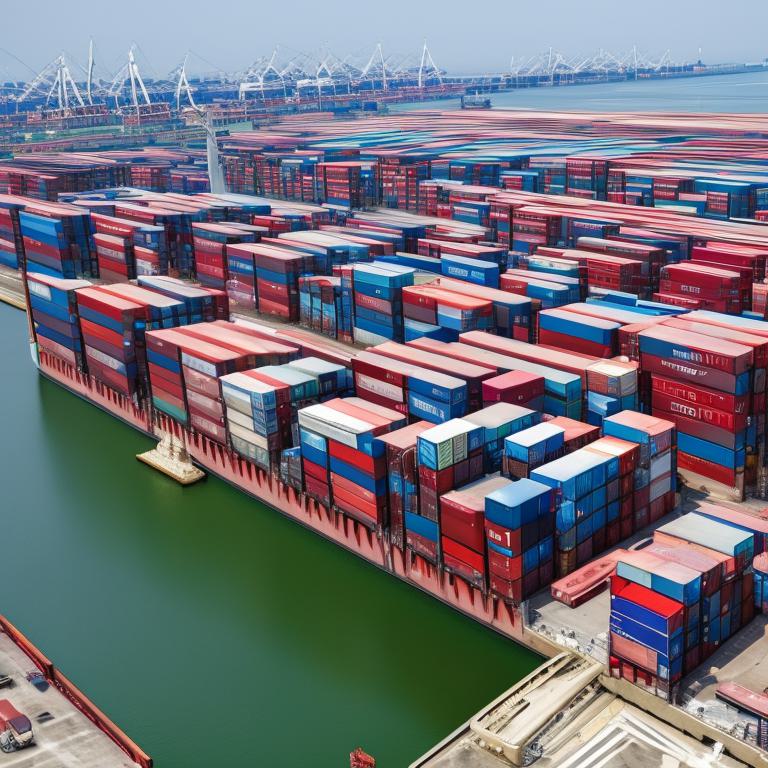} & \includegraphics[width=0.125\textwidth, height = 0.1\textwidth]{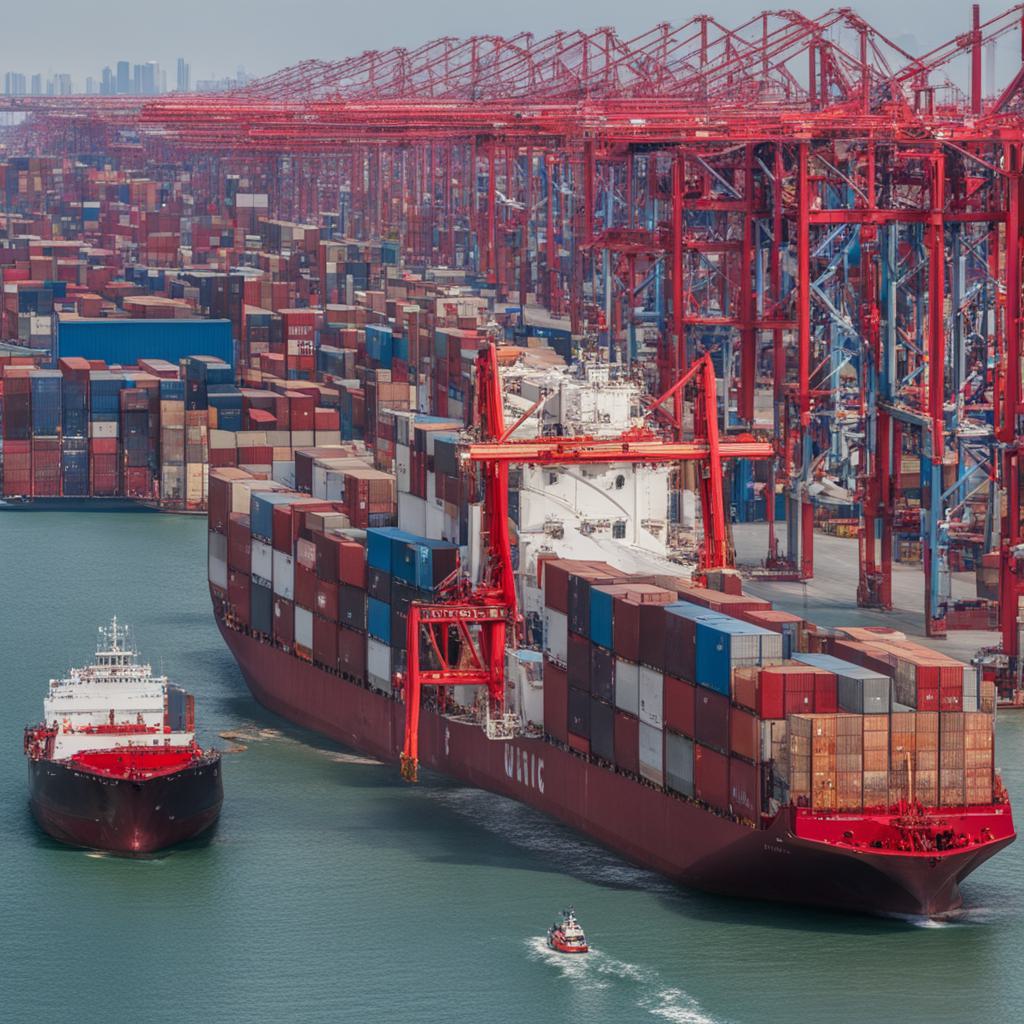} & \includegraphics[width=0.125\textwidth, height = 0.1\textwidth]{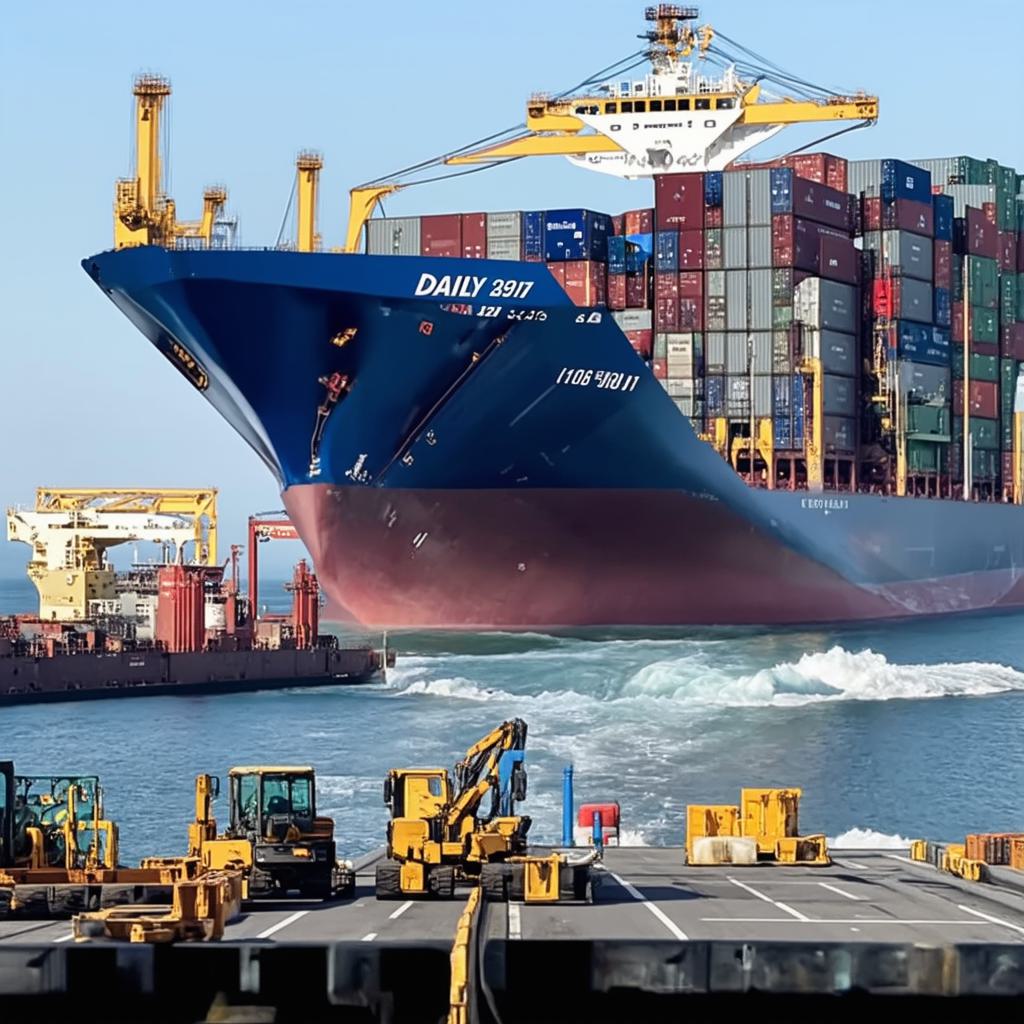} & \includegraphics[width=0.125\textwidth, height = 0.1\textwidth]{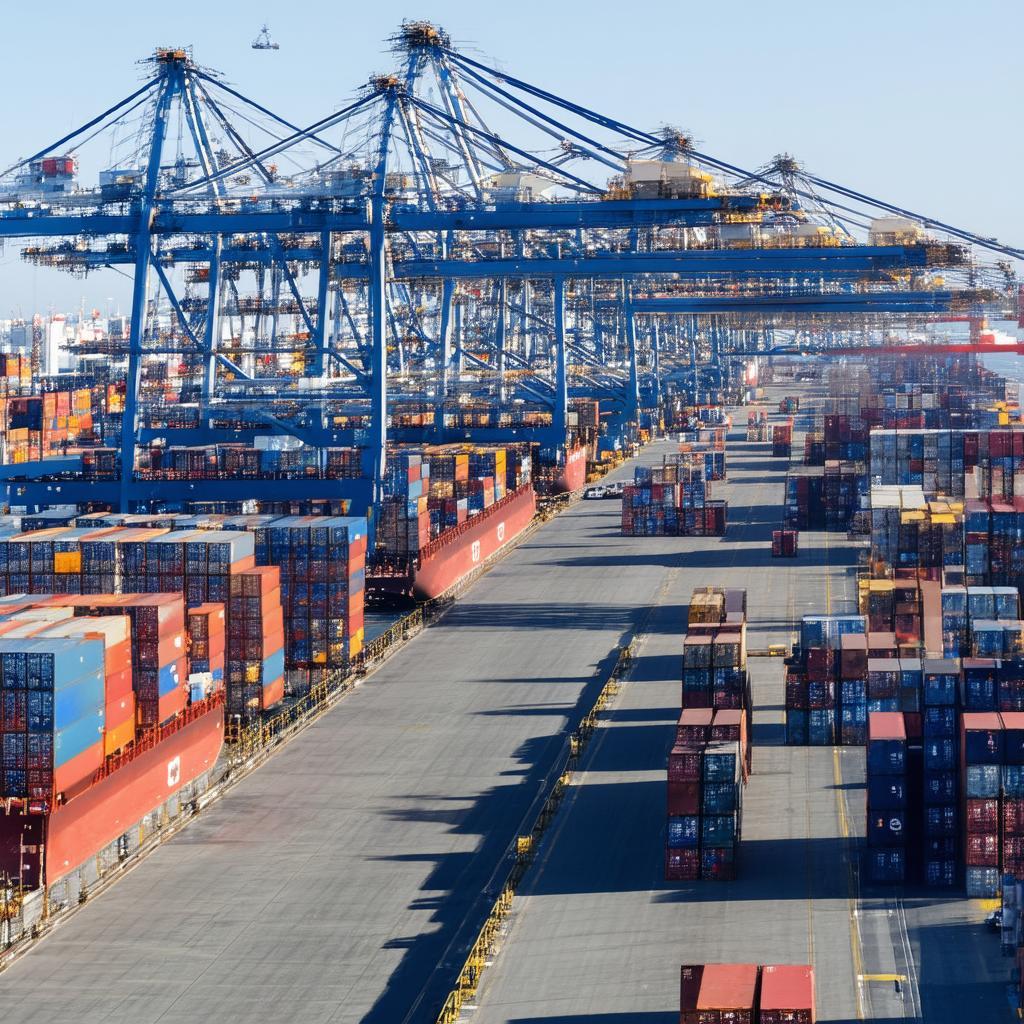} & \includegraphics[width=0.125\textwidth, height = 0.1\textwidth]{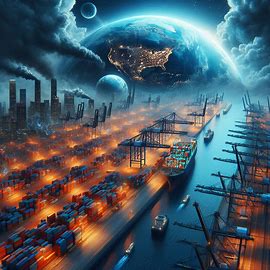} & \includegraphics[width=0.125\textwidth, height = 0.1\textwidth]{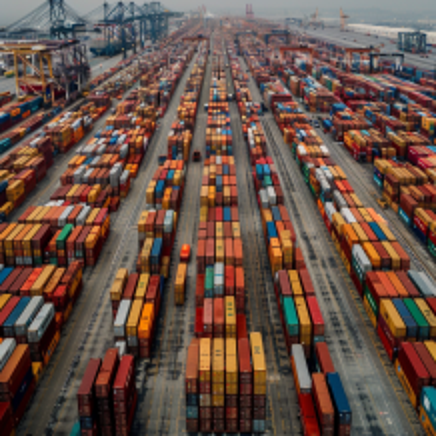} \\ 

\includegraphics[width=0.125\textwidth, height = 0.1\textwidth]{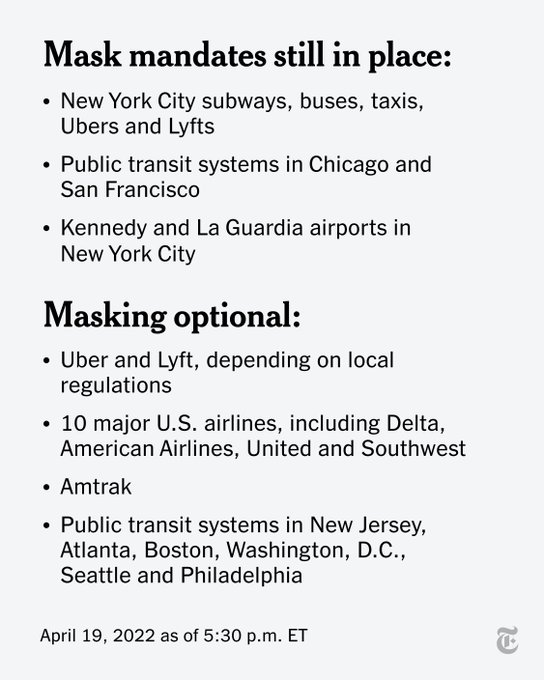} & \includegraphics[width=0.125\textwidth, height = 0.1\textwidth]{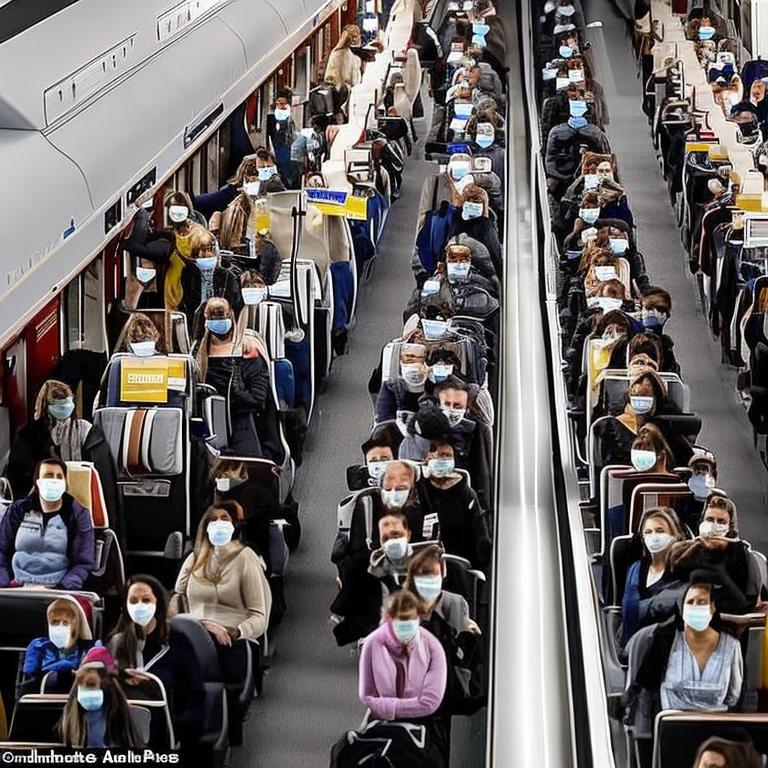} & \includegraphics[width=0.125\textwidth, height = 0.1\textwidth]{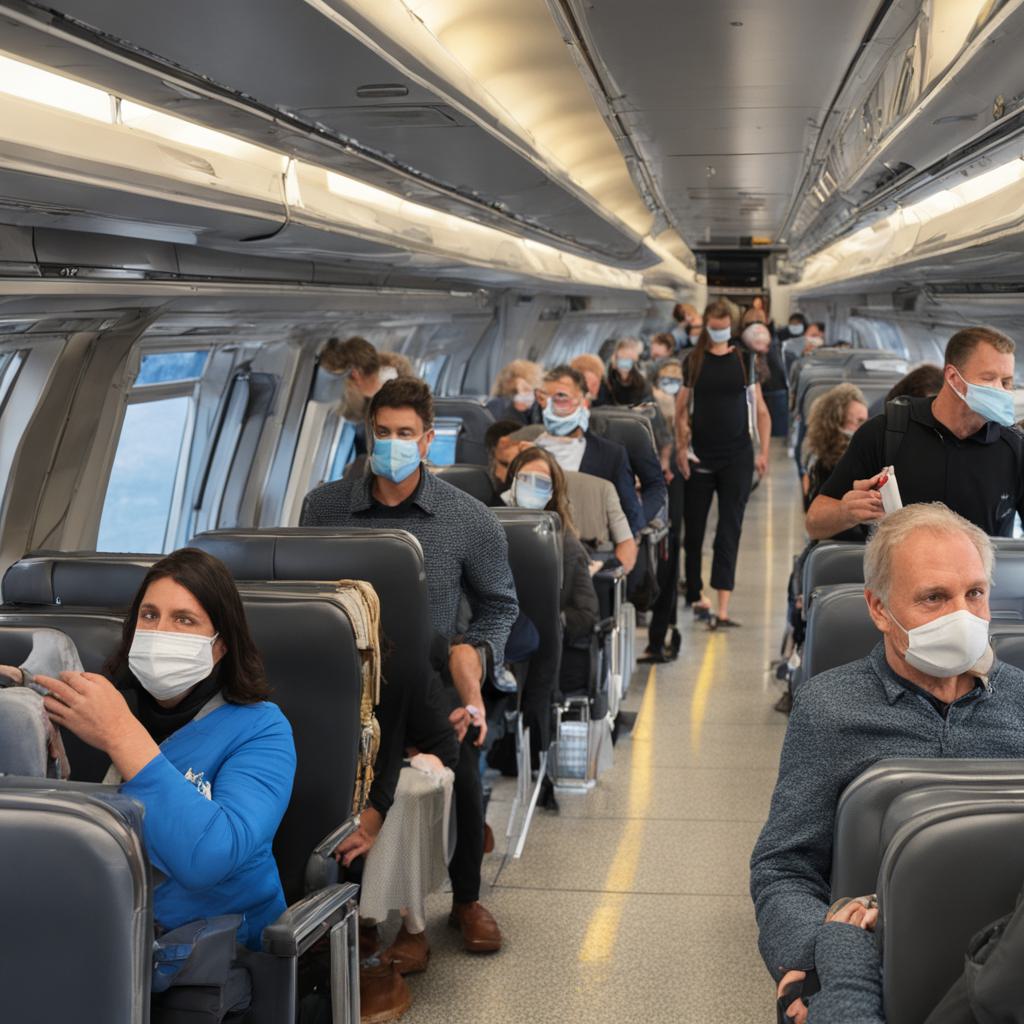} & \includegraphics[width=0.125\textwidth, height = 0.1\textwidth]{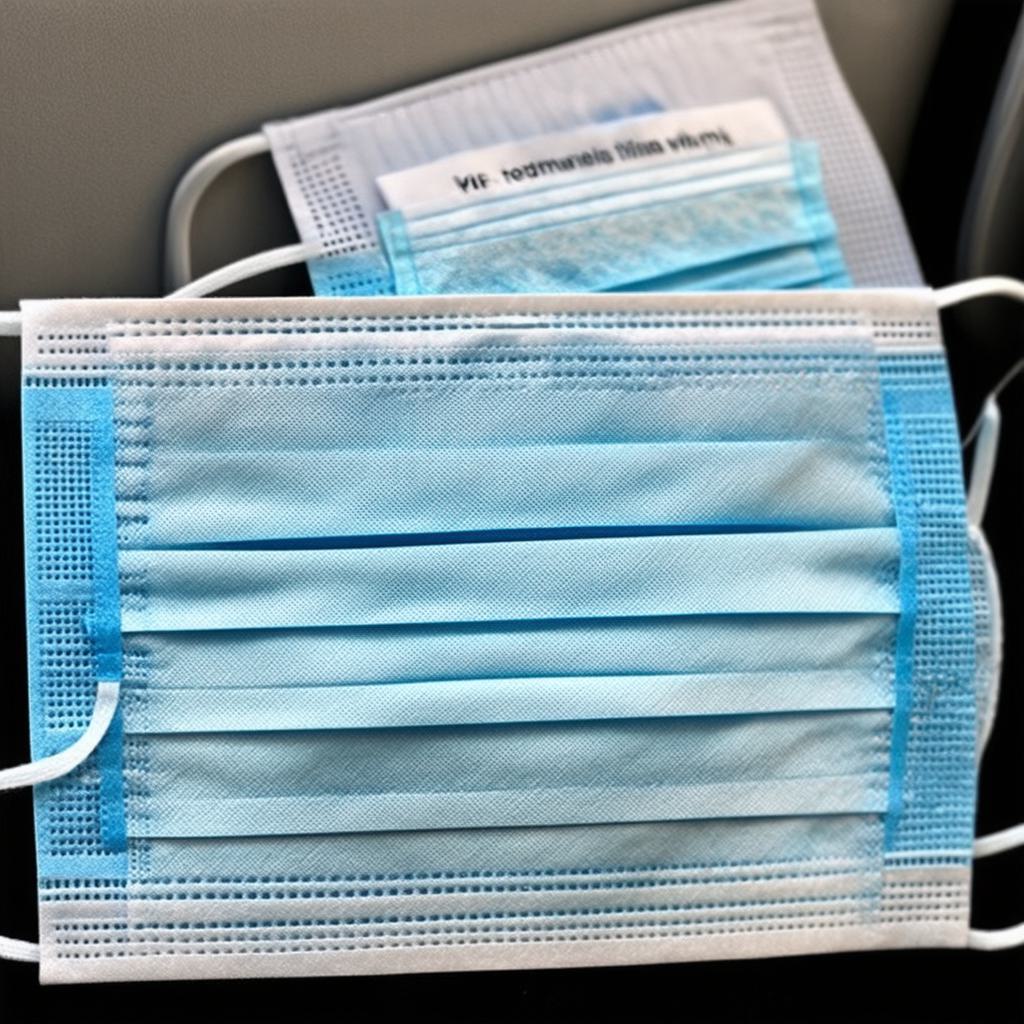} & \includegraphics[width=0.125\textwidth, height = 0.1\textwidth]{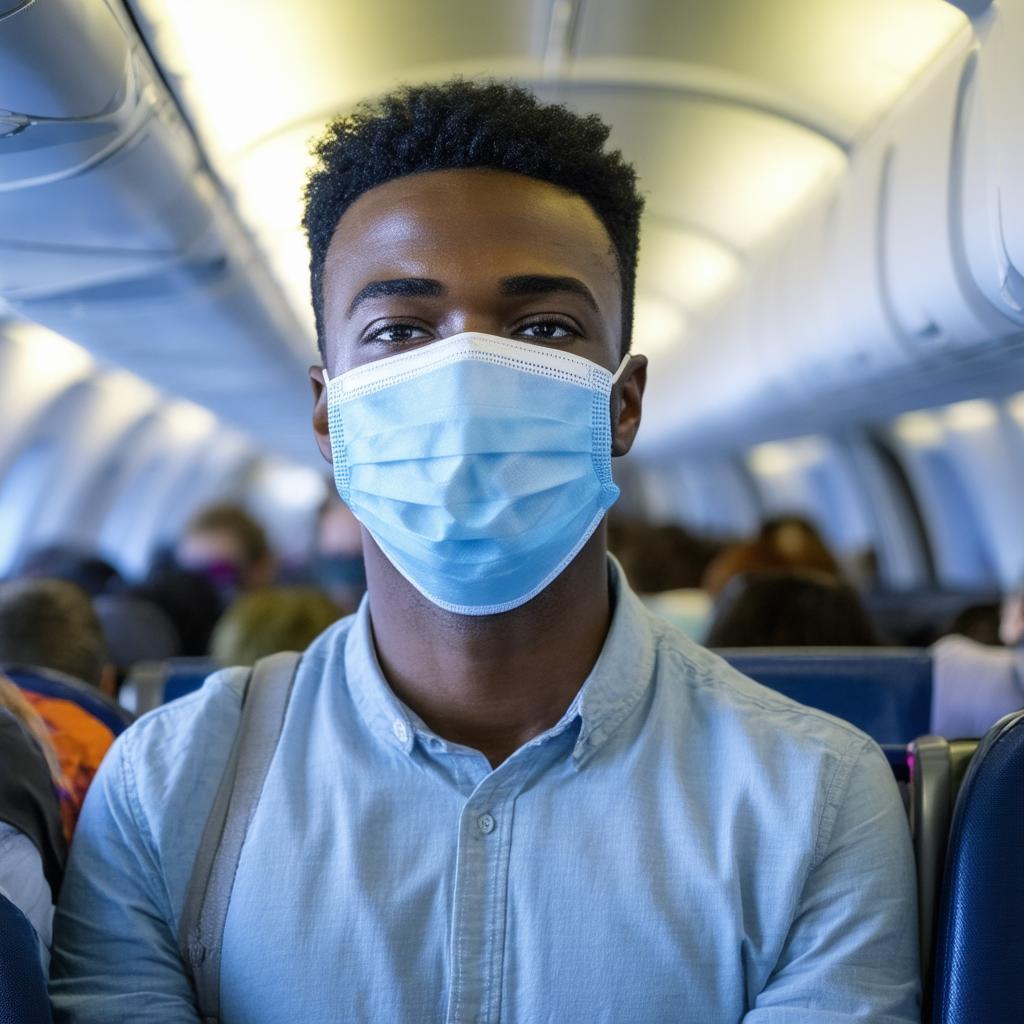} & \includegraphics[width=0.125\textwidth, height = 0.1\textwidth]{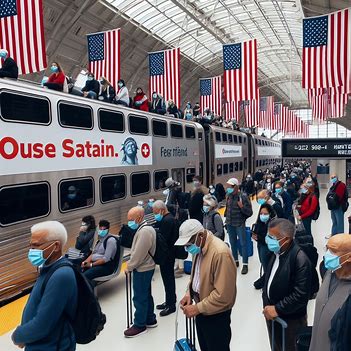} & \includegraphics[width=0.125\textwidth, height = 0.1\textwidth]{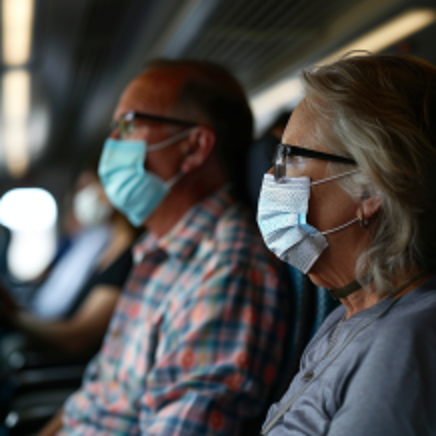} \\ 

\includegraphics[width=0.125\textwidth, height = 0.1\textwidth]{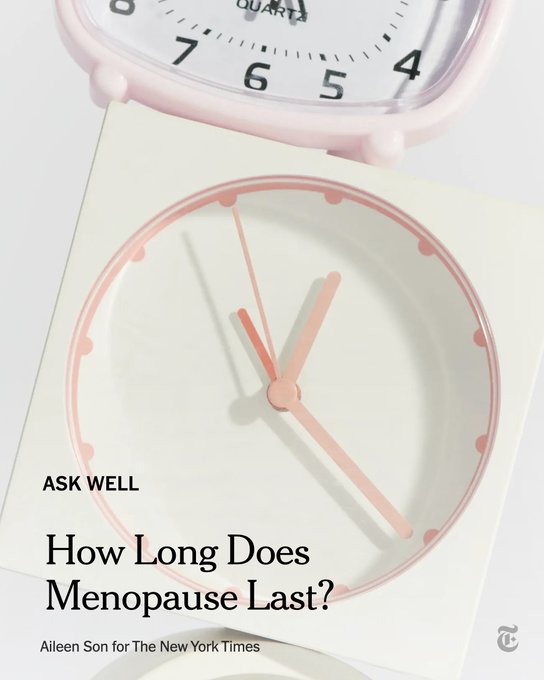} & \includegraphics[width=0.125\textwidth, height = 0.1\textwidth]{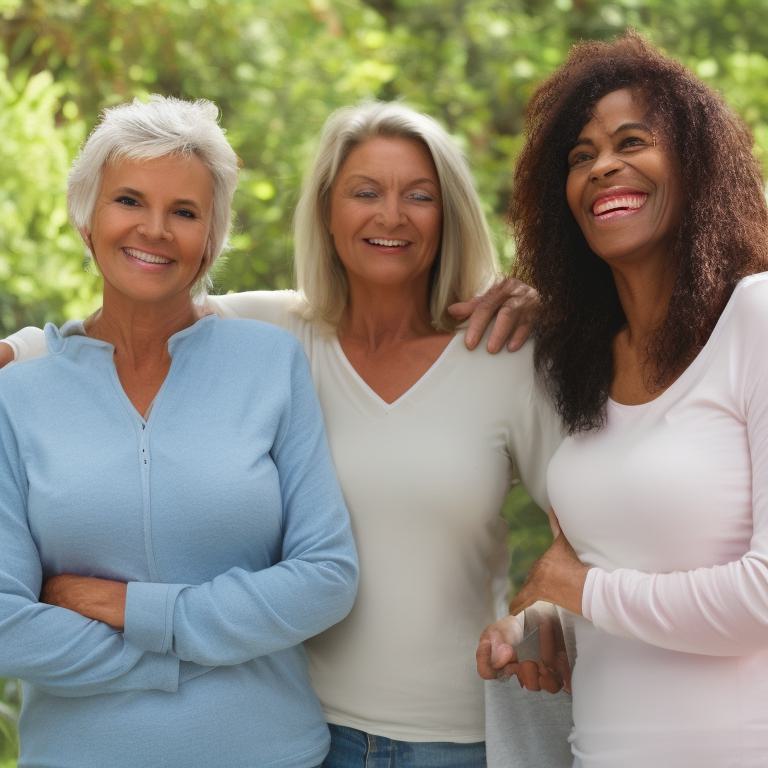} & \includegraphics[width=0.125\textwidth, height = 0.1\textwidth]{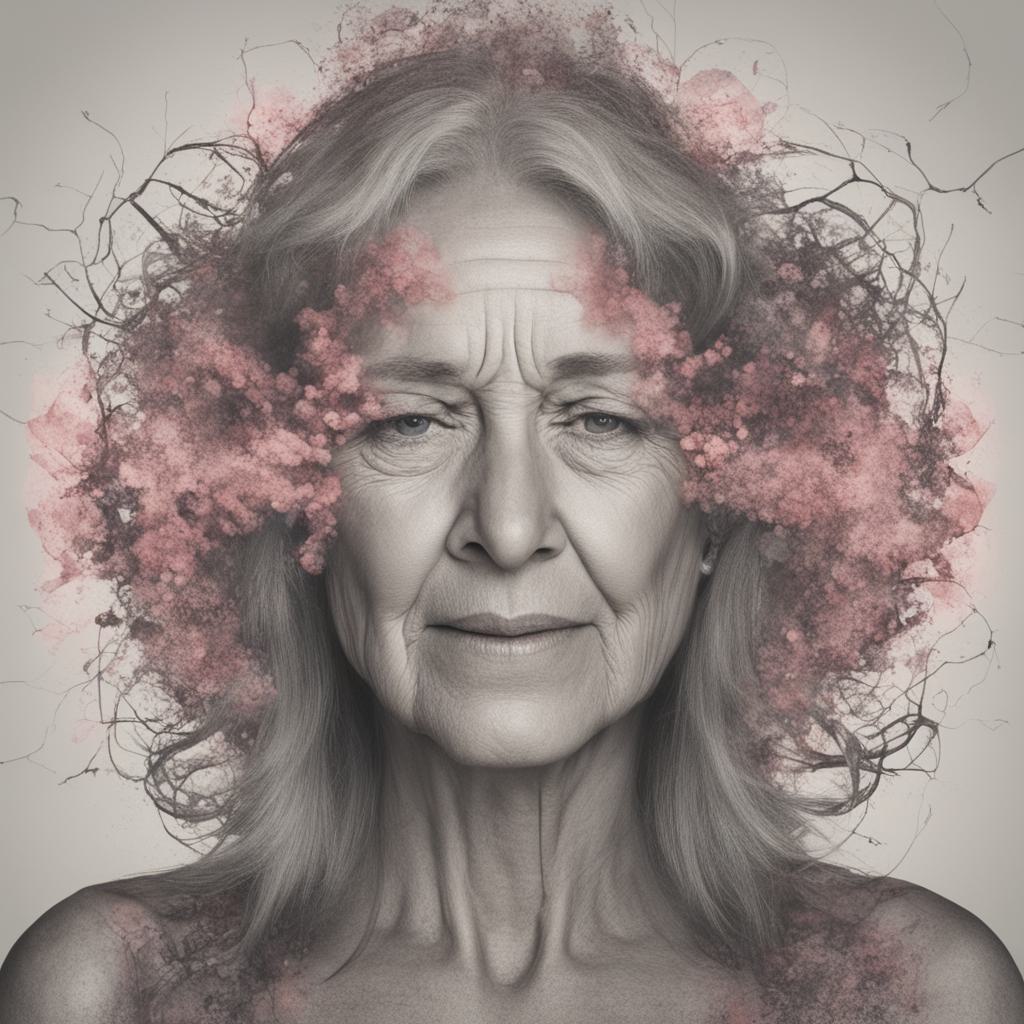} & \includegraphics[width=0.125\textwidth, height = 0.1\textwidth]{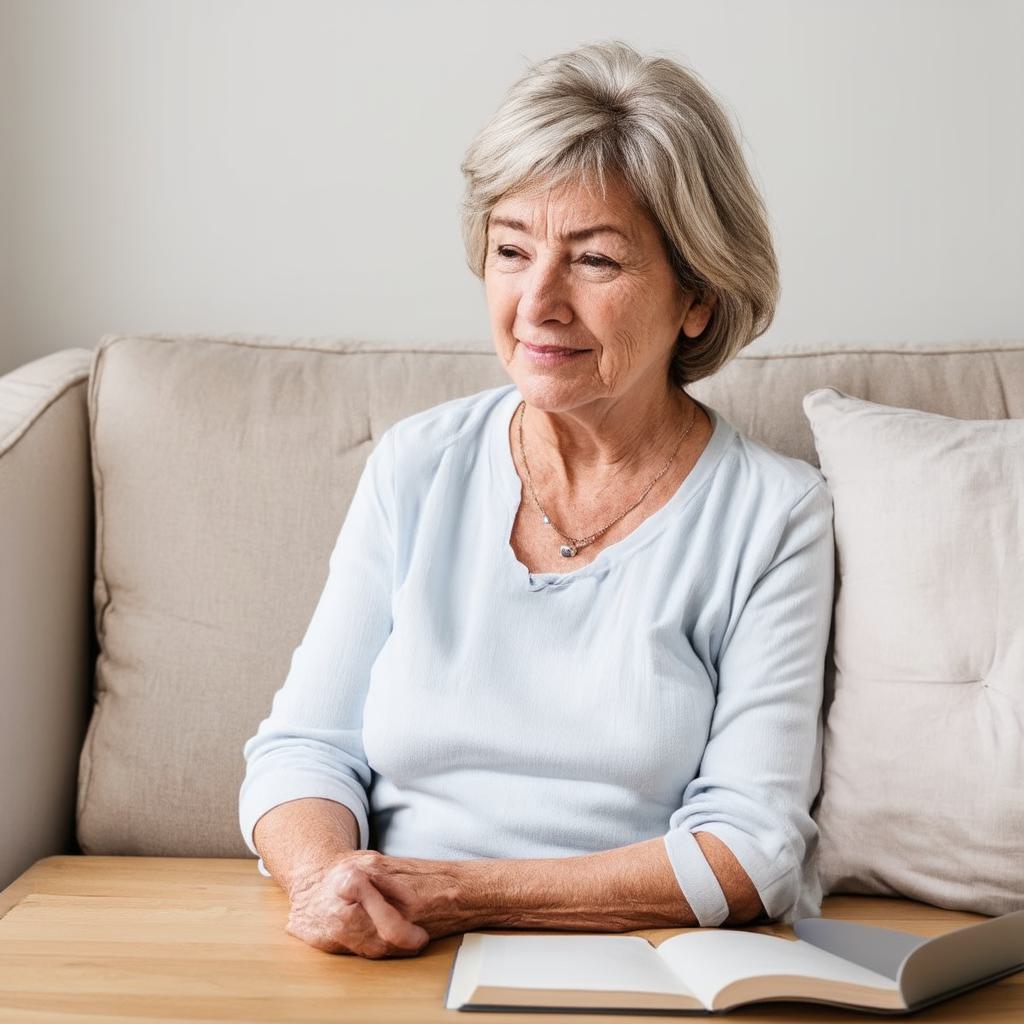} & \includegraphics[width=0.125\textwidth, height = 0.1\textwidth]{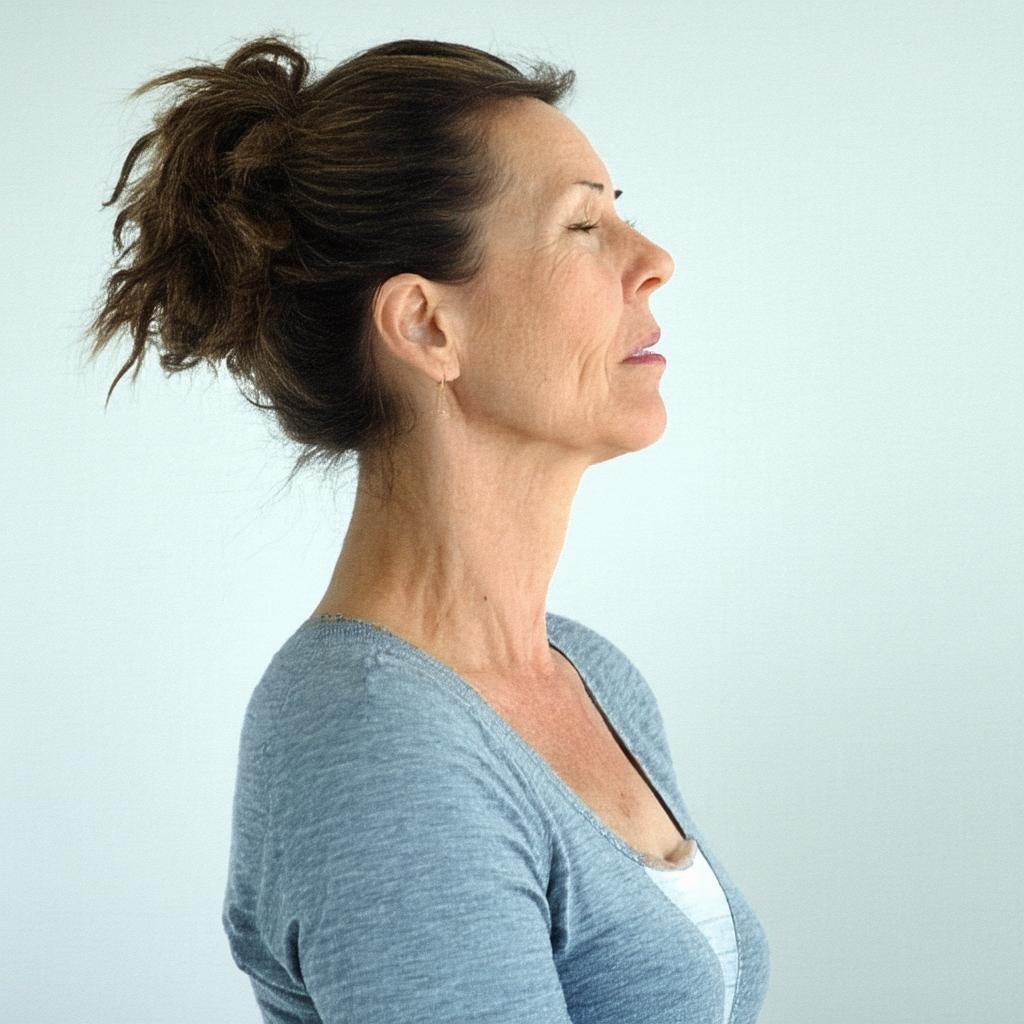} & \includegraphics[width=0.125\textwidth, height = 0.1\textwidth]{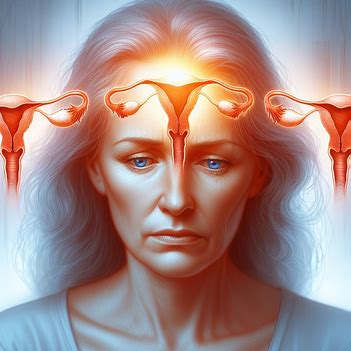} & \includegraphics[width=0.125\textwidth, height = 0.1\textwidth]{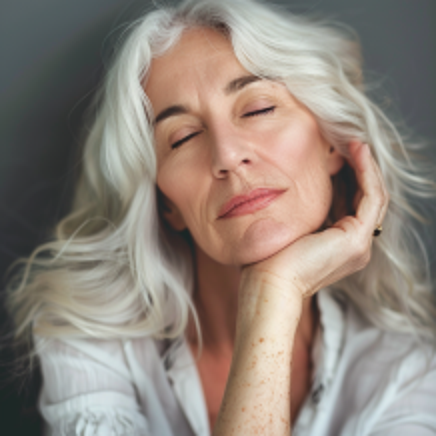} \\ 

\includegraphics[width=0.125\textwidth, height = 0.1\textwidth]{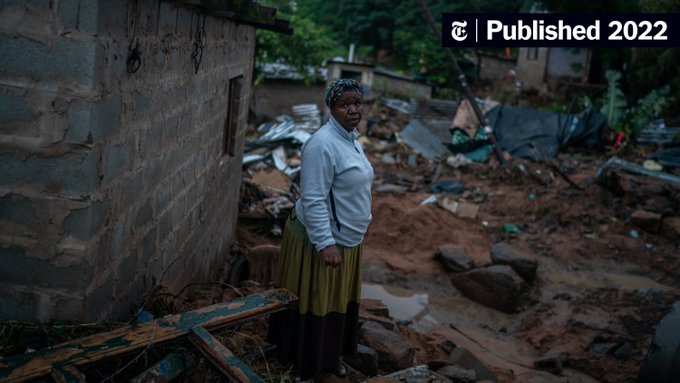} & \includegraphics[width=0.125\textwidth, height = 0.1\textwidth]{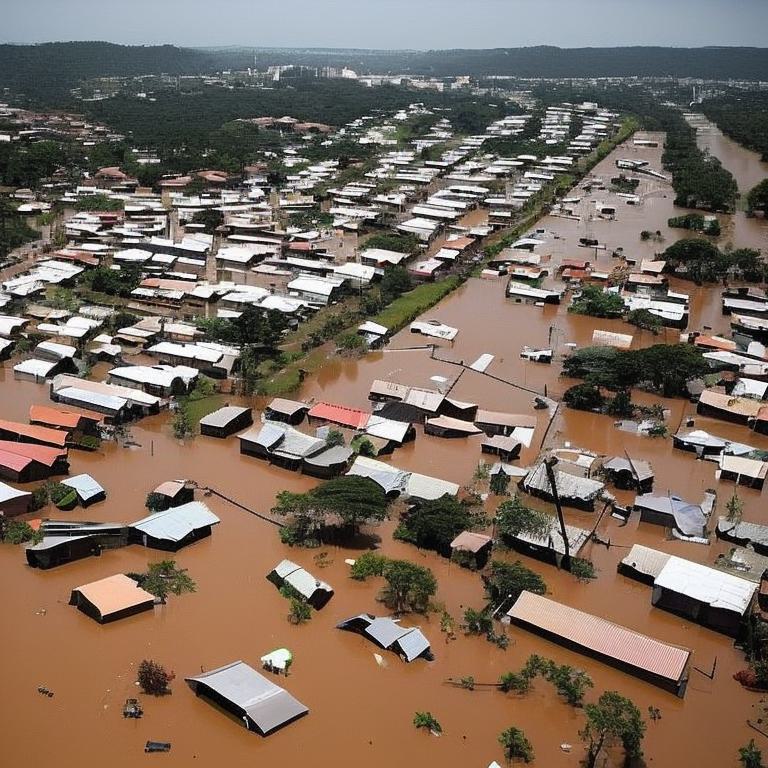} & \includegraphics[width=0.125\textwidth, height = 0.1\textwidth]{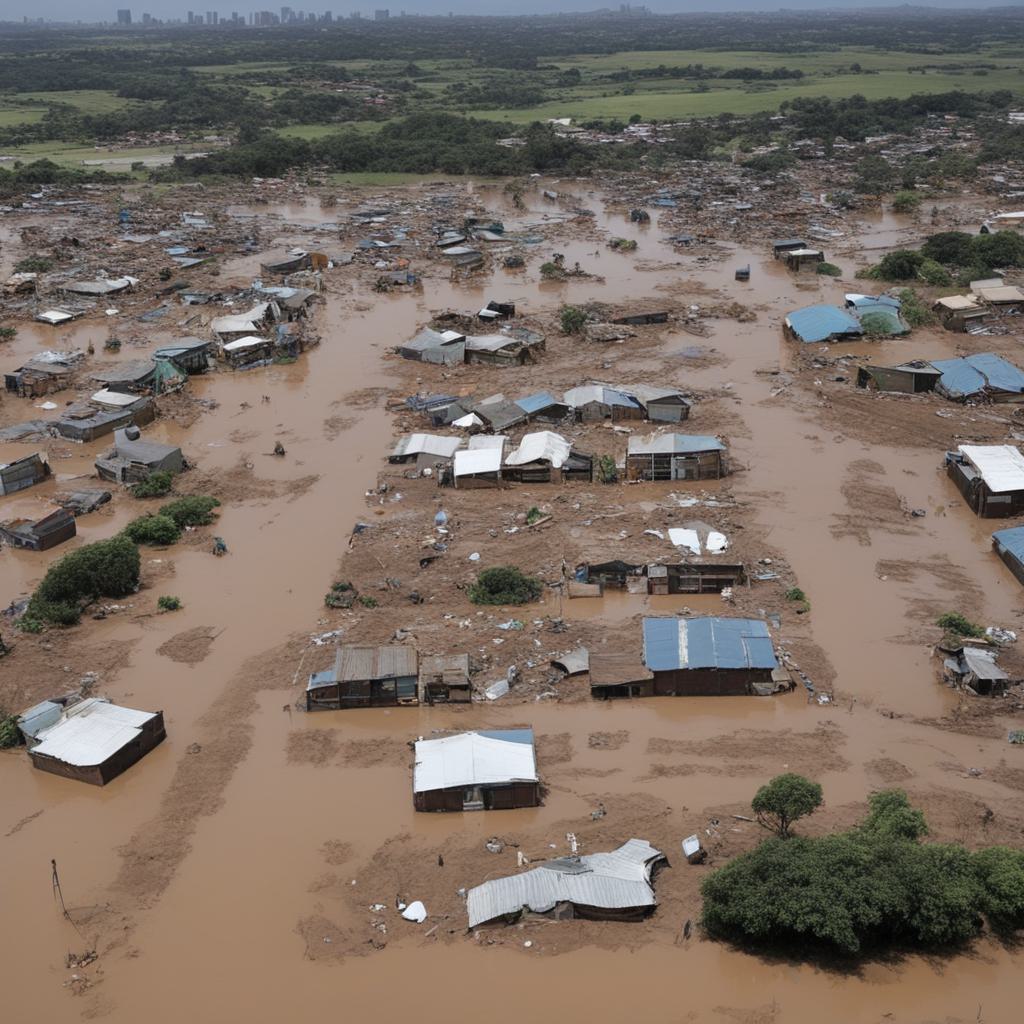} & \includegraphics[width=0.125\textwidth, height = 0.1\textwidth]{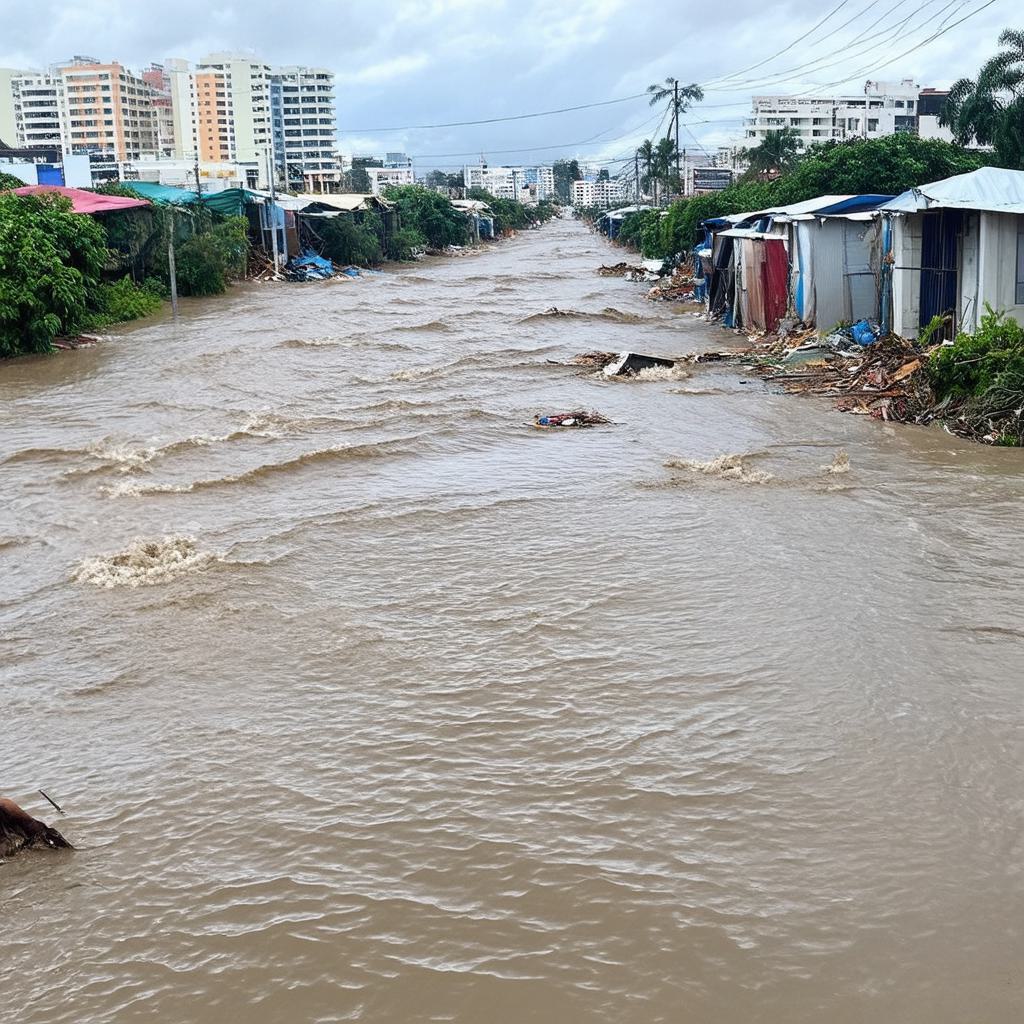} & \includegraphics[width=0.125\textwidth, height = 0.1\textwidth]{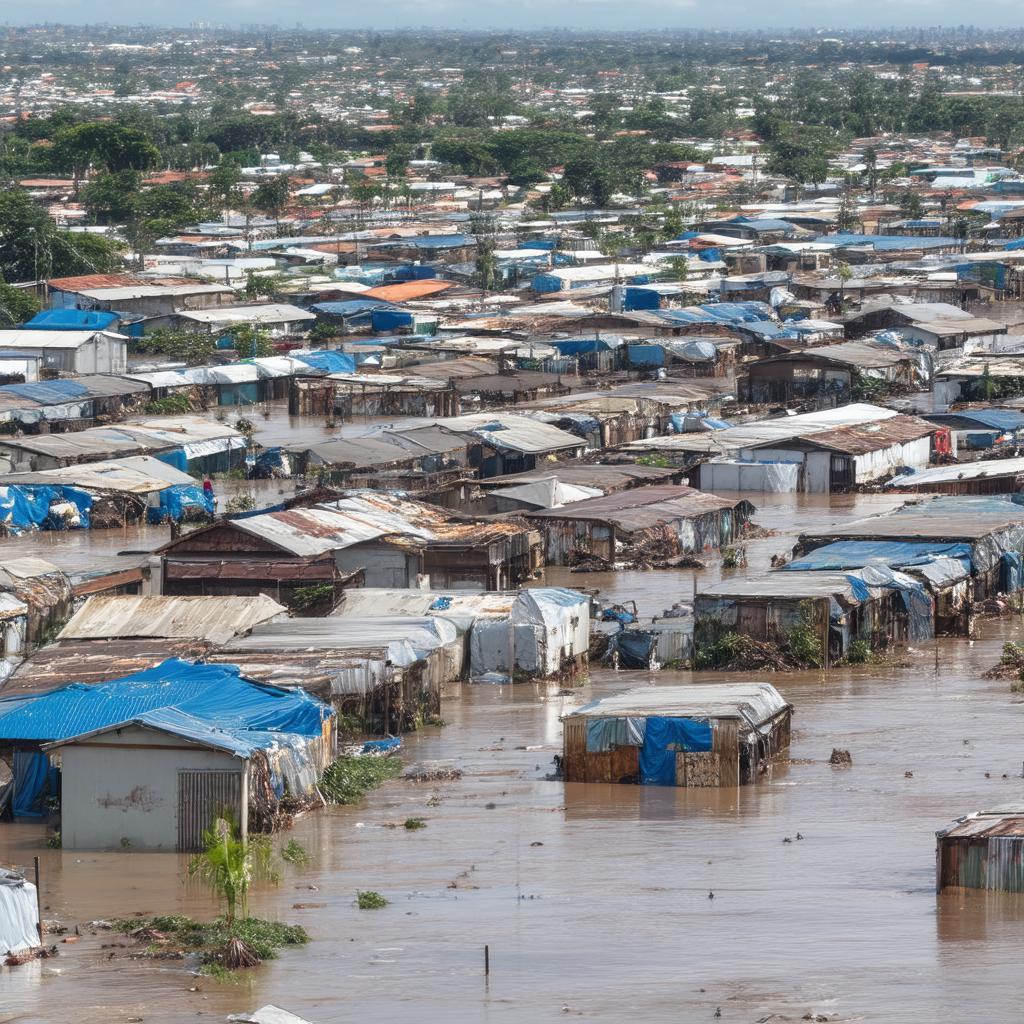} & \includegraphics[width=0.125\textwidth, height = 0.1\textwidth]{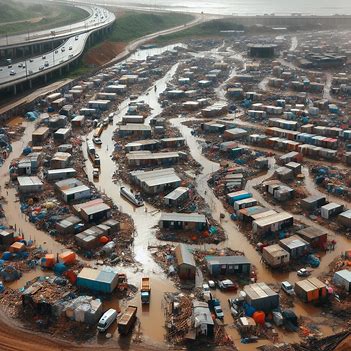} & \includegraphics[width=0.125\textwidth, height = 0.1\textwidth]{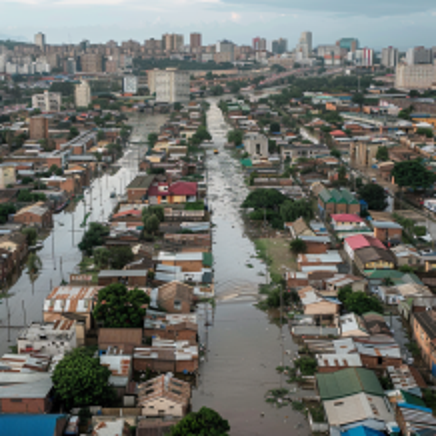} \\ 

\includegraphics[width=0.125\textwidth, height = 0.1\textwidth]{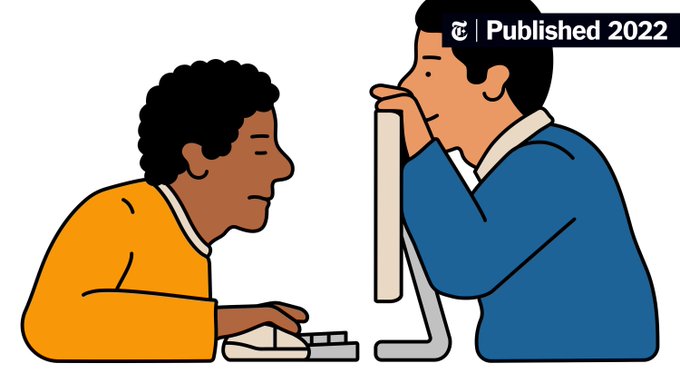} & \includegraphics[width=0.125\textwidth, height = 0.1\textwidth]{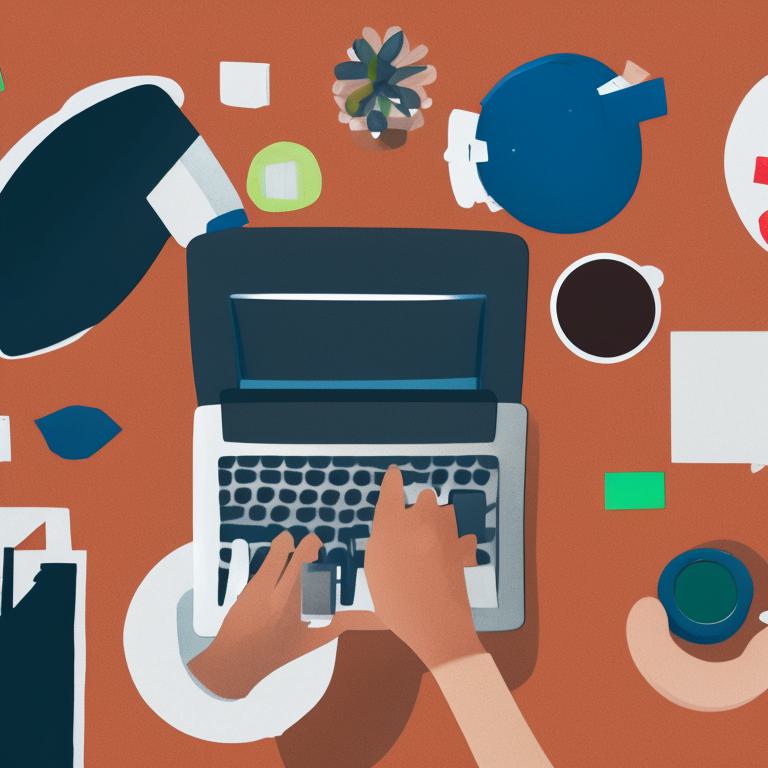} & \includegraphics[width=0.125\textwidth, height = 0.1\textwidth]{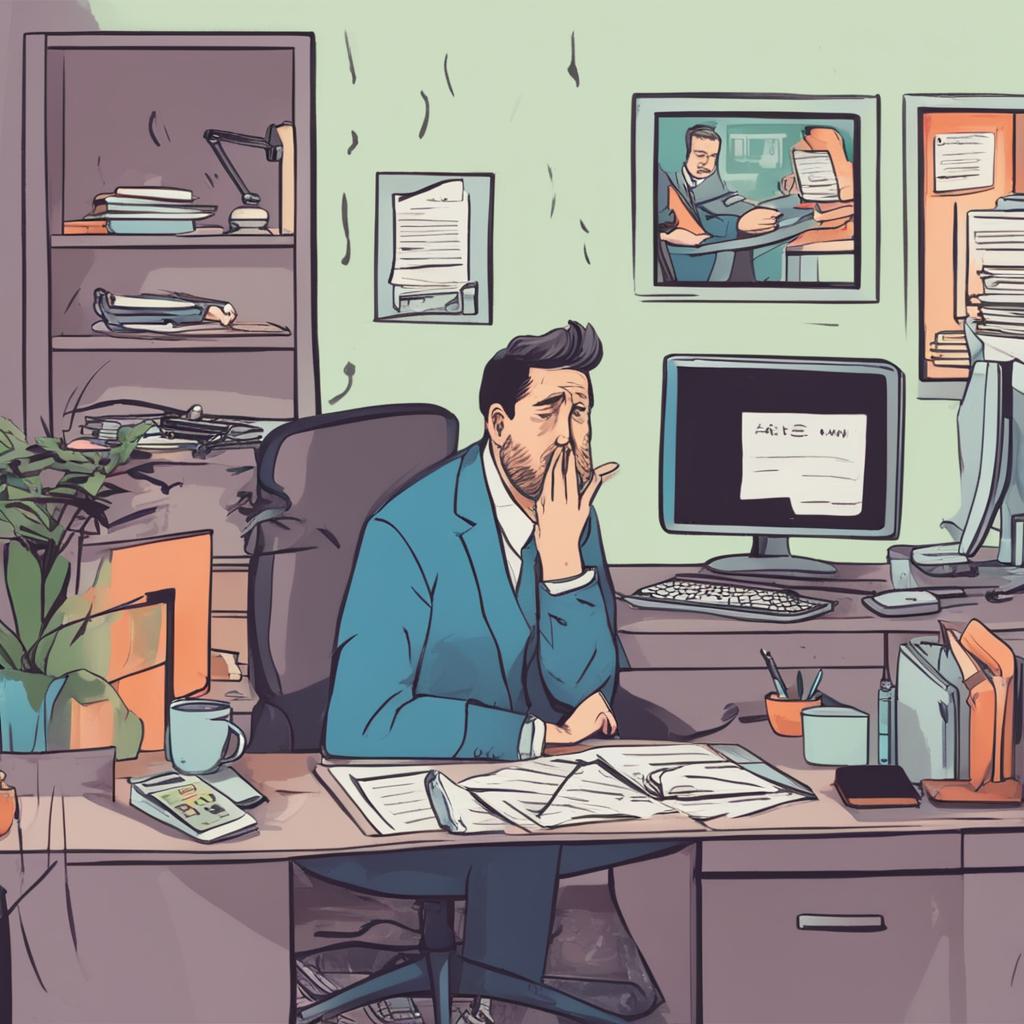} & \includegraphics[width=0.125\textwidth, height = 0.1\textwidth]{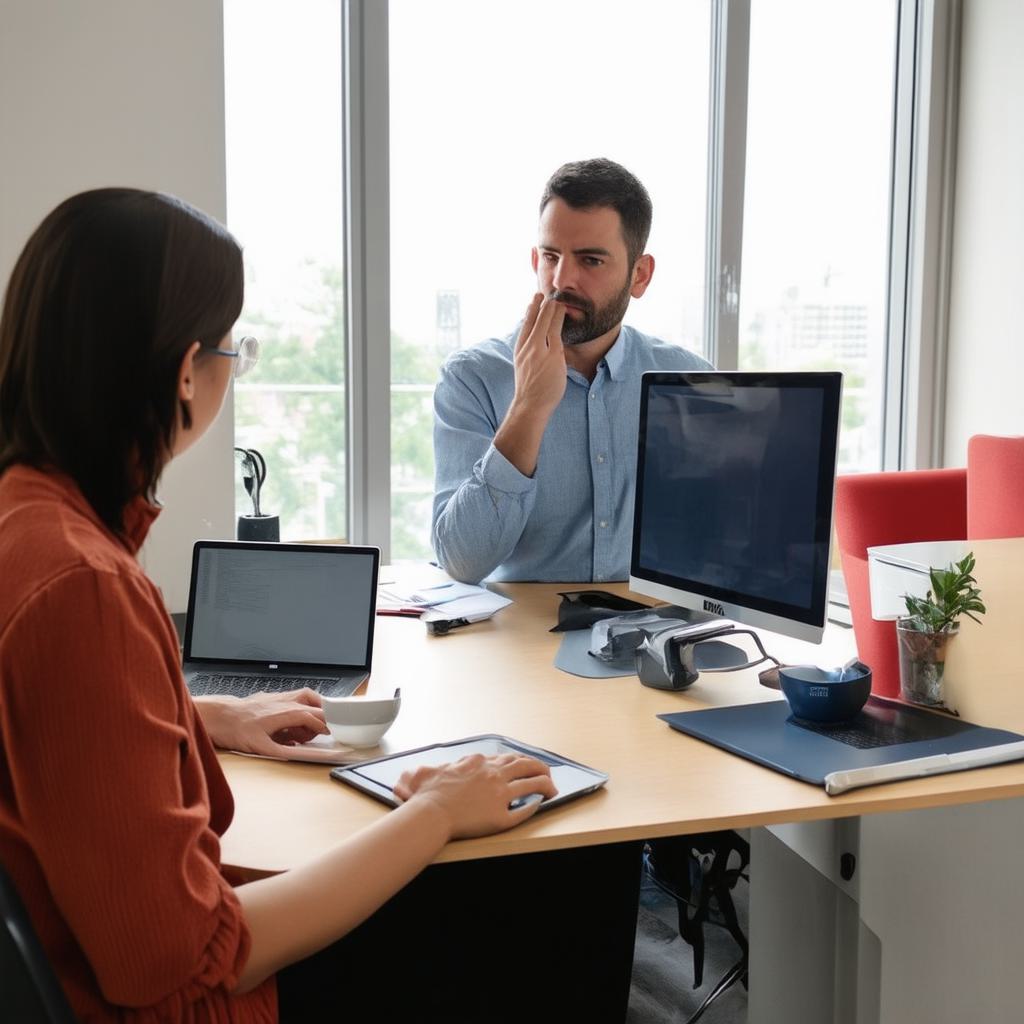} & \includegraphics[width=0.125\textwidth, height = 0.1\textwidth]{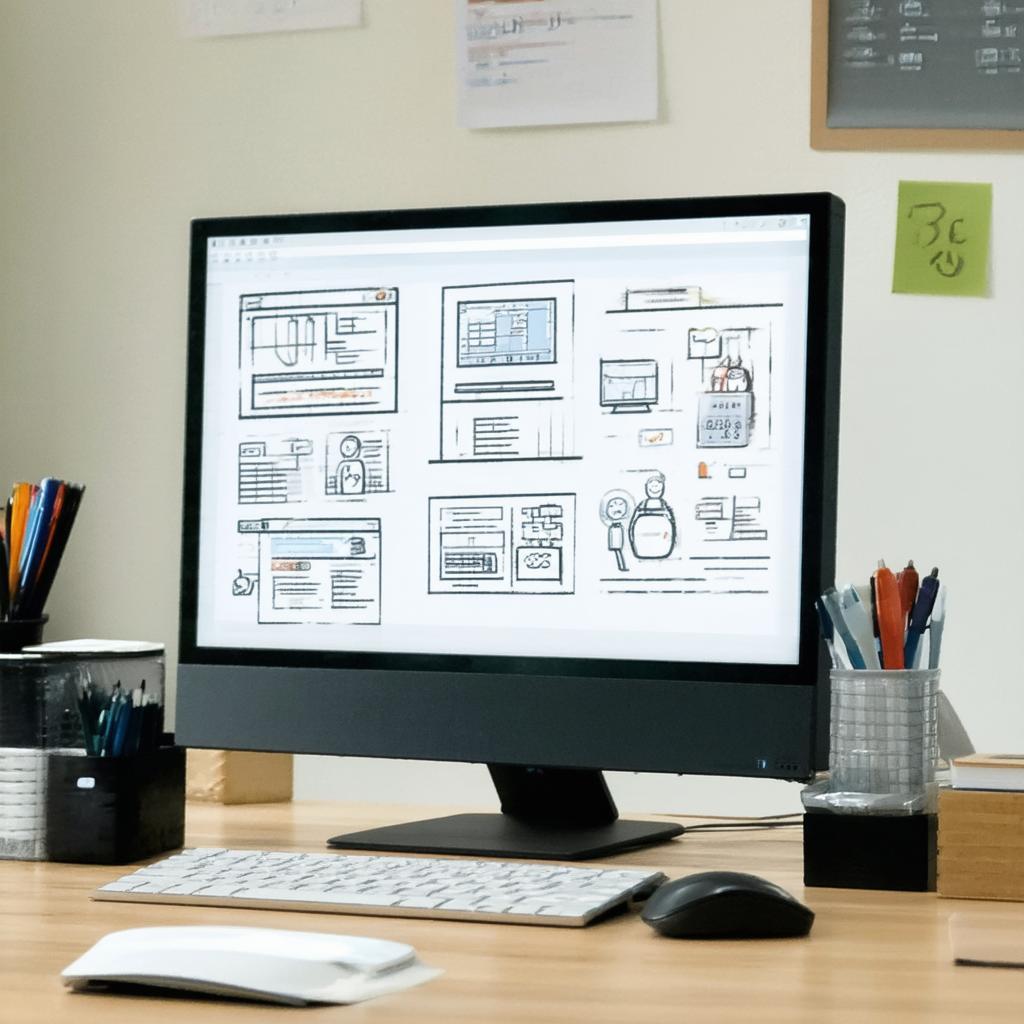} & \includegraphics[width=0.125\textwidth, height = 0.1\textwidth]{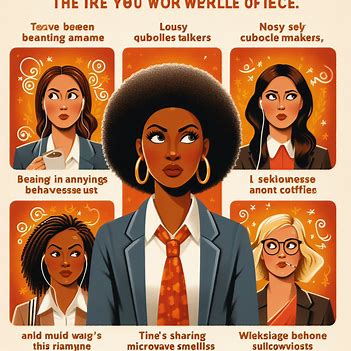} & \includegraphics[width=0.125\textwidth, height = 0.1\textwidth]{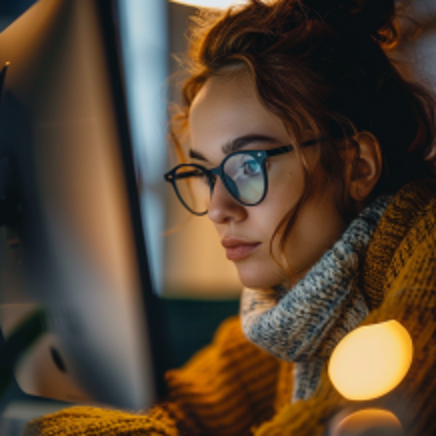} \\ 

\includegraphics[width=0.125\textwidth, height = 0.1\textwidth]{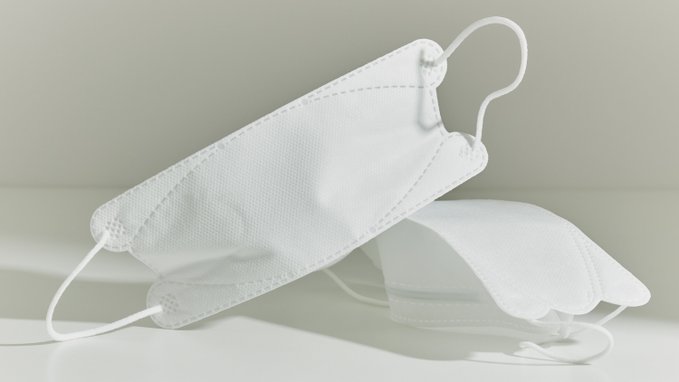} & \includegraphics[width=0.125\textwidth, height = 0.1\textwidth]{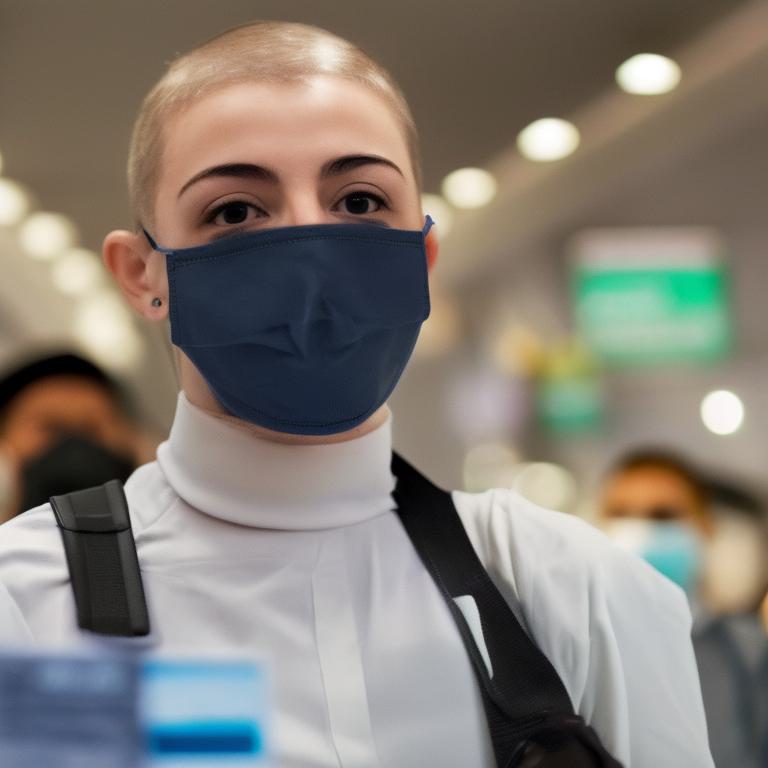} & \includegraphics[width=0.125\textwidth, height = 0.1\textwidth]{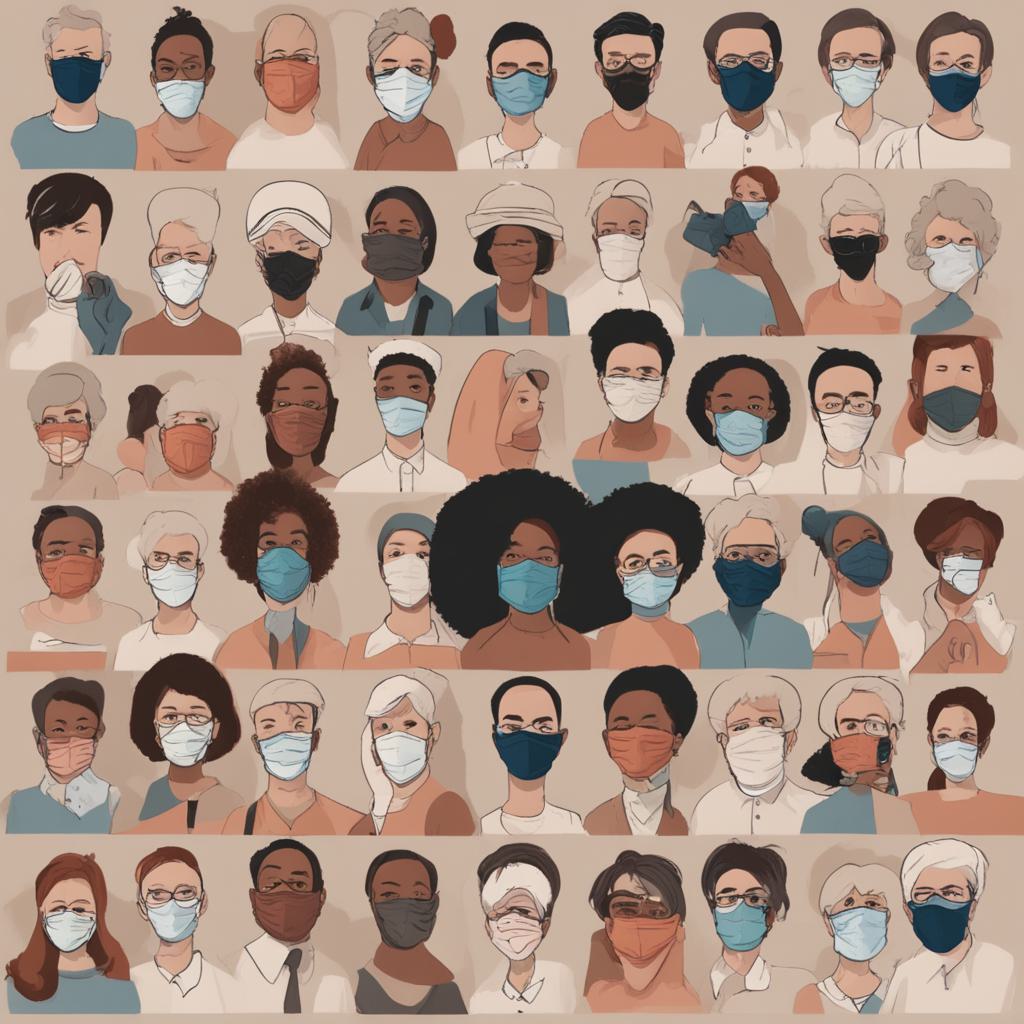} & \includegraphics[width=0.125\textwidth, height = 0.1\textwidth]{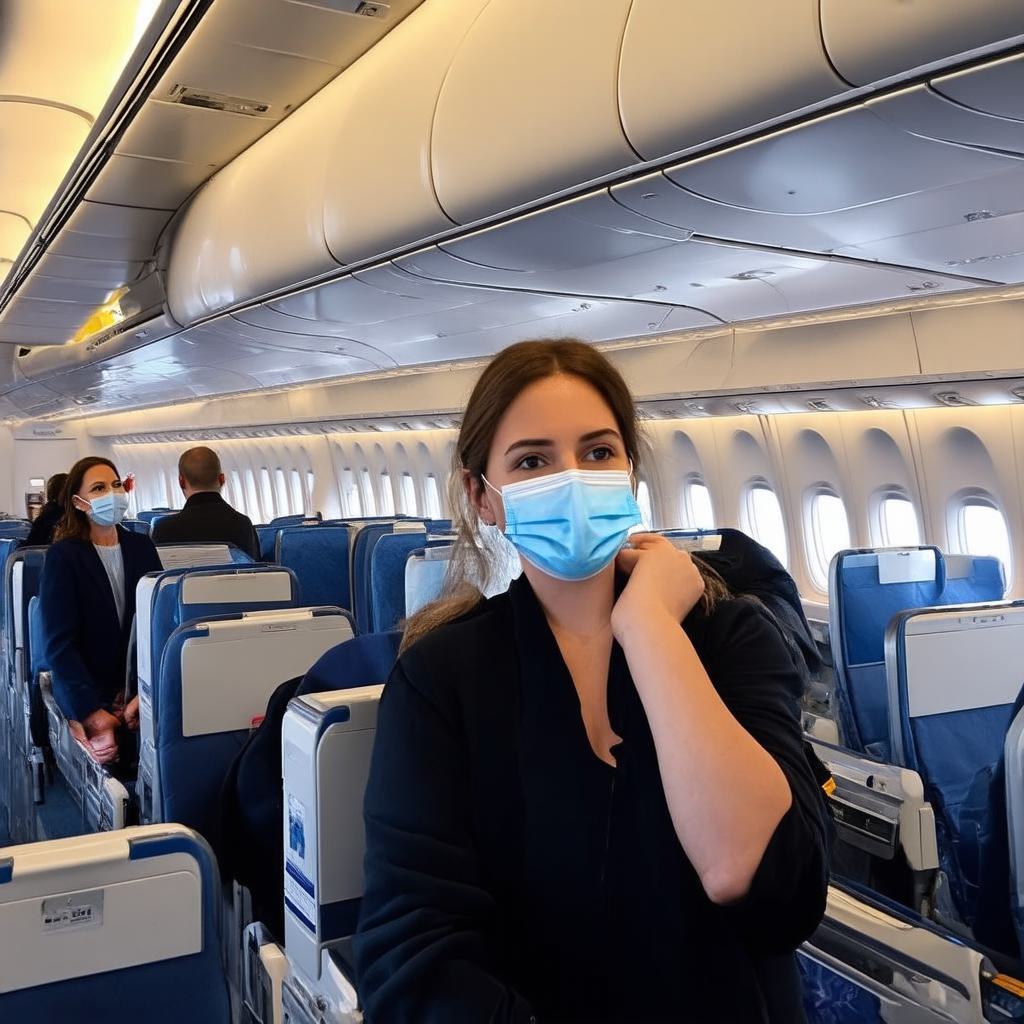} & \includegraphics[width=0.125\textwidth, height = 0.1\textwidth]{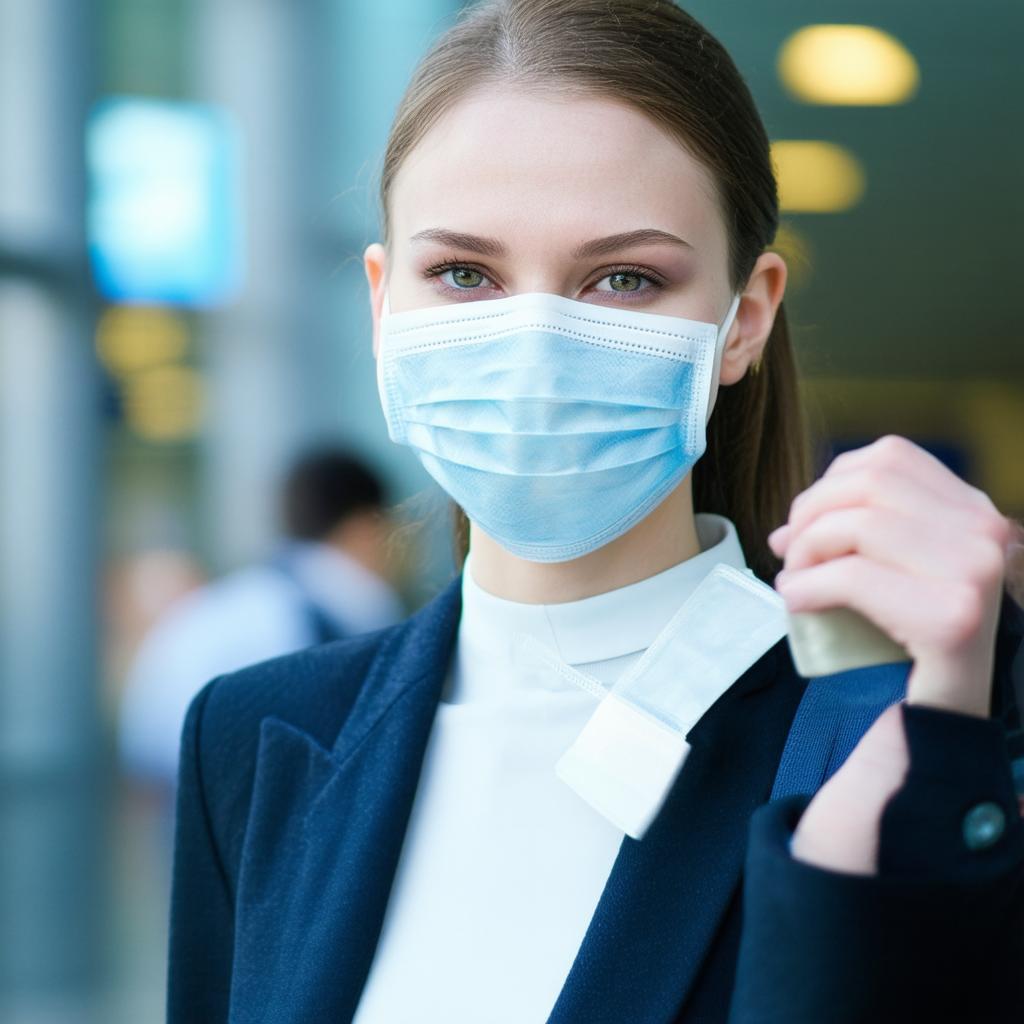} & \includegraphics[width=0.125\textwidth, height = 0.1\textwidth]{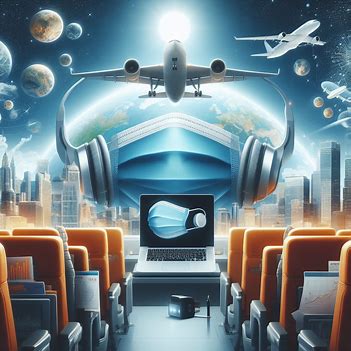} & \includegraphics[width=0.125\textwidth, height = 0.1\textwidth]{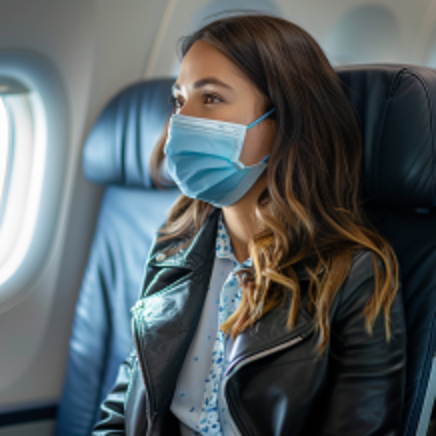} \\ \hline
\end{tabular}%
}
\end{table*}

\section{Factor-level analysis} 
\label{sec:vai}

\subsection{LBP Texture Analysis}
\label{lbp}
\begin{figure}[htbp]
    \centering

    \begin{minipage}{0.23\textwidth}
        \centering
        \includegraphics[width=\linewidth]{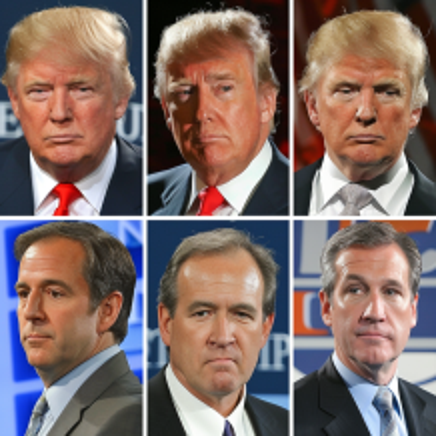}
        \caption*{Midjourney 6}
        \label{fig:midjourney-image}
    \end{minipage}
    \hfill
    \begin{minipage}{0.23\textwidth}
        \centering
        \includegraphics[width=\linewidth]{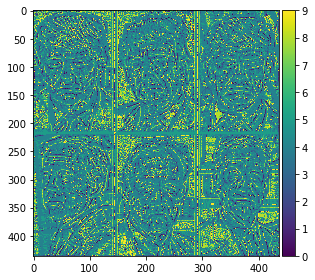}
        \caption*{LBP of Midjourney 6}
        \label{fig:midjourney-lbp}
    \end{minipage}
    \hfill
    \begin{minipage}{0.23\textwidth}
        \centering
        \includegraphics[width=\linewidth]{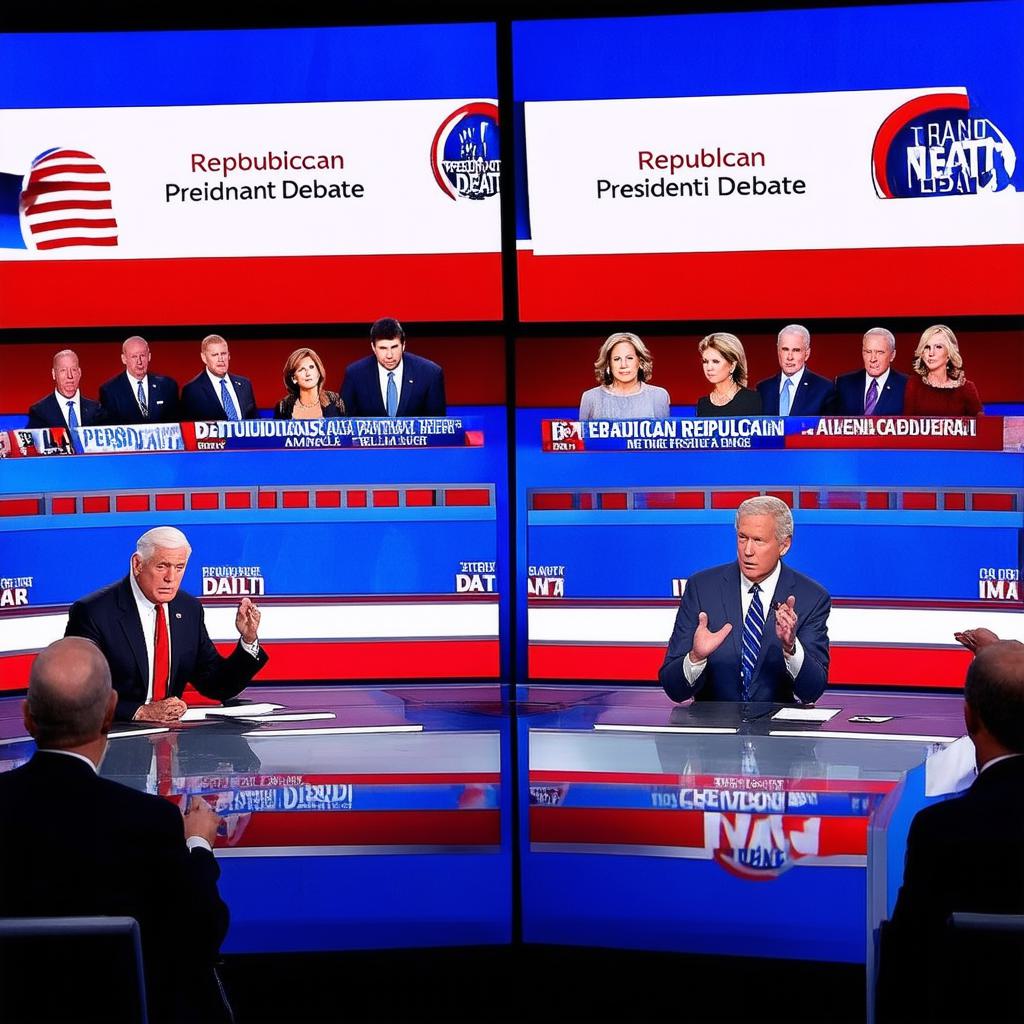}
        \caption*{Stable Diffusion 3}
        \label{fig:sd3-image}
    \end{minipage}
    \hfill
    \begin{minipage}{0.23\textwidth}
        \centering
        \includegraphics[width=\linewidth]{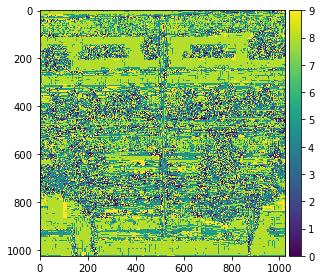}
        \caption*{LBP of Stable Diffusion 3}
        \label{fig:sd3-lbp}
    \end{minipage}

    \caption{Comparative analysis of texture patterns in images generated by different T2I models using Local Binary Pattern (LBP) representation.}
    \label{fig:LBP-horizontal}
\end{figure}

 Local Binary Pattern (LBP) is commonly used for texture analysis. AI-generated images vary in fine-grained texture generation. In Figure \ref{fig:LBP-horizontal} we can see that an image generated by Midjourney 6 has specific facial textures and subtle expression lines whereas image generated by SD3 has inconsistencies and lack of texture in certain areas. facial features, facial structures, hair lines, edges in clothing, and wrinkles are preserved in each segment for the Midjourney image but SD3 image completely lost it.

\subsection{Pairwise Scatter Plot Analysis}
\label{pairwise}
\begin{figure}[htbp]
    \centering
    \includegraphics[width=\linewidth]{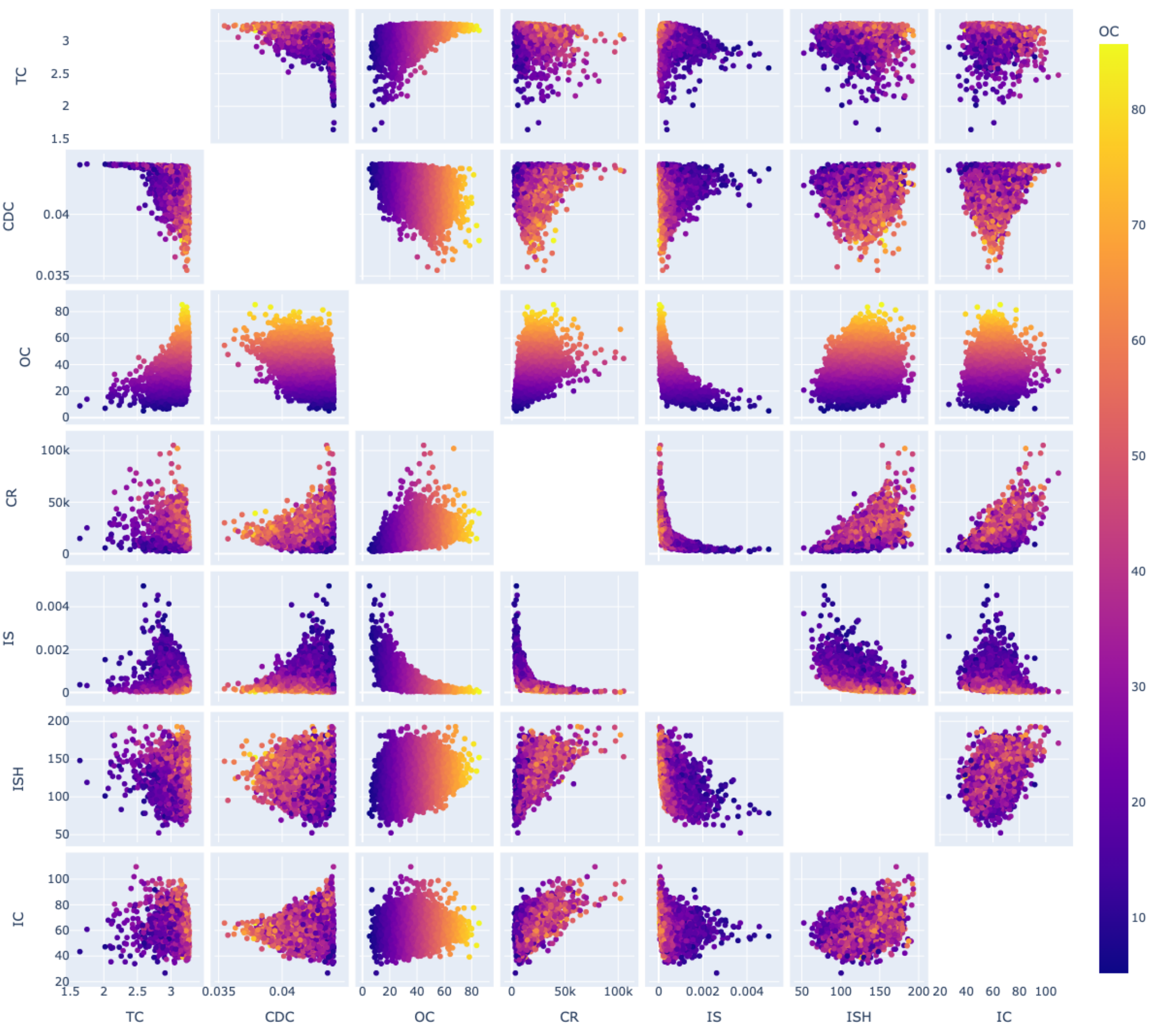}
    \caption{DALL·E 3}
    \label{fig:dalle}
\end{figure}

\begin{figure}[htbp]
    \centering
    \includegraphics[width=\linewidth]{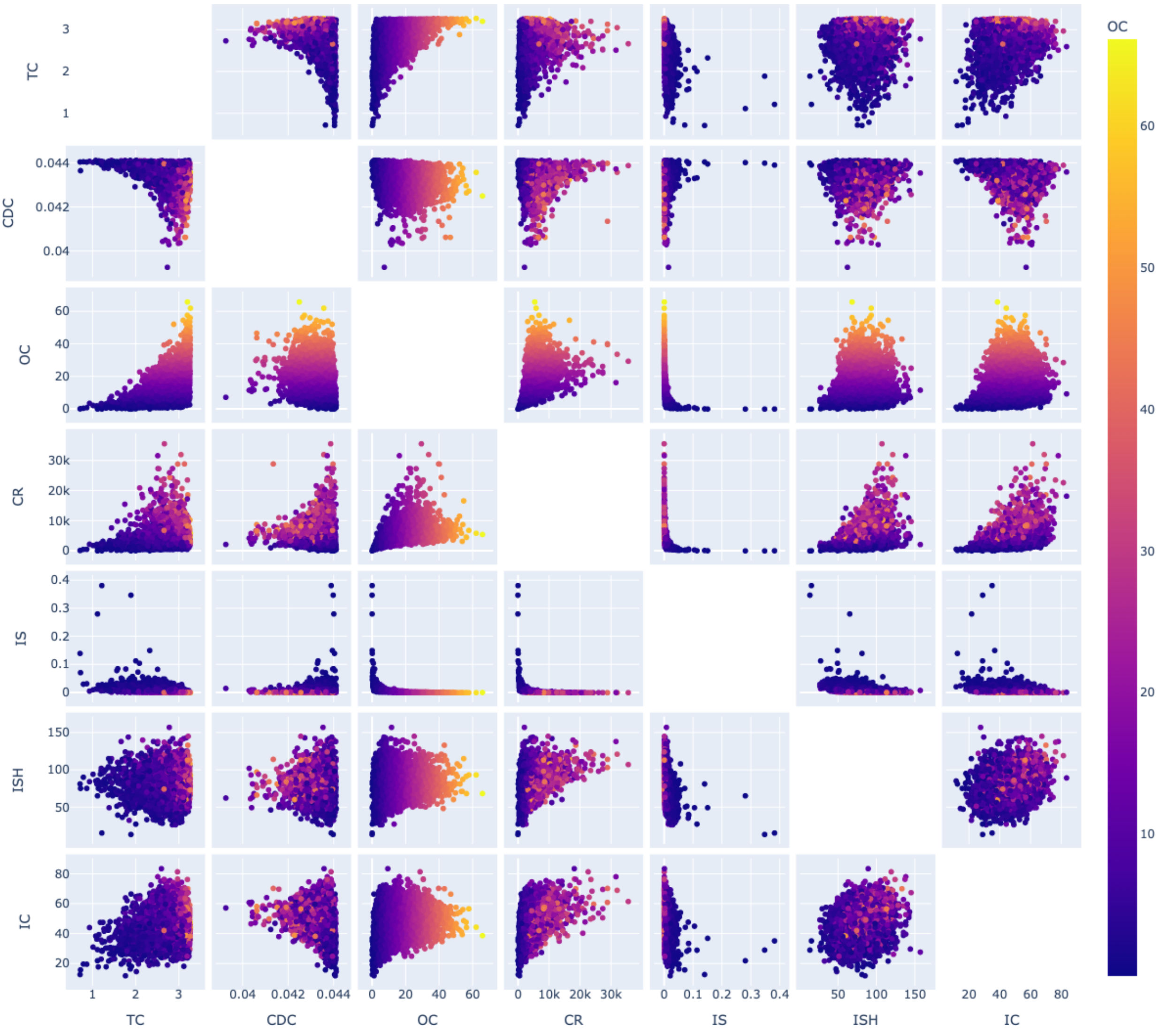}
    \caption{Midjourney 6}
    \label{fig:midjourney}
\end{figure}

\begin{figure}[htbp]
    \centering
    \includegraphics[width=\linewidth]{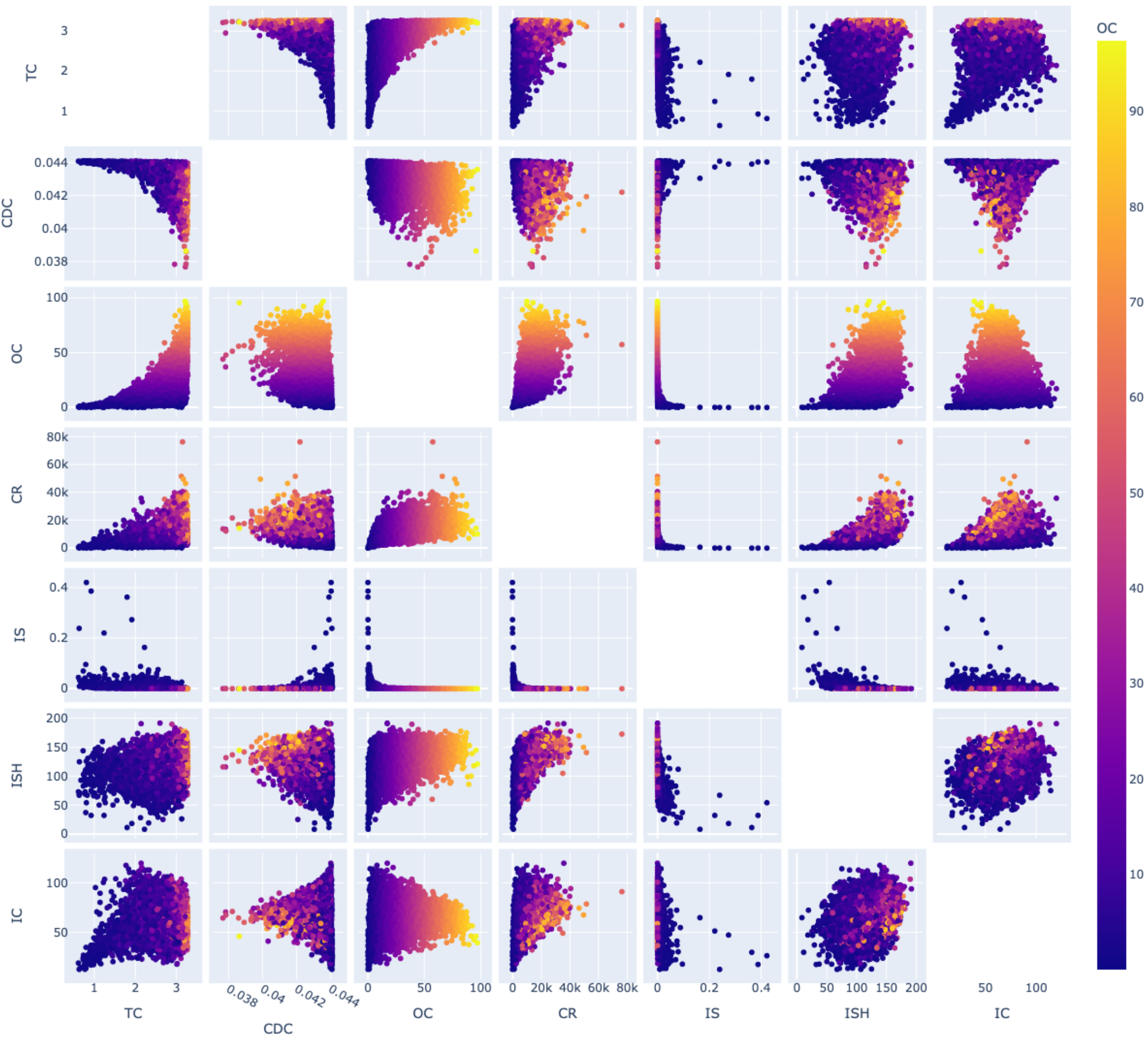}
    \caption{Stable Diffusion 3}
    \label{fig:sd3}
\end{figure}

\begin{figure}[htbp]
    \centering
    \includegraphics[width=\linewidth]{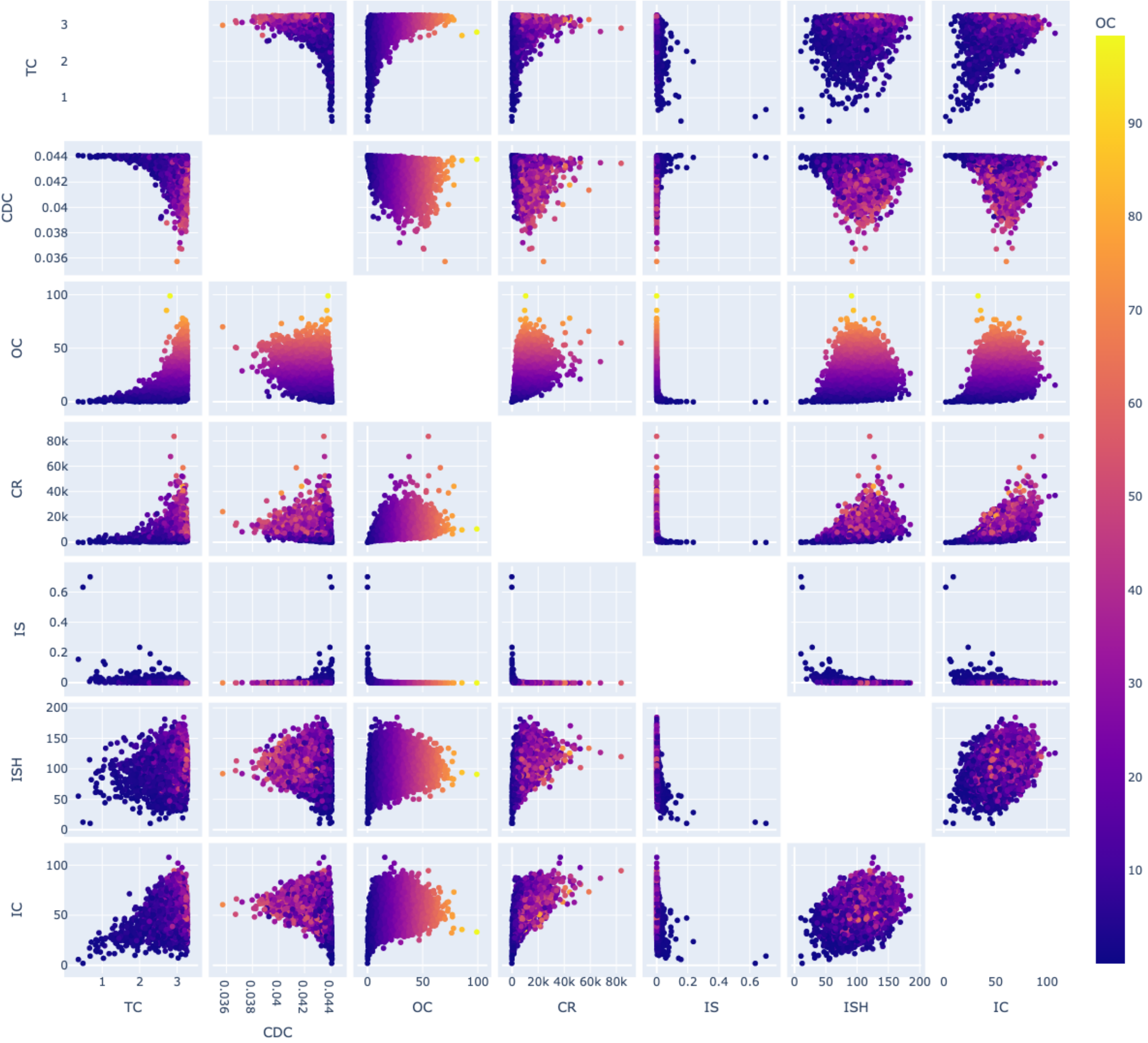}
    \caption{Stable Diffusion 2.1}
    \label{fig:sd21}
\end{figure}

\begin{figure}[htbp]
    \centering
    \includegraphics[width=\linewidth]{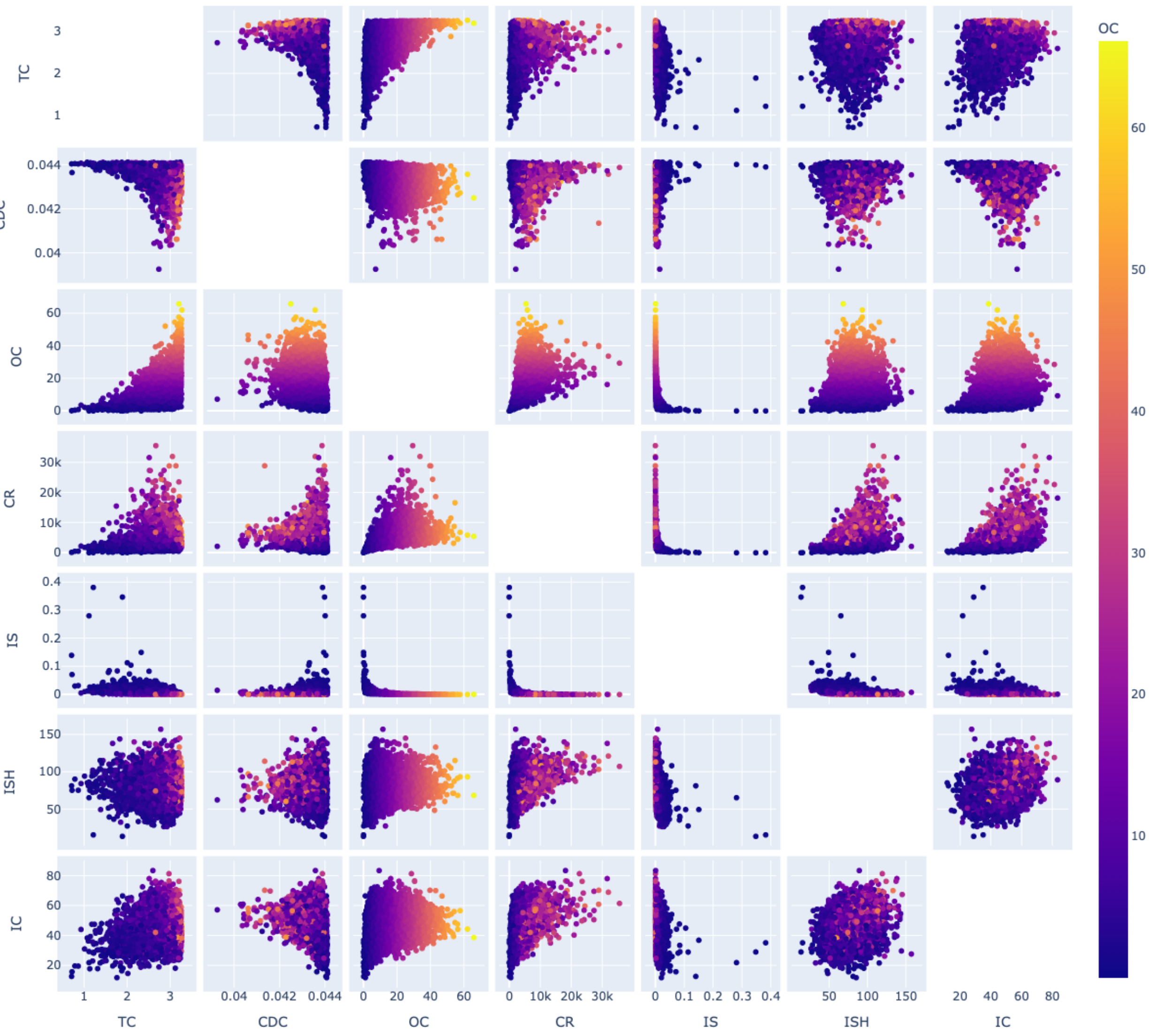}
    \caption{Stable Diffusion XL}
    \label{fig:sdxl}
\end{figure}

The pairwise scatter plots in Figures \ref{fig:dalle} through \ref{fig:sdxl} visualize relationships among different features such as texture complexity, object coherence, and contextual relevance across five text-to-image models.
DALL·E 3 shows a dense, compact cluster, indicating strong internal correlations between low- and high-level features and a high degree of structural stability.
Midjourney 6 produces a broader, moderately dispersed cluster, reflecting greater stylistic variability and weaker statistical coupling among features, consistent with its lower Visual AI Index.
Stable Diffusion 3, 3.5, and XL exhibit intermediate distributions, more organized than SD 2.1 but still fragmented, signifying partial improvements in texture-structure alignment.
Stable Diffusion 2.1 displays the widest and most irregular scatter, denoting high variance and minimal feature coherence.
Overall, the progression from SD 2.1 to DALL·E 3 illustrates a clear tightening of feature relationships and increasing internal consistency, highlighting how newer models achieve smoother integration of texture, structure, and semantic cues while reducing detectable artifacts.



\noindent

\setlength{\belowcaptionskip}{5pt} 

\noindent

\vspace{-5mm}



\noindent
\end{document}